\documentclass[10pt,journal,compsoc]{IEEEtran}

\usepackage[nocompress]{cite}

\usepackage{url}
\usepackage{ragged2e}
\usepackage{epsfig}
\usepackage{graphicx}
\usepackage{amsmath,amssymb} % define this before the line numbering.
\usepackage{mathptmx}      % use Times fonts if available on your TeX system
\usepackage{subfigure}
\usepackage{algorithm}
\usepackage{algorithmicx}
\usepackage{algpseudocode}
\usepackage{graphics}
\usepackage{mathrsfs}
\usepackage{threeparttable}
\usepackage{color}
\usepackage[normalem]{ulem}
\usepackage{multirow}
\usepackage{float}
\usepackage{amsfonts}
\usepackage{bm}
\usepackage{array}
\usepackage[table]{xcolor}
\usepackage{colortbl}
\usepackage{pifont}
\usepackage{diagbox}
\usepackage{rotating}
\usepackage{booktabs}
\usepackage{overpic}
\usepackage{textcomp}
\usepackage{contour}
\usepackage{enumitem}
\usepackage{stfloats}
\usepackage{csquotes}
\usepackage{wasysym}
\usepackage[colorlinks=true,breaklinks=true]{hyperref}

\usepackage[font=small,labelfont=bf,tableposition=top]{caption}

\usepackage{silence}
\def\ie{\emph{i.e.}}
\def\eg{\emph{e.g.}}

\def\etal{{\em et al.~}}

\definecolor{bblue}{rgb}{0,150,230}
\definecolor{mygray}{gray}{.92}

\def\Firstre#1{\textcolor{red}{#1}}
\def\Secondre#1{\textcolor{blue}{#1}}

\graphicspath{{./Imgs/}}
\DeclareGraphicsExtensions{.jpg,.pdf,.png}

\RequirePackage{silence}
\begin{document}

% \title{Bridging Salient Object Detection and Camouflaged Object Detection with Difficulty-aware Learning}
\title{Joint Salient Object Detection and Camouflaged Object Detection via Uncertainty-aware Learning}

\author{Aixuan Li,~
        Jing Zhang*,~
        Yunqiu Lv,~
        Tong Zhang,~
        Yiran Zhong,~
        Mingyi He,~
        Yuchao Dai*~\\
% <-this % stops a space
\IEEEcompsocitemizethanks{
\IEEEcompsocthanksitem A. Li, Y. Lv, M. He and Y. Dai are with School of Electronics and Information, Northwestern Polytechnical University, Xi'an, China and Shaanxi Key Laboratory of Information Acquisition and Processing.
\IEEEcompsocthanksitem J. Zhang is with School of Computing, the Australian National University, Canberra, Australia.
\IEEEcompsocthanksitem T. Zhang is with IVRL, EPFL, Switzerland. 
\IEEEcompsocthanksitem Y. Zhong is with Shanghai AI Laboratory, Shanghai, China. 
\IEEEcompsocthanksitem A preliminary version of this work appeared at~\cite{aixuan_cod_sod21}. Our code and data are available at: \url{https://npucvr.github.io/UJSCOD/}.  %https://github.com/JingZhang617/Joint_COD_SOD
\IEEEcompsocthanksitem A. Li and J. Zhang contributed equally.
\IEEEcompsocthanksitem Corresponding authors: Y. Dai (daiyuchao@gmail.com) and J. Zhang (zjnwpu@gmail.com). This research was supported in part by National Natural Science Foundation of China (62271410) and by the Fundamental Research Funds for the Central Universities.
}% <-this % stops an unwanted space
%\thanks{Manuscript received XXX; revised XXX.}
}

% The only time the second header will appear is for the odd numbered pages
% after the title page when using the twoside option.
% \twocolumn[{%
% \renewcommand\twocolumn[1][]{#1}%
% \maketitle

% }]

\IEEEtitleabstractindextext{%
\begin{abstract} 
\justifying
% Visual salient object detection (SOD) aims at finding the salient object(s) that attract human attention, while camouflaged object detection (COD) on the contrary intends to discover the camouflaged object(s) that hidden in the surrounding. These two tasks have been studied separately and are commonly believed to contradict each other. In this paper, surprisingly, we show that SOD and COD could be jointly learned within an uncertainty-aware framework. We first show that the easy samples in the COD dataset can serve as hard positive
% % hard \TZ{positive?} 
% samples to improve the robustness of SOD model. Then, based on the \enquote{contradicting} attributes of these two tasks, we introduce a \enquote{Similarity measure} module to explicitly model the \enquote{contradicting} attributes of these two tasks. Moreover, considering the uncertainty of labeling in both tasks' datasets,
% % and their different convergence rates, 
% we introduce an adversarial learning network to explicitly model the confidence of network predictions, leading to more interpretable predictions for downstream tasks.
% % \TZ{Can we add the function of confidence of networks predictions? what does this part do}.
% % , where the confidence serves as the weight of each task for effective multi-task learning.
% Experimental results on benchmark datasets demonstrate that our solution leads to state-of-the-art performance for both tasks.
Salient objects attract human attention and usually stand out clearly from their surroundings. In contrast, camouflaged objects
share similar colors or textures with the environment. In this case, salient objects are typically non-camouflaged, and camouflaged objects are usually not salient.
Due to this inherent \enquote{contradictory} attribute,
we introduce an uncertainty-aware learning pipeline to extensively explore the contradictory information of salient object detection (SOD) and camouflaged object detection (COD) via data-level and task-wise contradiction modeling.
We first exploit the \enquote{dataset correlation} of these two tasks and claim that the easy samples in the COD dataset can serve as hard samples for SOD to improve the robustness of the SOD model.
Based on the assumption that these two models should lead to activation maps highlighting different regions of the same input image, we further introduce a \enquote{contrastive} module with a joint-task contrastive learning framework to explicitly model the contradictory attributes of these two tasks. Different from conventional intra-task contrastive learning for unsupervised representation learning, our \enquote{contrastive} module is designed to model the task-wise correlation, leading to cross-task representation learning.
To better understand the two tasks from the perspective of uncertainty, we extensively investigate the uncertainty estimation techniques for modeling 
the main uncertainties of the two tasks, namely 
\enquote{task uncertainty} (for SOD) and \enquote{data uncertainty} (for COD), and aiming to effectively estimate the challenging regions for each task to achieve difficulty-aware learning. 
Experimental results on benchmark datasets demonstrate that our solution leads to both state-of-the-art performance and informative uncertainty estimation.

\end{abstract}

}

\maketitle

\IEEEdisplaynontitleabstractindextext
% \IEEEdisplaynontitleabstractindextext has no effect when using
% compsoc or transmag under a non-conference mode.

\IEEEpeerreviewmaketitle

\IEEEraisesectionheading{\section{Introduction}\label{sec:introduction}}
% \Jing{1.uncertainty modeling, reasonable uncertainty, representing awareness of prediction.
% 2. dataset interaction.
% 3. weakly supervised, single/dual?}
% \Jing{extend the introduction, discuss more about the relationship between cod and sod.}

\IEEEPARstart{V}{isual} salient object detection (SOD) aims to localize the salient object(s) of the image that attract human attention.
% , which usually have high contrast compared with their surroundings. 
The early work of saliency detection mainly relies on human visual priors based handcrafted features~\cite{itti1998model,Manifold-Ranking:CVPR-2013,wei2012geodesic} to detect high contrast regions. Deep SOD models~\cite{scrn_sal,wei2020f3net,wang2020progressive,basnet_sal} use deep saliency features instead of 
% visual prior based 
handcrafted features to achieve effective global and local context modeling, leading to better performance.
% compared with the handcrafted feature based techniques. 
In general, existing SOD models~\cite{scrn_sal,cpd_sal,zhou2020interactive,wei2020f3net,basnet_sal} focus on two directions: 1)
% Models belonging to the first direction focus on 
constructing effective saliency decoders~\cite{cpd_sal,zhou2020interactive} that facilitate high/low-level feature aggregation; and
2) designing appropriate loss functions~\cite{wei2020f3net,basnet_sal} to achieve structure-preserving saliency detection. 
% \TZ{Is this happening in encoder?} NO
% accurate saliency detection. 
% For the former, with the backbone features from
% % As SOD models usually take the 
% backbone networks, \eg~VGG16 \cite{vgg_network}, ResNet50 \cite{he2016deep} \etc~, as encoder, the large effort of many SOD models is then to design effective decoders for feature aggregation. Further, given the largely down-sampled backbone features, some other
% % the other direction of 
% saliency models aim to design appropriate loss functions~\cite{wei2020f3net,basnet_sal}, achieving structure-preserving saliency detection.
% The former digs into multi-level feature fusion for coarse accuracy saliency detection, while the latter focus more on fine-grained accuracy of saliency predictions.

% We argue that an effective training dataset and the awareness of model prediction considering the \enquote{subjective nature} \cite{ucnet_sal} of saliency can lead to more performance gain. Further, an explanatory model with extra output representing the awareness of it's prediction is more desirable than SOD models with only saliency map as output.
% the existing posting estimation based deep SOD models.
% that  even with no ground truth saliency map provided.

% To be qualified as a \enquote{salient} object,
% The salient objects are related to human visual attention, and saliency objects tend to have high contrast compared with its global and local context. 

\begin{figure}[tp]
%  \vspace{-5mm}
   \begin{center}
   \begin{tabular}{c@{ }}
   {\includegraphics[width=0.90\linewidth]{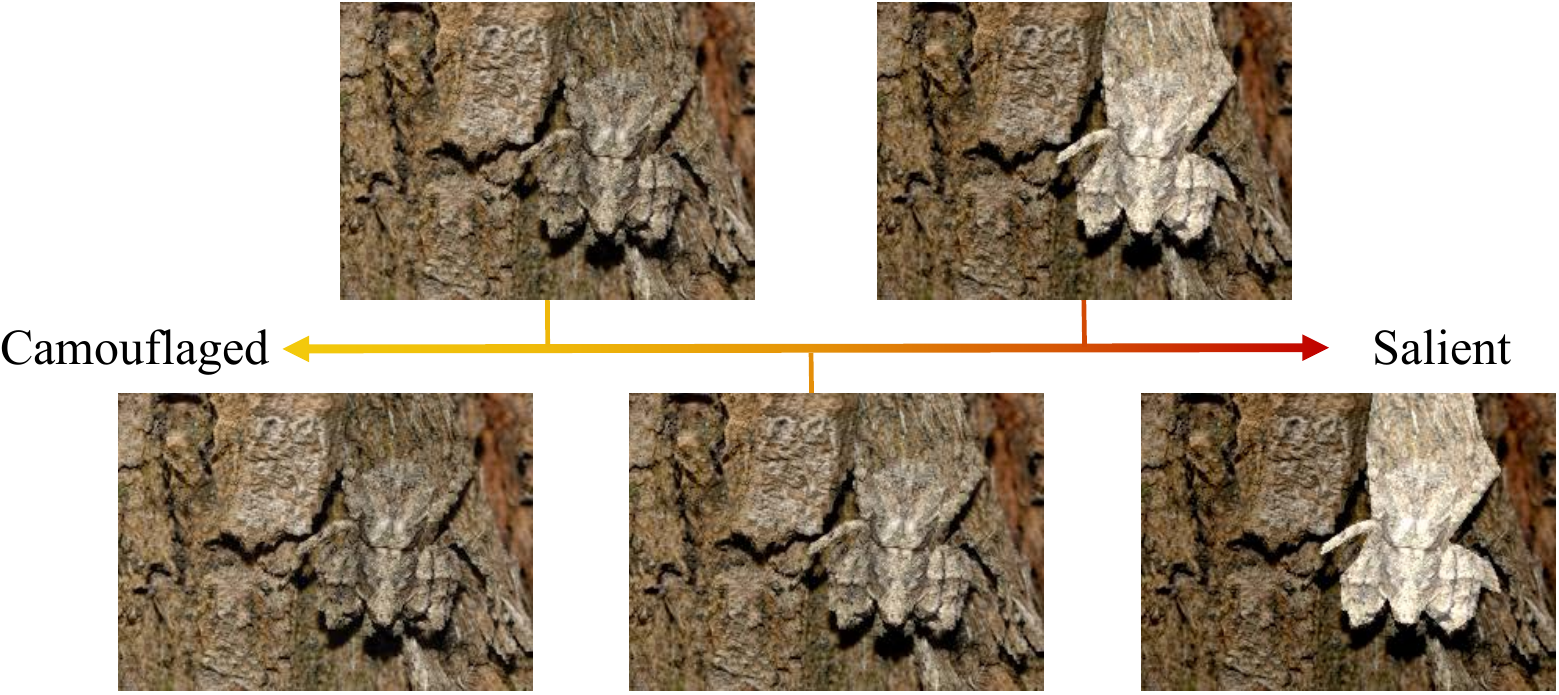}}
    \\
   \end{tabular}
   \end{center}
%   \vspace{-2mm}
   \caption{Illustration of the
%   Dataset level connection of camouflaged o We use an arrow to describe 
   transition from camouflaged objects to salient objects by gradually increasing the contrast of the foreground.
%   where the image
% %   the darker means more salient and the lighter means more camouflaged. 
% %   The images 
%   in the middle could belong to both camouflaged object dataset and salient object dataset.
%   \vspace{-2mm}
%   tasks.
%   appear in both tasks.
   }
   
%   From camouflaged object detection to salient object detection. The degree of red and yellow on the arrow represent the image 
   
%   The red and yellow arrows represent the degrees of salient or camouflaged respectively.
%   and the orange part is their overlapping part, the polar bear in the middle picture tries to hide but could still be found. And the pictures from left to right contain a snake, an owl, a polar bear, a snail, and a dog.
%   \vspace{-5mm}
\label{fig:relationship_sod_cod}
\end{figure}

Unlike salient objects that immediately attract human attention, camouflaged objects evolve to blend into their surroundings, effectively avoiding detection by predators.
% camouflaged objects evolve to hide in the environment, avoiding being noticed by predators. 
% with evolved similar pattern to it's habitat. 
The concept of camouflage has a long history~\cite{thayer1896law}, and finds application in various domains including biology~\cite{talas2020camogan,stevens2009animal,hanlon2007cephalopod}, military~\cite{lokken2020robustness,puzikova2008principles,lin2019metaheuristic} and other fields~\cite{duan2020adversarial,chen2017psychosocial}.
From a biological evolution perspective,
% the perspective of biological evolution,
% in order to avoid being noticed by the predator, 
prey species have developed adaptive mechanisms to camouflage themselves within their environment~\cite{hanlon1988adaptive,carvalho2006almost}, often by mimicking the structure or texture of their surroundings. These camouflaged objects can only be distinguished by subtle differences. Consequently, camouflaged object detection (COD) models~\cite{fan2020camouflaged,le2019anabranch,mei2021Ming,zhai2021Mutual,yunqiu_cod21} are designed to identify and localize these "subtle" differences, enabling the comprehensive detection of camouflaged objects.

To address the contradictory nature of SOD and COD, we propose a joint-task learning framework that  explores the relationship between these two tasks. Our investigation reveals an inverse relationship between saliency and camouflage, where a higher level of saliency typically indicates a lower level of camouflage, and vice versa. This oppositeness is clearly demonstrated in Fig.~\ref{fig:relationship_sod_cod}, where the object gradually transits from camouflaged to salient as the contrast level increases.  Hence, we explore the correlation of SOD and COD from both data-wise and task-wise perspectives.

\begin{figure*}[tp]
%  \vspace{-5mm}
   \begin{center}
   \begin{tabular}{{c@{ } c@{ } c@{ } c@{ } c@{ } c@{ } c@{ } c@{ } c@{ } c@{ } c@{ } c@{ }}}
   {\includegraphics[width=0.0765\linewidth, height=0.058\linewidth]{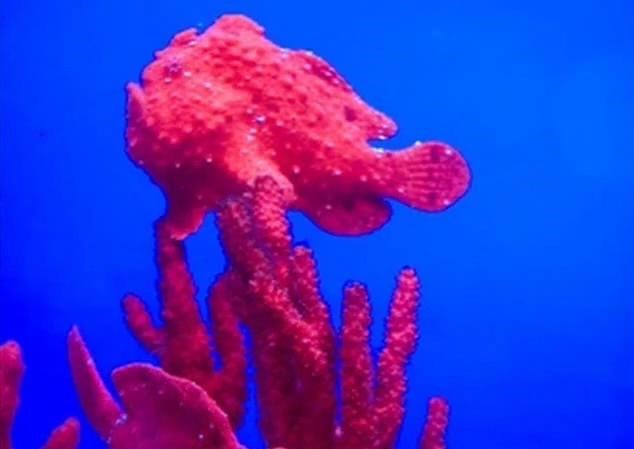}}&
   {\includegraphics[width=0.0765\linewidth, height=0.058\linewidth]{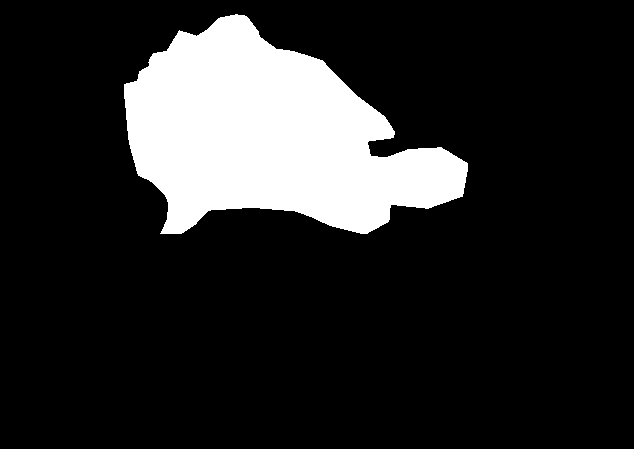}}&
   {\includegraphics[width=0.0765\linewidth, height=0.058\linewidth]{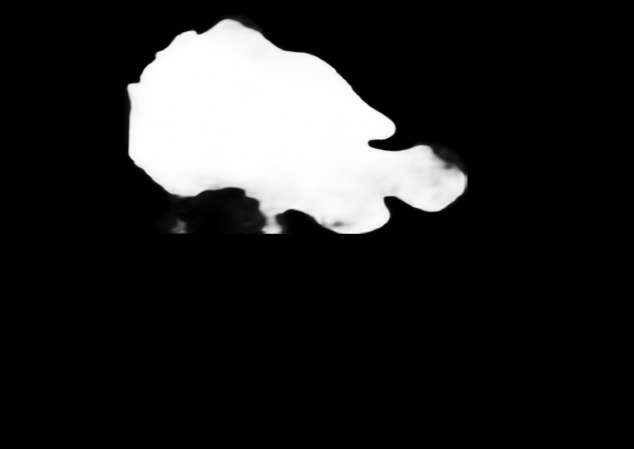}}&
   {\includegraphics[width=0.0765\linewidth, height=0.058\linewidth]{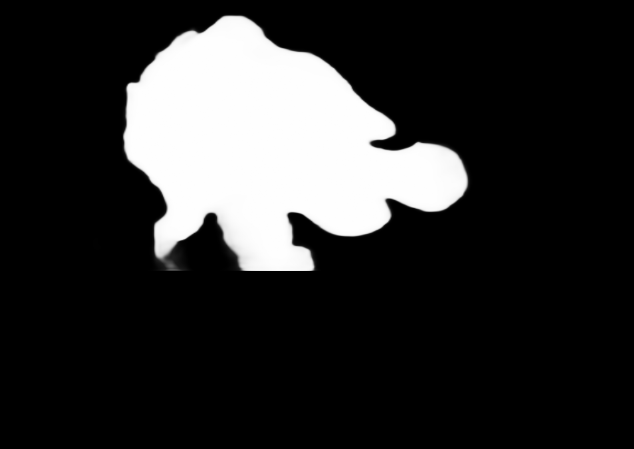}}&
   {\includegraphics[width=0.0765\linewidth, height=0.058\linewidth]{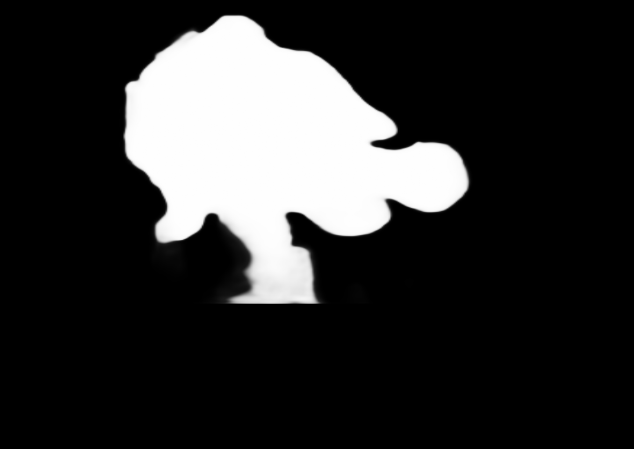}}&
   {\includegraphics[width=0.0765\linewidth, height=0.058\linewidth]{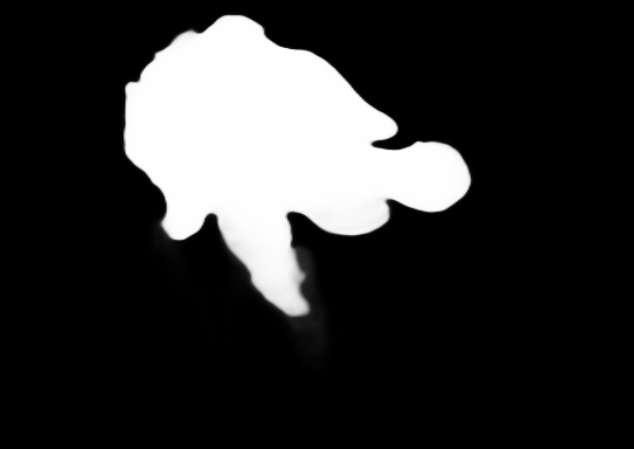}}&
   {\includegraphics[width=0.0765\linewidth, height=0.058\linewidth]{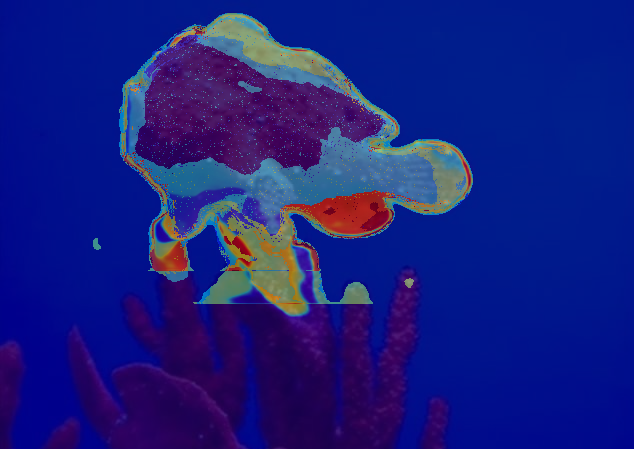}}&
   {\includegraphics[width=0.0765\linewidth, height=0.058\linewidth]{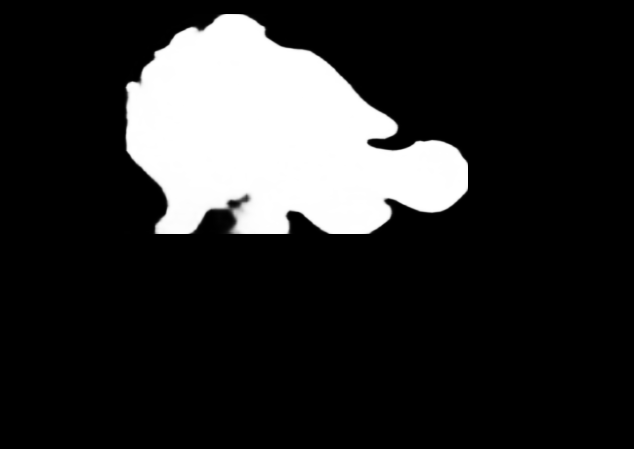}}&
   {\includegraphics[width=0.0765\linewidth, height=0.058\linewidth]{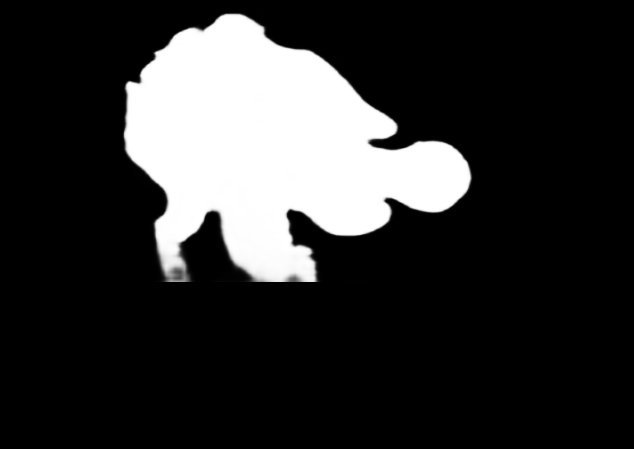}}&
   {\includegraphics[width=0.0765\linewidth, height=0.058\linewidth]{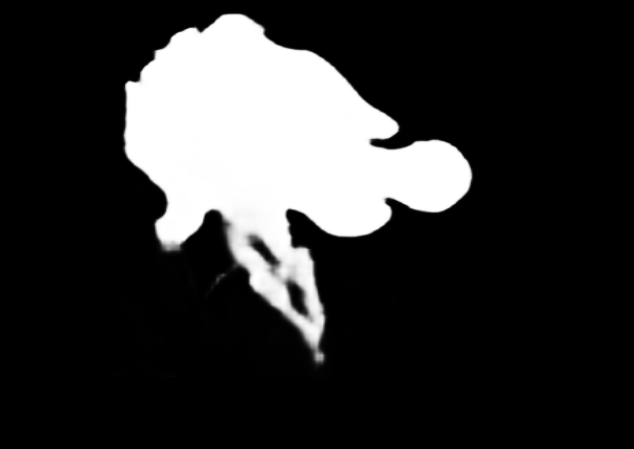}}&
   {\includegraphics[width=0.0765\linewidth, height=0.058\linewidth]{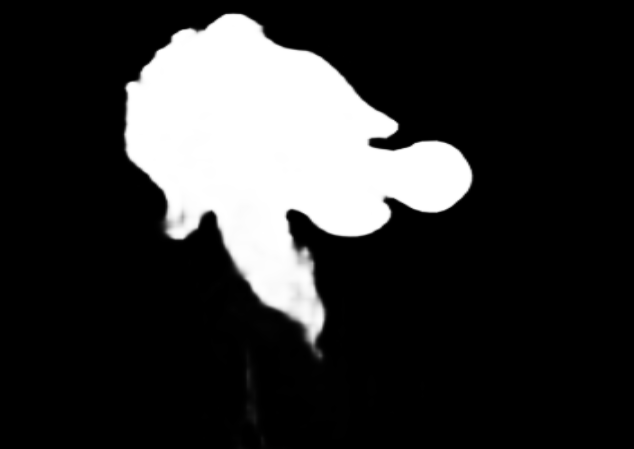}}&
   {\includegraphics[width=0.0765\linewidth, height=0.058\linewidth]{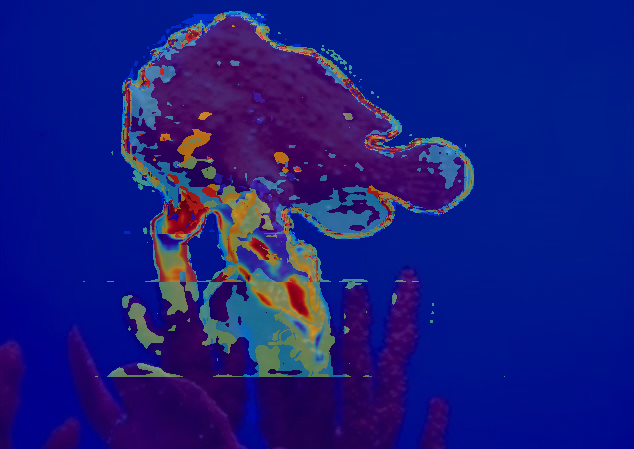}}\\
   
   {\includegraphics[width=0.0765\linewidth, height=0.058\linewidth]{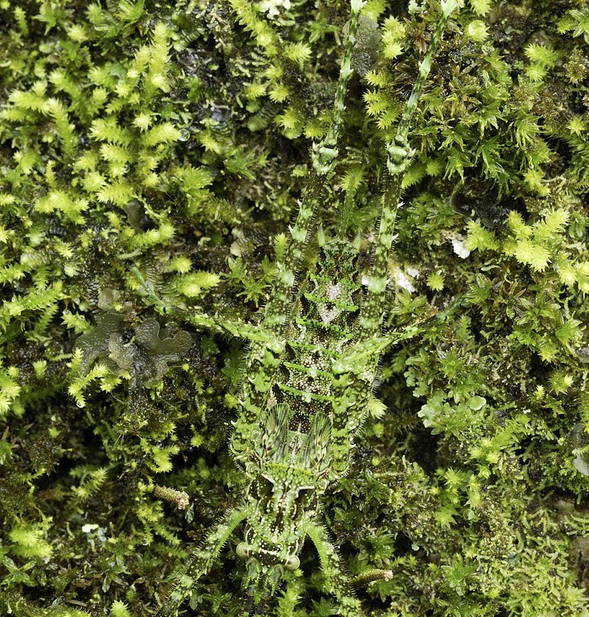}}&
   {\includegraphics[width=0.0765\linewidth, height=0.058\linewidth]{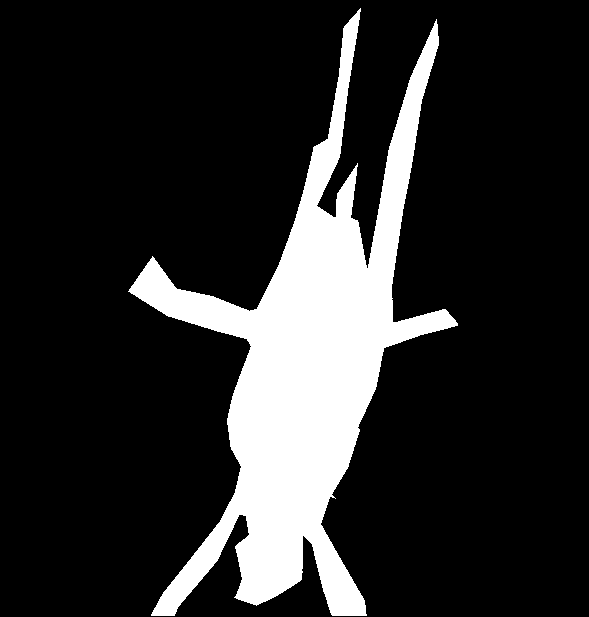}}&
   {\includegraphics[width=0.0765\linewidth, height=0.058\linewidth]{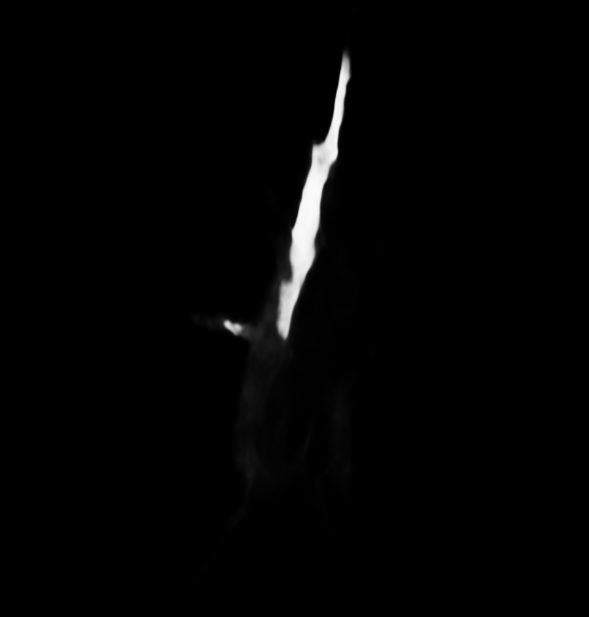}}&
   {\includegraphics[width=0.0765\linewidth, height=0.058\linewidth]{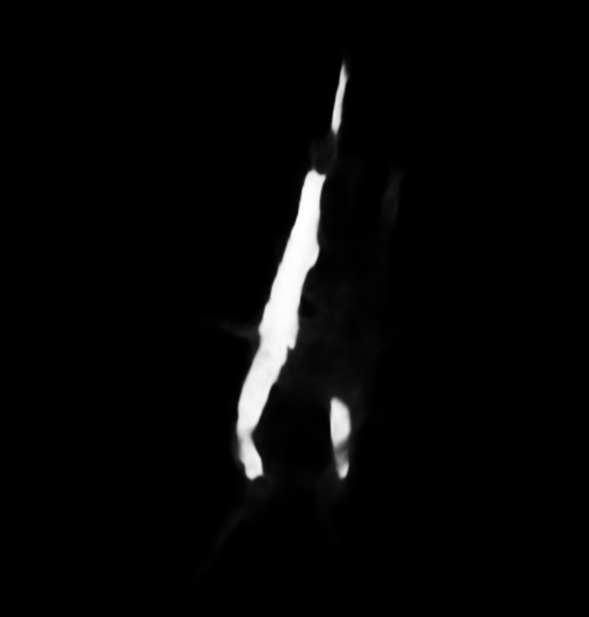}}&
   {\includegraphics[width=0.0765\linewidth, height=0.058\linewidth]{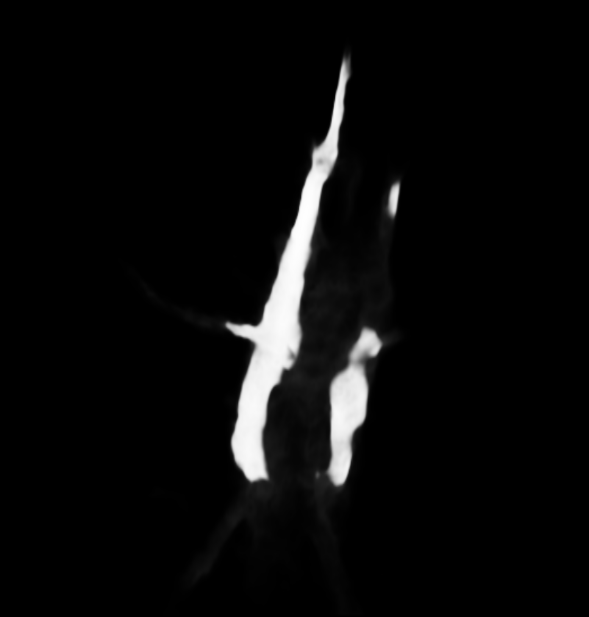}}&
   {\includegraphics[width=0.0765\linewidth, height=0.058\linewidth]{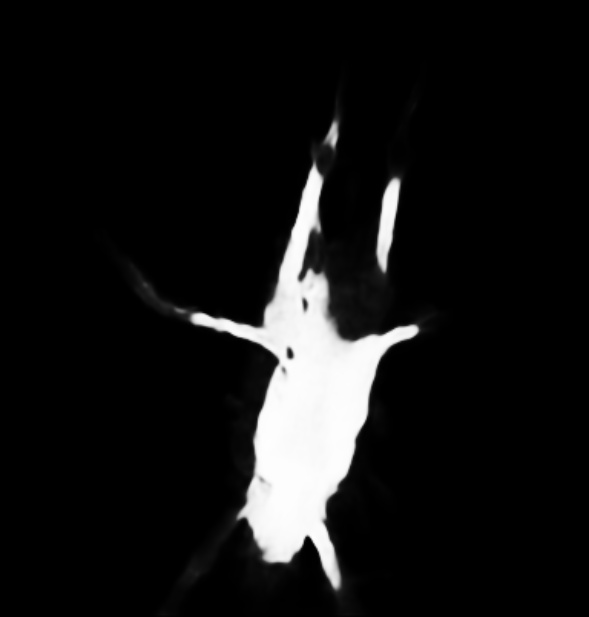}}&
   {\includegraphics[width=0.0765\linewidth, height=0.058\linewidth]{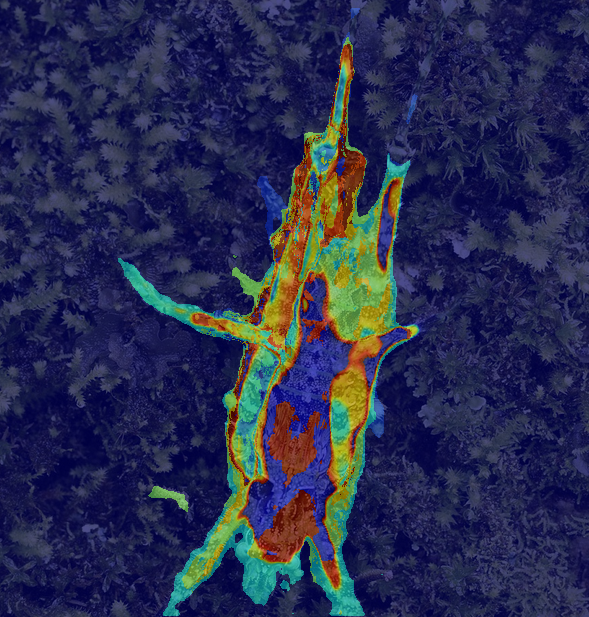}}&
   {\includegraphics[width=0.0765\linewidth, height=0.058\linewidth]{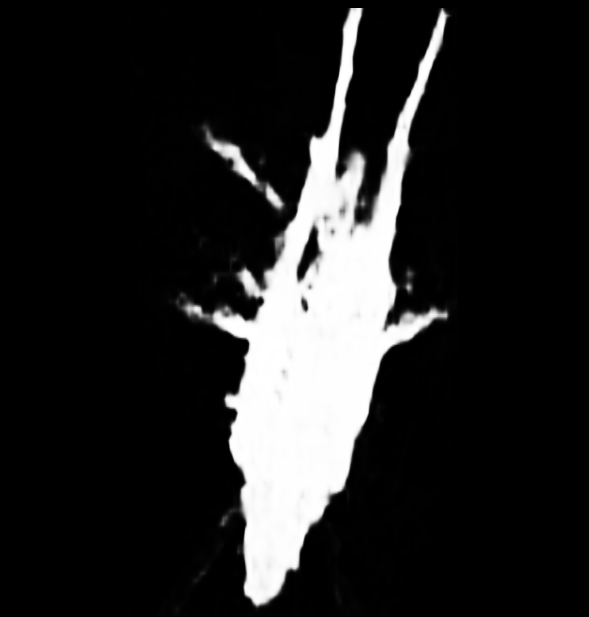}}&
   {\includegraphics[width=0.0765\linewidth, height=0.058\linewidth]{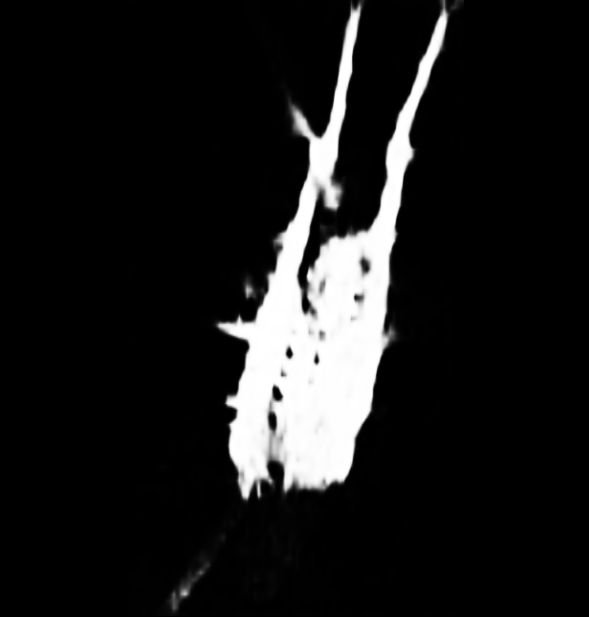}}&
   {\includegraphics[width=0.0765\linewidth, height=0.058\linewidth]{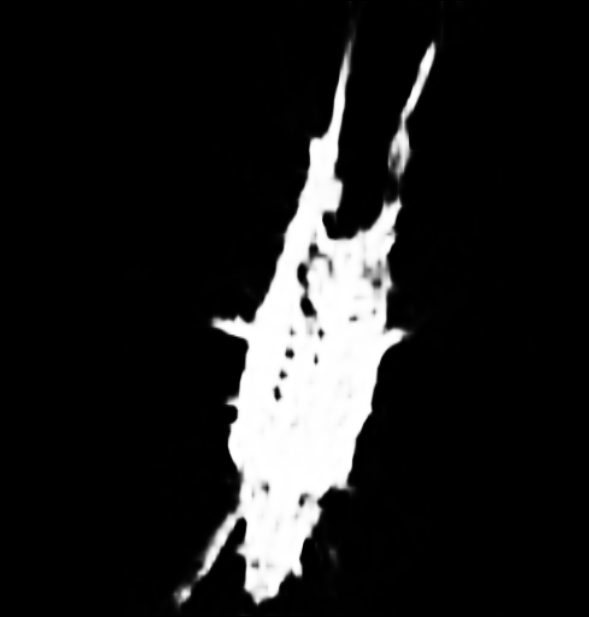}}&
   {\includegraphics[width=0.0765\linewidth, height=0.058\linewidth]{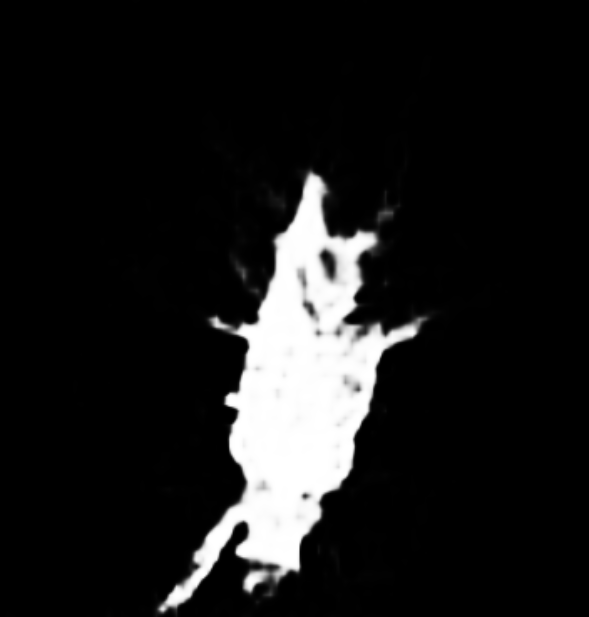}}&
   {\includegraphics[width=0.0765\linewidth, height=0.058\linewidth]{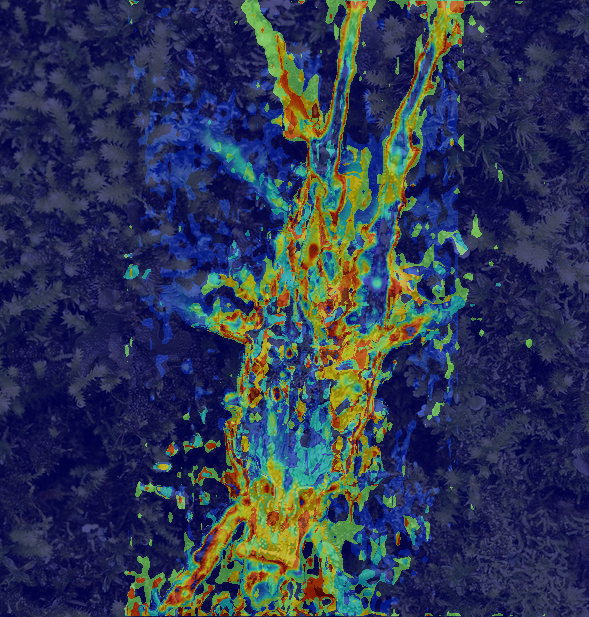}}\\

   {\includegraphics[width=0.0765\linewidth, height=0.058\linewidth]{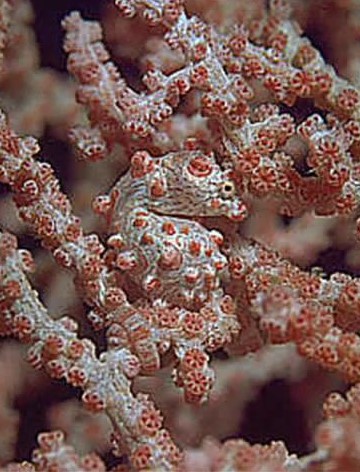}}&
   {\includegraphics[width=0.0765\linewidth, height=0.058\linewidth]{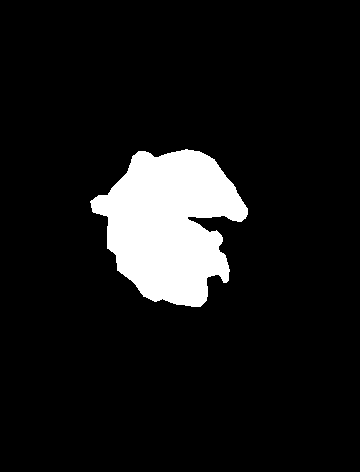}}&
   {\includegraphics[width=0.0765\linewidth, height=0.058\linewidth]{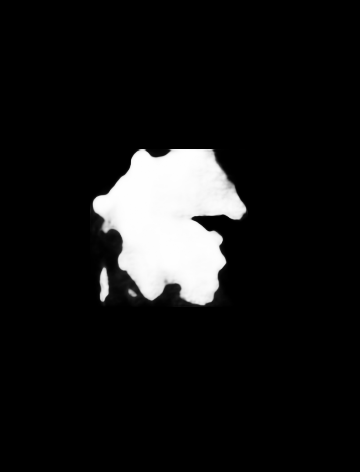}}&
   {\includegraphics[width=0.0765\linewidth, height=0.058\linewidth]{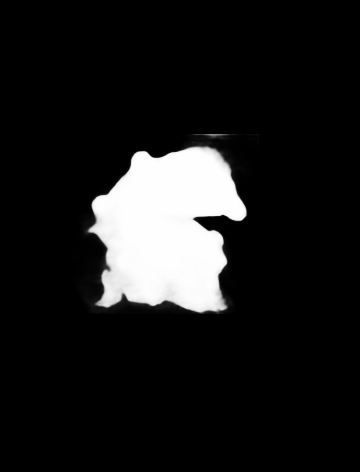}}&
   {\includegraphics[width=0.0765\linewidth, height=0.058\linewidth]{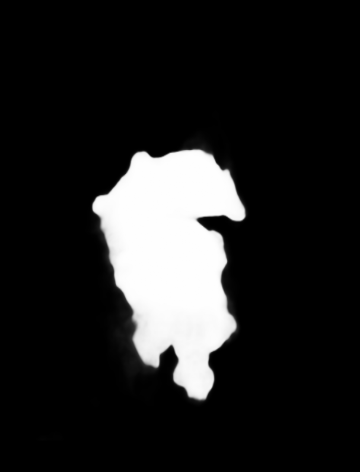}}&
   {\includegraphics[width=0.0765\linewidth, height=0.058\linewidth]{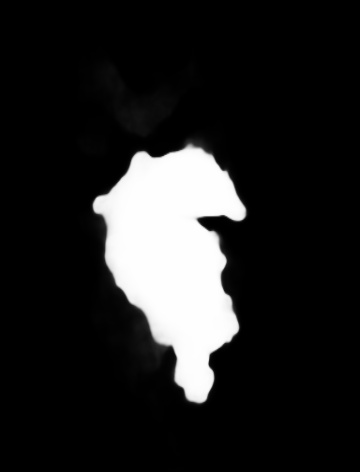}}&
   {\includegraphics[width=0.0765\linewidth, height=0.058\linewidth]{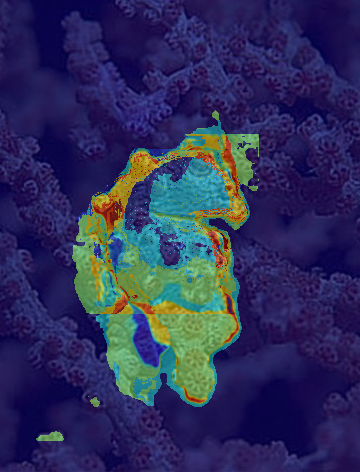}}&
   {\includegraphics[width=0.0765\linewidth, height=0.058\linewidth]{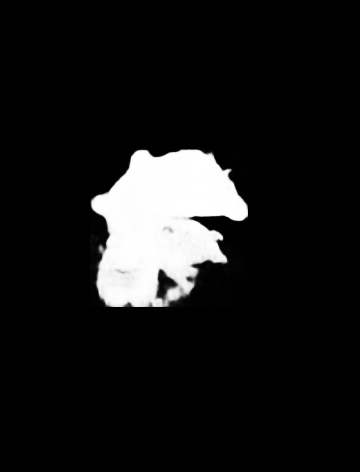}}&
   {\includegraphics[width=0.0765\linewidth, height=0.058\linewidth]{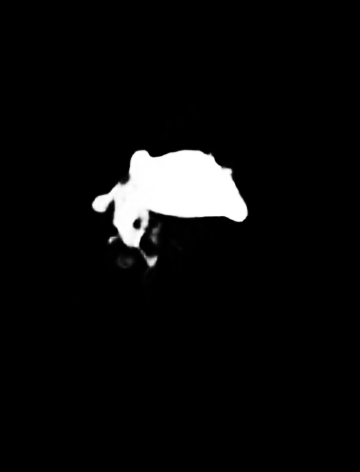}}&
   {\includegraphics[width=0.0765\linewidth, height=0.058\linewidth]{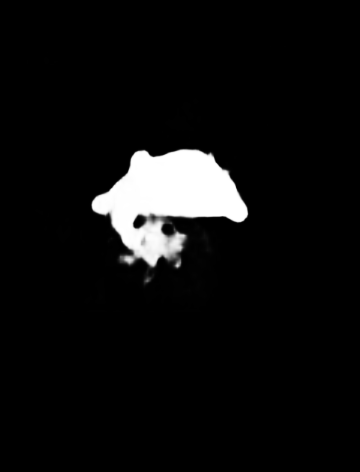}}&
   {\includegraphics[width=0.0765\linewidth, height=0.058\linewidth]{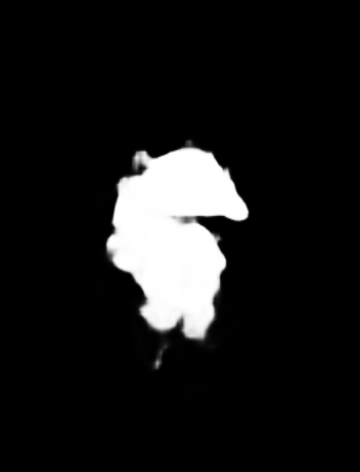}}&
   {\includegraphics[width=0.0765\linewidth, height=0.058\linewidth]{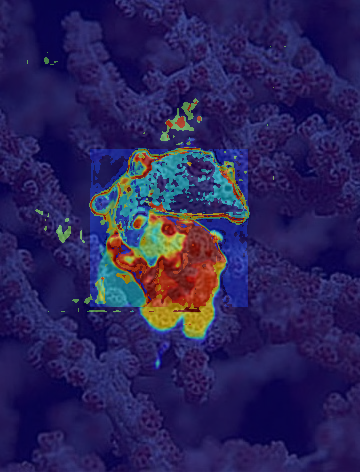}}\\

   \footnotesize{Image} &\footnotesize{GT} &\multicolumn{5}{c}{\footnotesize{ZoomNet \cite{ZoomNet_CVPR2022}}}&\multicolumn{5}{c}{\footnotesize{Ours}}\\
   \end{tabular}
   \end{center}
    \caption{Illumination of COD model performance for different percentages of background regions. For each method, from left to right are: image, ground truth, prediction of \enquote{CCrop}, \enquote{MCrop}, \enquote{LCrop},\enquote{Raw}  and the uncertainty maps of the four predictions. \enquote{CCrop} is achieved via cropping the minimal bounding box that contains the camouflaged objects. \enquote{MCrop} and \enquote{LCrop} are approximately 65\% and 80\% of the size of \enquote{Raw} image , respectively. Uncertainty maps represent the different
    % \YC{differential or different}
    predictive regions for the four types of cropped images.
    } 

    \label{fig:foreground_cropping_performance}
\end{figure*}

For data-wise correlation modeling, 
% we design \enquote{data interaction as data augmentation} by defining easy samples from COD as hard samples for SOD.
we re-interpret the data augmentation by defining easy samples from COD as hard samples for SOD. By doing so, we achieve contradiction modeling from the dataset perspective. Fig.~\ref{fig:relationship_sod_cod} illustrates that typical camouflaged objects are never salient, but samples in the middle can be defined as hard samples for SOD. 
Thus, we achieve context-aware data augmentation by the proposed \enquote{data interaction as data augmentation} method.
In addition, for COD, we find the performance is sensitive to the size of camouflaged objects. 
% Besides investigating the conventional uncertainty modeling techniques, we find the performance of COD is sensitive to the size of camouflaged objects. 
% As a context based task, we argue that the detectability of camouflage is sensitive to the complexity of the background. 
To explain this, we crop the foreground camouflaged objects with different percentages of background, and show their corresponding prediction maps and uncertainty maps in Fig.~\ref{fig:foreground_cropping_performance}. We observe that the cropping based prediction uncertainty, \ie~variance of multiple predictions, is relatively consistent with
% where the uncertainty map explains the ignorance of the model for different percentages of the background.
% Fig.~\ref{fig:foreground_cropping_performance} shows that the uncertainty maps are well-aligned with the 
region-level detectability of the camouflaged objects, validating that performance of the model can be influenced by the complexity of the background.
% For the compact cropping (\enquote{CCrop})...
% We also argue that camouflaged objects in the wild may appear in different environments. 
% The foreground-cropping strategy can serve as an effective data augmentation technique and a promising uncertainty generation strategy for COD.
The foreground-cropping strategy can serve as an effective data augmentation technique and a promising uncertainty generation strategy for COD, which also simulates real-world scenarios that camouflaged objects in the wild may appear in different environments.
We have also investigated the foreground cropping strategy for SOD, and observed relatively stable predictions, thus the foreground cropping is only applied to COD training dataset.
Aside from data augmentation, we integrate contrastive learning into our framework to address task-wise contradiction modeling. Conventional contrastive learning typically
% where positive/negative pairs are decided based on category information for category-related tasks~\cite{chopra2005learning,dimension_reduction_lecun,Distance_Metric_Learning,large_scale_online_learning,facenet,npair_loss}, 
% or whether derive from the same sample in self-supervised learning~\cite{khosla2020supervised,wen2021toward,wang2021dense},
constructs their positive/negative pairs based on semantic invariance. 
However, since both SOD and COD are class-agnostic tasks that rely on contrast-based object identification, we adopt a different approach for selecting positive/negative pairs based on region contrast. Specifically, given the same input image and its corresponding detected regions for the two tasks, we define region features with similar contrast as positive pairs, while features with different contrast serve as negative pairs. This contrastive module is  designed to cater to class-agnostic tasks and effectively captures the contrast differences between the foreground objects in both tasks.

\begin{figure}[tp]
%  \vspace{-5mm}
   \begin{center}
   \begin{tabular}{{c@{ } c@{ } c@{ } c@{ } c@{ }}}
   {\includegraphics[width=0.185\linewidth,height=0.115\linewidth]{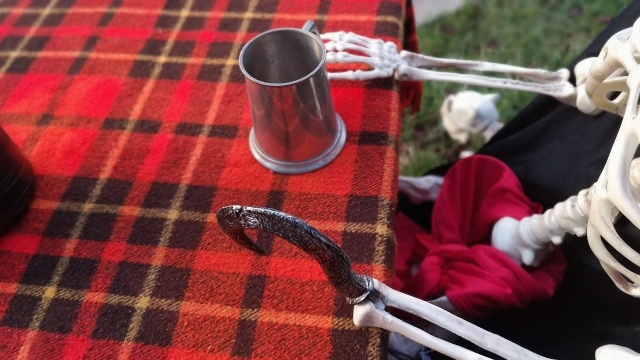}}&
   {\includegraphics[width=0.185\linewidth,height=0.115\linewidth]{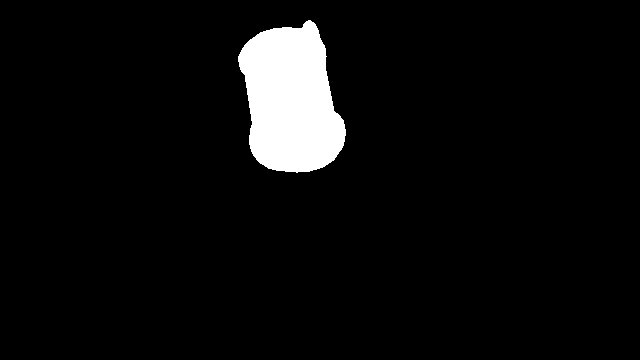}}&
   {\includegraphics[width=0.185\linewidth,height=0.115\linewidth]{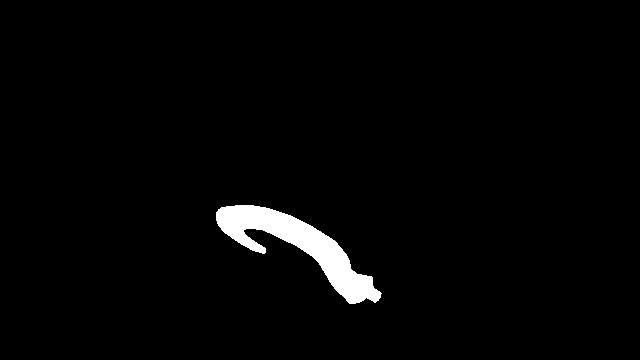}}&
   {\includegraphics[width=0.185\linewidth,height=0.115\linewidth]{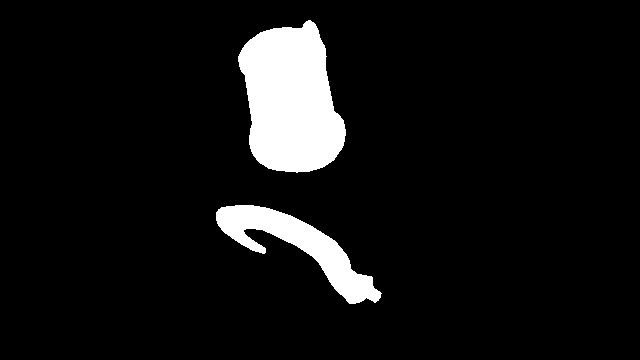}}&
    %   {\includegraphics[width=0.185\linewidth]{pic/figure2_uncertainty/sod/COME_Train_991_un_binary.png}}
   {\includegraphics[width=0.185\linewidth,height=0.115\linewidth]{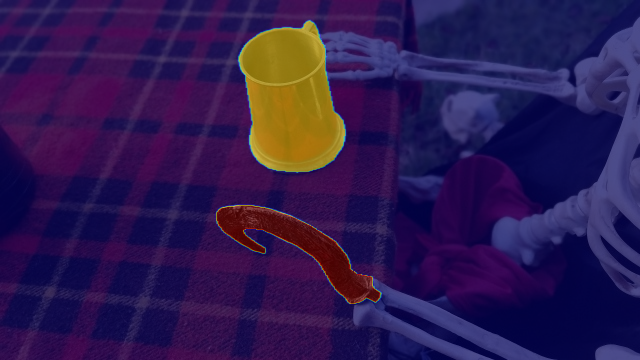}}\\
   {\includegraphics[width=0.185\linewidth,height=0.115\linewidth]{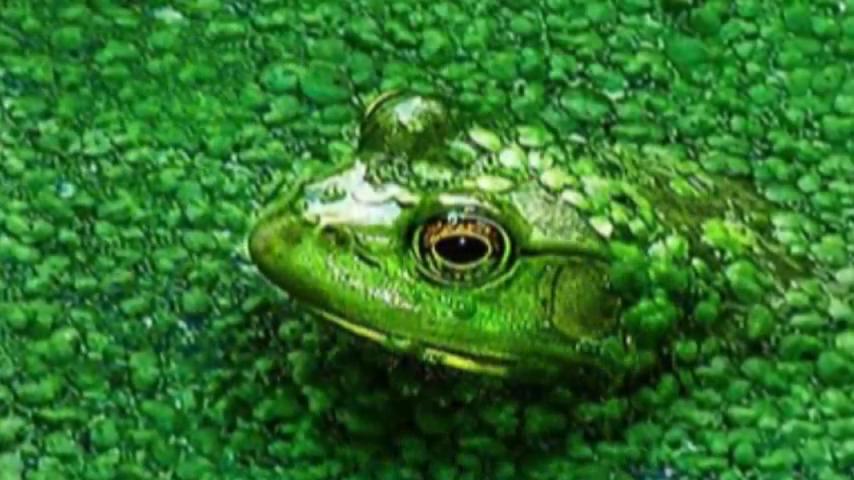}}&
   {\includegraphics[width=0.185\linewidth,height=0.115\linewidth]{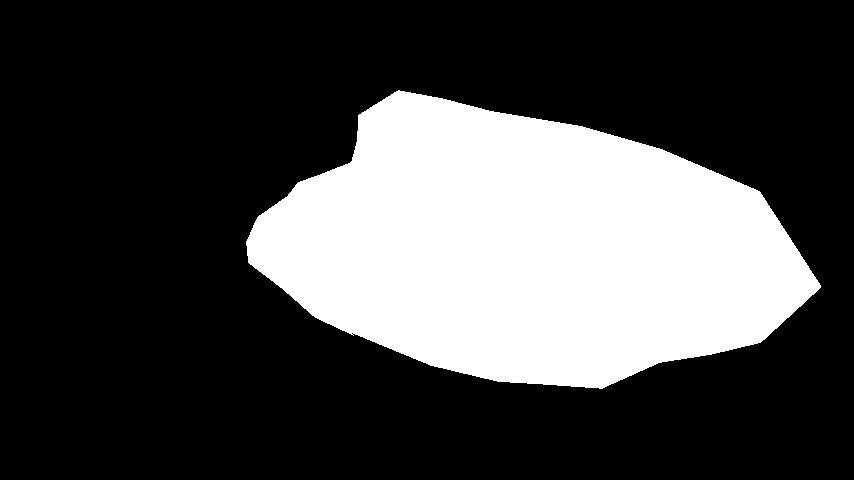}}&
   {\includegraphics[width=0.185\linewidth,height=0.115\linewidth]{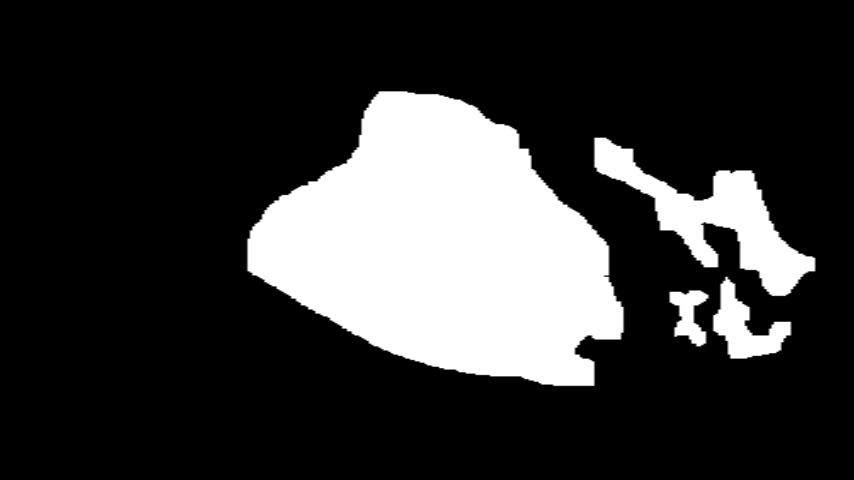}}&
   {\includegraphics[width=0.185\linewidth,height=0.115\linewidth]{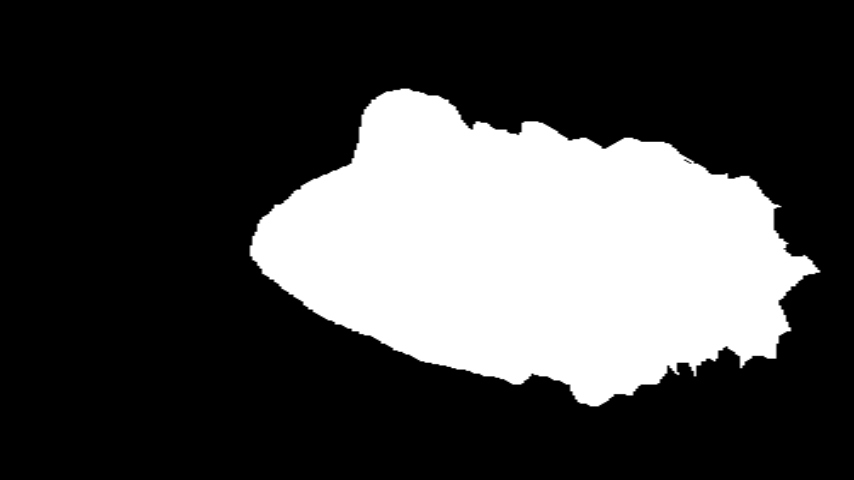}}&
    %   {\includegraphics[width=0.185\linewidth]{pic/figure2_uncertainty/cod/camourflage_00045_un_binary.png}}&
   {\includegraphics[width=0.185\linewidth,height=0.115\linewidth]{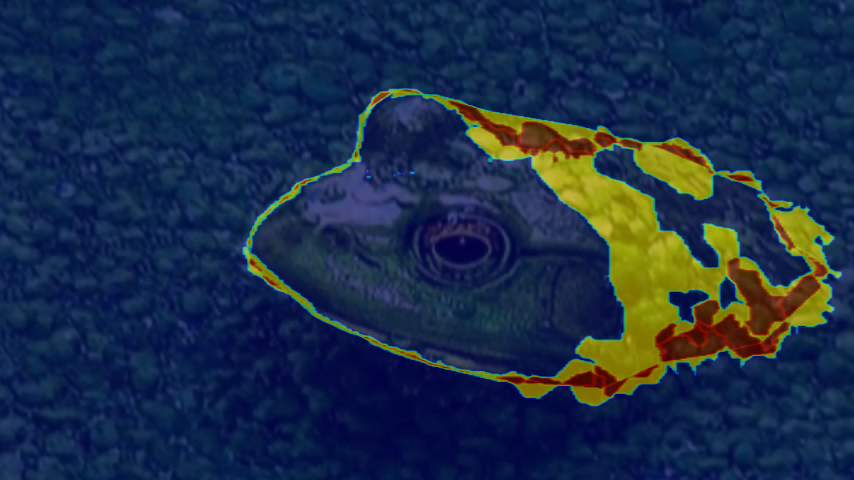}}\\
   \footnotesize{Image} &\footnotesize{GT} &\multicolumn{2}{c}{\footnotesize{Candidate Annotations}}&\footnotesize{Uncertainty}\\
   \end{tabular}
   \end{center}
    \caption{Inherent uncertainty for SOD ($1^{st}$ row) and COD ($2^{nd}$ row), where the \enquote{Candidate Annotations} represent two possible annotations for both tasks, and the \enquote{Uncertainty} is the \enquote{task uncertainty} ($1^{st}$ row) and \enquote{data uncertainty} ($2^{nd}$ row) respectively.
    } 
    \label{fig:task_uncertainty}
\end{figure}

Additionally, we observe two types of \enquote{uncertainty} for SOD and COD, respectively, as depicted in Fig.~\ref{fig:task_uncertainty}. For SOD, the subjective nature~\cite{ucnet_sal,zhang2020uncertainty,zhang2020learning} and the prediction uncertainty due to the\enquote{majority voting} mechanism in labeling procedure, which we define as \enquote{task uncertainty.} On the other hand, in COD, uncertainty arises from the difficulty of accurately annotating camouflaged objects due to their resemblance to the background, which we refer to as \enquote{data uncertainty.} To address these uncertainties, as shown in the fifth column of Fig.~\ref{fig:task_uncertainty}, we extensively investigate uncertainty estimation techniques to achieve two main benefits: (1) a self-explanatory model that is aware of its prediction, with an additional uncertainty map to explain the model's confidence, and (2) difficulty-aware learning, where the estimated uncertainty map serves as an indicator for pixel-wise difficulty representation, facilitating practical hard negative mining.

A preliminary version of our work appeared at~\cite{aixuan_cod_sod21}. Compared with the previous version, we have made the following extensions: \textbf{1)}: We have fully analyzed the relationship between SOD and COD from both dataset and task connection perspectives to further build their relationships. 
\textbf{2)}: To further investigate the cross-task correlations from the \enquote{contrast} perspective, we have introduced contrastive learning to our dual-task learning framework.
% \textbf{3)}: To explore the dataset correlation, we introduce a new hard sample selection criteria based on weighted mean absolute error (wMAE) for data interaction, where the selected samples from the COD dataset are included in the SOD training dataset, achieving hard negative mining for SOD. As the images in SOD training dataset can hardly be defined as samples for COD, 
% we perform random sampling based foreground-cropping for COD training dataset, yielding significant performance gains. 
\textbf{3)}: As an adversarial training based framework, we have investigated more training strategies for the discriminator, leading to more stable training.
% for We propose using the 
% residuals between ground truth and prediction, as the uncertainty value  for the prediction, and setting the uncertainty value  for the ground truth to an all-zero image. We utilize an adversarial learning module to help model the model uncertainty and inherent uncertainty.
\textbf{4)}: We have conducted additional experiments to fully explain the task connections, the uncertainty estimation techniques, the experiment setting, and the hyper-parameters.
% in our network. 

% \YC{We still need to summarize our main contributions. Think about the reviewers.}
Our main contributions are summarized as:
\begin{itemize}
    % \item We propose the first joint salient object detection and camouflaged object detection framework to explore the contradicting nature of these two tasks.
    
    % \item We design a contrastive module to model the task-wise correlation, and a data interaction strategy to achieve context-aware data augmentation for SOD. 
    
    % \item We propose random sampling-based foreground-cropping as the COD data augmentation technique to consider the local attribute of camouflage. 
    
    % \item We explore uncertainty-aware learning to produce uncertainty estimation with single forward.
    \item We propose that salient object detection and camouflaged object detection are tasks with opposing attributes for the first time and introduce the first joint learning framework which utilizes category-agnostic \enquote{contrastive module} to model the contradictory attributes of two tasks.
    
    \item Based on the transitional nature between saliency and camouflage, we introduce \enquote{data interaction as data augmentation} by defining simple COD samples as hard SOD samples to achieve context-aware data augmentation for SOD. 
    
    \item We analyze the main sources of uncertainty in SOD and COD annotations. In order to achieve reliable model predictions, we propose an \enquote{uncertainty-aware learning module} as an indicator of model prediction confidence.
    
    \item  Considering the inherent differences between COD and SOD tasks, we propose  \enquote{random sampling-based foreground-cropping} as the COD data augmentation technique to simulate the real-world scenarios of camouflaged objects, which significantly improves the performance.
    
%    \item From the perspective of the dataset for task relations, we introduce data interaction as a data augmentation, and propose a new sample selection principle to avoid selection bias due to the different sizes of foreground objects.
%    \item To achieve task-wise contradiction modeling, we introduce contrastive learning based on contrast information, implementing contrastive learning for category-agnostic tasks.
%    \item We summarise the sources of inherent uncertainty in SOD and COD respectively, and  propose a stable uncertainty modeling approach for reliable model generation.
%    \item Considering the difference between the properties of camouflaged and saliency, we propose a Foreground Cropping  data augmentation method to simulate real-world scenarios, which yields significant performance gains.
\end{itemize}

\section{Related Work}
% \YC{To be more concise.}

% We first introduce state-of-the-art salient object detection and camouflaged object detection models. Then we discuss multi-task learning frameworks and uncertainty-aware learning strategies.

% \Jing{we want to achieve confidence-aware learning. adversarial learning can be defined as one special case of confidence-aware learning. Introduce more work on confidence-aware leaning, difficulty-aware learning. Meanwhile, more work on cod and sod.}

\noindent\textbf{Salient Object Detection.} 
% Saliency detection models mainly focus on achieving mapping from the given input images to the corresponding ground truth saliency maps. 
Existing deep saliency detection models~\cite{cpd_sal,scrn_sal,wang2020progressive,Liu19PoolNet,wei2020f3net,feng2019attentive,basnet_sal} are mainly designed to achieve structure-preserving saliency predictions. \cite{basnet_sal,scrn_sal} introduced an auxiliary edge detection branch to produce a saliency map with precise structure information. Wei \etal~\cite{wei2020f3net} presented structure-aware loss function to penalize prediction along object edges. Wu \etal~\cite{cpd_sal} designed a cascade partial decoder to achieve accurate saliency detection with finer detailed information. Feng \etal~\cite{feng2019attentive} proposed a boundary-aware mechanism to improve the accuracy of network prediction on the boundary. There also exist salient object detection models that benefit from data of other sources.~\cite{wang2018salient,saliency_unified} integrated fixation prediction and salient object detection in a unified framework to explore the connections of the two related tasks. Zeng \etal~\cite{zeng2019joint} presented to jointly learn a weakly supervised semantic segmentation and fully supervised salient object detection model to benefit from both tasks. Zhang  \etal~\cite{high_resolution_sod} used two refinement structures, combining expanded field of perception and dilated convolution, to increase structural detail without consuming significant computational resources, which are used for salient object detection task on high-resolution images.
Liu \etal~\cite{liu2021samnet} designed the stereoscopically attentive multi-scale module to ensure the effectiveness of the lightweight salient object detection model, which uses a soft attention mechanism in any channel at any position, ensuring the presence of multiple scales and reducing the number of parameters.

% argued that semantic segmentation and saliency detection can be combined to achieve boosted performance for each task, and they then worked on the joint learning of 
% combined weakly supervised semantic segmentation with salient object detection to 

\noindent\textbf{Camouflaged Object Detection.}
% Camouflage models are designed to discover the camouflaged object(s) hidden in the surrounding. Different from salient objects, which are those attracting human attention, camouflaged objects are those trying to decrease the conspicuousness.
% that it should be the background. 
The concept of camouflage is usually associated with context~\cite{behrens1988theories,behrens2018seeing,cuthill2019camouflage}, and the camouflaged object detection models are designed to discover the camouflaged object(s) hidden in the environment.
%~\cite{cuthill2019camouflage} identified the signal processing process of the camouflage mechanism.
Cuthill \etal~\cite{cuthill2005disruptive} concluded that an effective camouflage includes two mechanisms: background pattern matching, where the color is similar to the environment, and disruptive coloration, which usually involves bright colors along edge, and makes the boundary between camouflaged objects and the background unnoticeable.
% to separate the body visually.
Bhajantri \etal~\cite{bhajantri2006camouflage} utilized co-occurrence matrix to detect defective. Pike \etal~\cite{pike2018quantifying} combined several salient visual features to quantify camouflage, which could simulate the visual mechanism of a predator.
Le \etal~\cite{le2019anabranch} fused a classification network with a segmentation network and used the classification network to determine the likelihood that the image contains camouflaged objects to produce more accurate camouflaged object detection. In the field of deep learning, Fan \etal~\cite{fan2020camouflaged} proposed the first publicly available camouflage deep network with the largest camouflaged object training set.
% Fan \etal~\cite{fan2020pranet} applied camouflaged object detection to polyp segmentation, using global information for initial guidance of the segmentation and mining edge cues with the help of reverse attention module.
Mei \etal~\cite{mei2021Ming} incorporated the predation mechanism of organisms into the camouflaged object detection model and proposed a distraction mining strategy.
% which first used the positioning module to locate areas that may be artifacts in the global picture, and then used the focus module to gradually add details to the predicted areas.
Zhai \etal~\cite{zhai2021Mutual} introduced a joint learning model for COD and edge detection based on graph networks, where the two modules simultaneously mine complementary information.
% for information interaction.
Lv \etal~\cite{lv2023towards} presented a triple-task learning framework to simultaneously rank, localize and segment the camouflaged objects.
% ,  the degree of camouflaged object, integrating the three tasks of ranking, segmenting, and localizing camouflaged object.

\noindent\textbf{Multi-task Learning.} 
The basic assumption of multi-task learning is that there exists shared information among different tasks. In this way, multi-task learning is widely used to extract complementary information about positively related tasks. Kalogeiton \etal~\cite{kalogeiton2017joint} jointly detected objects and actions in a video scene.
% Based on the observation that features learned for one task help learning the other one, they fuse the features extracted from their respective convolutional streams, and finally obtain the output for each task.
Zhen \etal~\cite{zhen2020joint} designed a joint semantic segmentation and boundary detection framework by iteratively fusing feature maps generated for each task with a pyramid context module.
% Besides, they fuse the edge map derived from semantic segmentation result and the module predicted semantic boundary to suppress the non-semantic edge.
In order to solve the problem of insufficient supervision in semantic alignment and object landmark detection, Jeon \etal~\cite{jeon2019joint} designed a joint loss function to impose constraints between tasks, and only reliable matched pairs were used to improve the model robustness with weak supervision.
Joung \etal~\cite{joung2020cylindrical} solved the problem of object viewpoint changes in 3D object detection and viewpoint estimation with a cylindrical convolutional network, which obtains view-speciﬁc features with structural information at each viewpoint for both two tasks.
Luo \etal~\cite{luo2020multi} presented a multi-task framework for referring expression comprehension and segmentation.

\noindent\textbf{Uncertainty-aware Learning.}
% For detection and segmentation tasks, t
% The difficulty of
% detecting and 
% segmenting different parts of the image is different. In this way, d
%  While the confidence-aware learning is to estimate the confidence of the network prediction, which can help people understand how confident the network is in generating predictions.
% Gal \etal~\cite{gal2016dropout} used the dropout method to approximate Bayesian theory, combined the Bayesian model with the deep learning model, and solved the problem that traditional Bayesian models require high computational costs in deep learning.
% Srivastava \etal~\cite{srivastava2014dropout} used the dropout method to help the network not to be overconfident during the test.
% ~\cite{moon2020confidence} proposed a correctness ranking loss as a regularization method to use the consistency of the confidence ranking and the predicted probability ranking to help alleviate the overconfidence of the classification task prediction.
% Instead of using confidence estimation methods directly, Hendrycks \etal~\cite{hendrycks2016baseline} used the probability distribution of the softmax output to help determine if the network is outputting incorrectly or out-of-distribution .
Difficulty-aware (or uncertainty-aware, confidence-aware) learning aims to explore the contribution of hard samples,
% part of the objects, thus the model can pay more attention to the difficult parts, 
leading to hard-negative mining \cite{lin2017focal},
% prompting the network to learn the hard-to-predict parts.
% Difficult perceptual 
% Uncertainty-aware learning 
which has been widely used in medical image segmentation~\cite{nie2019difficulty,li2020difficulty,huang2021difficulty,yu2020difficulty}, semantic segmentation~\cite{li2017not,Xie_2020_ACCV,lin2017focal}, and other fields~\cite{li2020accelerating}. To achieve difficulty-aware learning, one needs to estimate model confidence. To achieve this, Gal \etal~\cite{gal2016dropout} used Monte Carlo dropout (MC-Dropout) as a Bayesian approximation, where model uncertainty can be obtained with dropout neural networks. Deep Ensemble \cite{2016deep_ensemble,chitta2018adaptive_ensemble,osband2016deep_ensemble} is another popular type of uncertainty modeling technique, which usually involves generating an ensemble of predictions to obtain variance of predictions as the uncertainty estimation. With extra latent variable involved, the latent variable models \cite{vae_raw,CVAE,gan_raw} can also be used to achieve predictive distribution estimation, leading to uncertainty modeling.
% extracting information from existing models that has been thrown away so far. T  of the  dropout method to approximate Bayesian theory, combined the Bayesian model with the deep learning model, and solved the problem that traditional Bayesian models require high computational costs in deep learning.
% Srivastava \etal~\cite{srivastava2014dropout} used the dropout method to help the network not to be overconfident during the test.
% ~\cite{moon2020confidence} proposed a correctness ranking loss as a regularization method to use the consistency of the confidence ranking and the predicted probability ranking to help alleviate the overconfidence of the classification task prediction.
% Instead of using confidence estimation methods directly, Hendrycks \etal~\cite{hendrycks2016baseline} used the probability distribution of the softmax output to help determine if the network is outputting incorrectly or out-of-distribution .
% m, 
% to help models improve performance.
Following the uncertainty-aware learning pipeline, Lin \etal~\cite{lin2017focal} introduced focal loss to balance the contribution of simple and hard samples for loss updating.
% a modification of the cross-entropy loss function to help the network reduce its focus on simple samples and prevent the network from converging to the optimum due to too many simple samples when the ratio of simple samples to difficult samples is imbalanced.
Li \etal~\cite{li2017not} presented
% argued that the difficulty of each pixel segmentation is inconsistent, and therefore proposed
a deep layer cascade model for
% applied to the 
semantic segmentation
% task, which sended the difficult parts to subbranches and allowed the network 
to pay more attention to the difficult parts.
Nie \etal~\cite{nie2019difficulty}
% applied difficulty-aware learning to medical image segmentation, using 
adopted adversarial learning to generate confidence levels for predicting segmentation maps, and then used the generated confidence levels to achieve difficulty-aware learning.
% to represent the predicted difficulty of pixels in a sample to help the network focus on difficult pixels.
Xie \etal~\cite{Xie_2020_ACCV} applied difficulty-aware learning to an active learning task, where the difficult samples are claimed to be more informative.
% believed that active learning needs to select the samples that contain the most information, i.e. difficult to segment samples, so the error mask are considered as difficult to segment pixels, and is used to learn to segment the difficult graph, which is used to help select data for active learning.
% At the same time, we can consider to some extent that the part of the network that is not confident is usually a sample of what the network considers difficult, so that the perception of confidence-aware learning and the perception of difficulty-aware learning have a compatible part.

% As the deep learning can be regarded as a data regression model, the network is sometimes overconfident about the false predictions generated by the data. This overconfidence makes the deep model unreliable, so the application of this type of model in practice will bring risks.
% We want networks not to be overconfident about positions that are perceived as difficult or uncertain.

\noindent\textbf{Contrastive learning.} The initial goal of contrastive learning~\cite{chopra2005learning,dimension_reduction_lecun,Distance_Metric_Learning,large_scale_online_learning,facenet,npair_loss} is to achieve effective feature representation via self-supervised learning.
The main strategy to achieve this is through constructing positive/negative pairs via data augmentation techniques~\cite{xie2021propagate,li2021dense,leverage2022, van2021unsupervised,wang2021dense,o2020unsupervised,chaitanya2020contrastive,xie2021detco}, where the basic principle is that similar concepts should have similar representation, thus stay close to each other in the embedding space. On the contrary, dissimilar concepts should stay apart in the embedding space.
Different from augmentation based self-supervised contrastive learning, supervised contrastive learning builds the positive/negative pairs based on the given labels~\cite{Distance_Metric_Learning,large_scale_online_learning,npair_loss}. 
Especially for image segmentation, the widely used loss function is cross-entropy loss. However, it's well known that cross-entropy loss is not robust to labeling noise~\cite{wang2019symmetric} and the produced category margins are not separable enough for better generalizing. Further, it penalizes pixel-wise predictions independently without modeling the cross-pixel relationships. Supervised contrastive learning~\cite{khosla2020supervised} can fix the above issues with robust feature embedding exploration, following the similar training pipeline as self-supervised contrastive learning.

\section{Our Method}
% \Jing{present a big picture of the framework, and the motivation for each module.}

\begin{figure}[t!]
%  \vspace{-5mm}
   \begin{center}
   \begin{tabular}{c@{ }}
   {\includegraphics[width=0.95\linewidth]{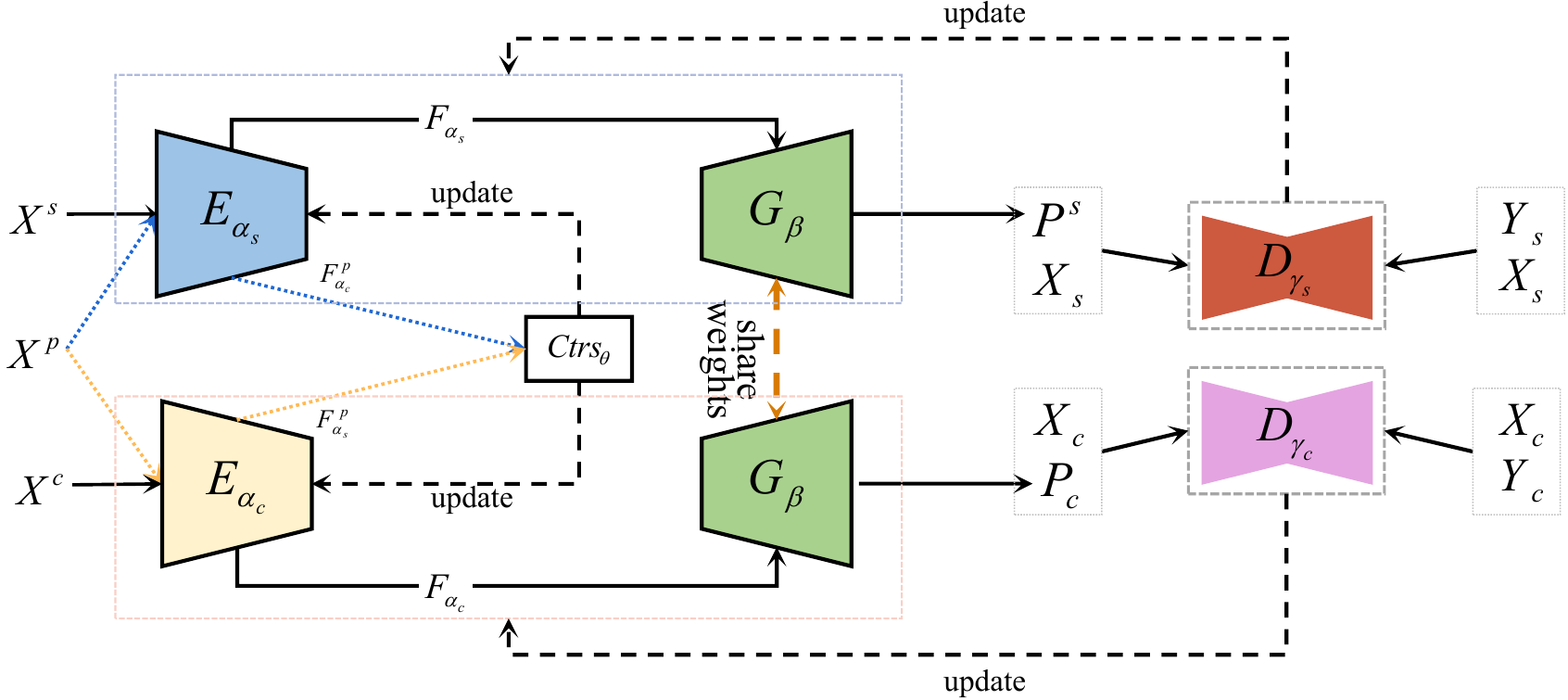}}
    \\
   \end{tabular}
   \end{center}
%   \vspace{-7mm}
   \caption{Overview of the proposed uncertainty-aware joint learning network. The \enquote{Feature Encoders} for SOD $E_{\alpha_{s}}$ and COD $E_{\alpha_{c}}$ are used to extract independent feature representation $F_{\alpha_{s}}$ and  $F_{\alpha_{c}}$, respectively. Given the feature representation of auxiliary image $X^p$, the contrastive module $\mathit{Ctrs}_{\theta}$ is presented to model the contradicting attributes of SOD and COD, achieving effective multi-task learning, and based on which the \enquote{Prediction Decoders} $G_{\beta}$ share weights to produce fine-grained SOD prediction $P_{s}$ and COD prediction $P_{c}$.
   % by sharing information. 
   % Further, a
   As an adversarial learning network, we extract confidence indicators from the two discriminators $D_{\gamma_{s}}$ and $D_{\gamma_{c}}$ for uncertainty-aware learning to distinguish the prediction from the ground truth.
   % whether the input is a \textit{prediction-image} pair or a \textit{ground truth-image} pair.
   % In addition, given the feature representation of auxiliary data image $X^p$, the task contrastive  module $\mathit{Ctrs}_{\theta}$ is presented to model the contradicting attribute of SOD and COD, achieving effective multi-task learning.
%   The \enquote{Feature encoder} module is used to extract task-specific features for image $X^s$ and $X^c$ from SOD and COD dataset respectively. And for the connection modeling data $X^p$, we introduce \enquote{Similarity measure} to explicitly model the contradicting attribute of SOD and COD.
%   The prediction feature of SOD and COD dataset will be fed to 
%   The shared \enquote{Prediction decoder} module is used to generate predictions for both tasks.
% %   a prediction for each task. 
%   The shared \enquote{Confidence estimation} module is a fully convolutional discriminator, which estimates the pixel-wise confidence of network prediction.  
   }
  \vspace{-2mm}
\label{fig:network_overview}
\end{figure}

We propose an uncertainty-aware joint learning framework via contrastive learning (see Fig.~\ref{fig:network_overview}) to learn SOD and COD in a unified framework. Firstly, we explain that these two tasks are both contradictory and closely related (Sec.~\ref{sub:task_analysis}), and a joint learning pipeline can benefit each other with effective context modeling. 
% as a data augmentation technique, we
% % we introduce adversarial training strategy and 
% select a group of easy samples from the COD training dataset
% % for salient object detection 
% to achieve robust SOD.
% salient object detection. 
Then, we present a \enquote{Contrastive Module} to explicitly model the \enquote{contradicting} attributes of these two tasks (Sec.~\ref{sub:contradicting_modeling}), with a data-interaction technique to achieve context-level data augmentation. Further, considering uncertainty for both tasks, we introduce a difficulty-aware learning network (Sec.~\ref{sub:adversarial_learning}) to produce predictions with corresponding uncertainty maps, representing the model's awareness of the predictions.
% We also observe that random-sampling based foreground cropping (Sec.~\ref{random_sampling_cropping}) is an effective data augmentation technique for camouflaged object detection.
% We will first introduce uncertainty estimation techniques and the uncertainty-aware learning pipeline for each task, and then we present the proposed joint learning framework.

\subsection{Tasks Analysis}
\label{sub:task_analysis}

% We discuss the correlation of the two tasks, serving as prior for our contrastive learning based joint task learning framework.
% Then, we analyze the inherent uncertainty within each task, especially the subjective nature of salient object detection (task uncertainty) and the labeling uncertainty of camouflaged object detection (data uncertainty), which is our basic assumption for the uncertainty-aware joint learning framework. 

\subsubsection{Tasks Relationship Exploration}
\label{sec:task_relationship}
% We analyse the relationship of salient object detection and camouflaged object detection to validate our joint learning framework.
\noindent\textbf{Model Perspective:}
At the task level, both SOD and COD are class-agnostic binary segmentation tasks, where a UNet \cite{unet} structure is usually designed to achieve mapping from input (image) space to output (segmentation) space. Differently, the foreground of SOD usually  stands out highly from the context,
% with higher contrast. W
while camouflaged instances are evolved to conceal in the environment.
% , and the contrast based techniques fail to detect the camouflaged instances. 
With the above understanding about both SOD and COD, we observe \enquote{complementary} information between the two tasks. Given the same image, we claim that due to the \enquote{contradicting} attributes of saliency and camouflage, the extracted features for each task should be different from each other, and the localized region of each task should be different as well.
% usually we can simply consider both sod and cod as dichotomous tasks, only the properties of the objects they want to segment are different, in other words, they are both a segmentation model, but positioned to different objects. Therefore, on a model level, SOD and COD can use the same model and furthermore, they can share the same segmentation module as long as the objects to be segmented are identified.Based on the above perspective, we have designed separate localisation modules for the SOD and COD models to determine the location properties of the object to be segmented, and once the object is located, both tasks use the same segmentation module to segment the located object.

\noindent\textbf{Dataset Perspective:}
At the dataset level, we observe some samples within the COD dataset can also be included in the SOD dataset (see Fig.~\ref{fig:selected_easy_samples_from_cod}), where the camouflaged region is consistent with the salient region. However, due to the similar appearance of foreground and background, these samples are easy for COD but challenging for SOD, making them effective for serving as
% which can then serve as 
hard samples for SOD to achieve hard negative mining. On the other side, most of the salient foreground in the SOD dataset has high contrast, and the camouflaged regions of the same image usually differ from the salient regions. In this way,
% \enquote{high contrast} attribute of foreground in SOD dataset, 
samples in the SOD dataset usually cannot serve as simple samples for COD.
% as the salient region 
% Given the \enquote{contradicting} attributes of the two tasks, we further observe different  Saliency and camouflage are two opposing object properties, but in fact, 
Considering the dataset relationships of both tasks, we claim that easy samples in the COD dataset can effectively serve as hard samples for SOD to achieve context-level data augmentation.
% the sharing and contradicting samples,
% At the same time, these two attributes have overlapping parts, and there is no clear demarcation between salience and camouflage, a similar point is made in \cite{le2019anabranch}. We find that there are difficult salient images in the camouflage dataset that share similar colours or textures with their surroundings and at the same time attract attention, so it is difficult to say exactly whether these objects are salient or camouflage objects, and these samples can be considered as difficult SOD datasets, so selecting these samples to add to the SOD dataset can help improve the robustness of the SOD model.

\subsubsection{Inherent Uncertainty}
\label{subsub:Inherent_Uncertainty}
\noindent\textbf{Subjective Nature of SOD:}
To reflect the human visual system,
% ensure the quality and validity of
% % the annotation of 
% the SOD dataset, 
the initial saliency annotation of each image is obtained with multiple annotators \cite{wang2017learning,yan2013hierarchical,Manifold-Ranking:CVPR-2013,li2015visual}, and then \enquote{majority voting} is performed to generate the final ground truth saliency map that represents the majority salient regions,
% the annotation of the SOD dataset is usually provided by more than one annotators to cross-check the correctness of the annotation \cite{wang2017learning,yan2013hierarchical,Manifold-Ranking:CVPR-2013,li2015visual}. 
\eg~the DUTS dataset \cite{wang2017learning}, ECSSD \cite{yan2013hierarchical}, DUT \cite{Manifold-Ranking:CVPR-2013} dataset are annotated by
% \Jing{50?} annotators,
% % at pixel level, similarly
% ECSSD\cite{yan2013hierarchical}, DUT\cite{Manifold-Ranking:CVPR-2013} annotated by 
five annotators and HKU-IS~\cite{li2015visual} is annotated by three annotators.
Further, to maintain consistency of the annotated data, some SOD datasets adopt the pre-selection strategy, where the images contain no common salient regions across all the annotators
% with significant saliency disagreement
will be removed before the labeling process,
% some SOD datasets undergo a pre-selection process of salient objects or images before pixel-level annotation.
\eg~HKU-IS~\cite{li2015visual} dataset first evaluates the consistency of the annotation of the three annotators, and removes
% removed 
the images with greater disagreement. In the end, 4,447 images are obtained from an initial dataset with 7,320 images.
% , then voted on the salient objects, and annotated the objects that more than two people considered to be significant as the final ground-truth, and  SOC dataset first had 5 people draw bounding boxes for objects considered salient, selected consistent salient objects, and then had 10 people perform pixel-level labeling.
We argue that both the majority voting process for final label generation and the pre-selection process for candidate dataset preparation introduce bias to both the dataset and the models trained on it. We explain this as the \enquote{subjective nature} of saliency.

% In this approach of selecting the most common object as the final ground truth, the process of voting for the salient object introduces data uncertainty.

\noindent\textbf{Labeling Uncertainty of COD:}
% Camouflage is related to biological evolution, where
Camouflaged objects are evolved to have similar texture and color information to their surroundings \cite{behrens1988theories,cuthill2005disruptive}.
% thus the risk of detection by predators can be reduced.
% , and therefore camouflaged objects have similar texture, structure, and colour information to their surroundings due to their properties \cite{behrens1988theories,cuthill2005disruptive}.
Due to the similar appearance of camouflaged objects and their habitats, it's more difficult to accurately annotate the camouflaged instance than generic object segmentation, especially along instance boundaries. This poses severe and inevitable labeling noise while generating the camouflaged object detection dataset, which we define as \enquote{labeling uncertainty} of camouflage.
% Further, images in the existing camouflaged object detection dataset are collected from the Internet, where 
% The camouflage dataset contains images where the camouflaged object is clearly visible and also where the camouflaged object is extremely difficult to detect \cite{Chameleon2018}, so it is extremely difficult to annotate with very precise pixel-level annotations, especially at the junction of camouflaged objects and the environment. 
% Compared to the SOD dataset, where salient objects typically have higher contrast, camouflaged objects have much greater uncertainty in pixel-level labelling.

\subsection{Joint-task Contrastive Learning}
\label{sub:contradicting_modeling}
% \subsubsection{Overview-New}
As a joint learning framework, we have two sets of training dataset for each individual task, namely a SOD dataset $D_s=\{x_i^s,y_i^s\}_{i=1}^{N_s}$ for SOD and a COD dataset $D_c=\{x_i^c,y_i^c\}_{i=1}^{N_c}$ for COD, where $\{x_i^s,y_i^s\}$ is the SOD image/ground truth pair and $\{x_i^c,y_i^c\}$ is the COD image/ground truth pair, and $i$ indexes images, $N_s$ and $N_c$ are the size of training dataset for each task.
% The goal of our joint learning framework is then to explore the correlation between the two tasks for effective multi-task learning. 
Motivated by both the task contradiction and data sharing attributes of the two tasks,
% and their inherent task/labeling uncertainty,
we introduce a contrastive learning based joint-task learning pipeline
% uncertainty-aware joint learning framework 
for joint salient object detection and camouflaged object detection. Firstly, we
% aim to 
model the \enquote{task contradiction}  (Section \ref{subsub_contrative_learning_subsec}) with a 
% contrastive learning based 
\enquote{contrastive module}. Secondly, we select easy samples by weighted MAE from the COD training dataset (Section \ref{subsub:data_interaction}), serving as hard samples for SOD.

\subsubsection{Task Correlation Modeling via Contrastive Learning}
\label{subsub_contrative_learning_subsec}
To model the task-wise correlation, we design a \enquote{Contrastive Module} in Fig.~\ref{fig:network_overview} and introduce another set of images from the PASCAL VOC 2007 dataset~\cite{everingham2007pascal} as \enquote{connection modeling} dataset $D_p=\{x_i^p\}_{i=1}^{N_p}$, from which we extract both the camouflaged features and the salient features. With the three datasets (SOD dataset $D_s$, COD dataset $D_c$ and connection modeling dataset $D_p$), our contradicting modeling framework uses the \enquote{Feature Encoder} module to extract both the camouflage feature and the saliency feature.
% , where the output features are feed to the \enquote{Contrastive Module} to further enhance their representativeness.
% with the contrastive mo dule.
% and produce prediction of each task with 
The \enquote{Prediction Decoder} is then used to produce the prediction of each task.
% And the foreground and background corresponding encoded feature of SOD and COD could be obtained separately from features with corresponding prediction maps . 
We further present a \enquote{Contrastive Module} to model the connection of the two tasks with the connection modeling dataset. 

% \Jing{I'm here.}

\noindent\textbf{Feature Encoder:}
% As shown in Fig. \ref{fig:network_overview}, we feed images $X^c$, $X^p$ and $X^s$ in $D_c$, $D_p$ and $D_s$ to the feature encoder module to extract the task-specific feature.
% The goal of t
The \enquote{Feature Encoder}
% is to extract image features for each task, which 
takes the RGB image ($x^s$ or $x^c$) as input to produce task-specific predictions and also serves as the feature extractor for the \enquote{Contrastive Module}. We design both the saliency encoder $E_{\alpha_s}$ and camouflage encoder $E_{\alpha_c}$ with the same backbone network, \eg~the ResNet50~\cite{he2016deep}, where $\alpha_s$ and $\alpha_c$ are the corresponding network parameter sets.
% of each of them respectively. 
% Initially, t
The ResNet50 backbone network has four groups\footnote{We define feature maps of the same spatial size as
% belong to the 
same group.} of convolutional layers of channel size 256, 512, 1024 and 2048 respectively.
% We then add one convolutional layer of kernel size $3\times3$ after the last convolutional layer of each group to obtain feature maps of the same channel size $C=32$.
We then define the output features of both encoders as $F_{\alpha_s}=\{f^s_k\}_{k=1}^4$ and $F_{\alpha_c}=\{f^c_k\}_{k=1}^4$, where $k$ indexes the feature group.
% $f_{k}, k=1,...,4$ is feature map of the $k$-th group.
% $F_{\alpha_s}$ and $F_{\alpha_c}$  are from $E_{\alpha_s}$ and $E_{\alpha_c}$ respectively.

\begin{figure}[tp]
%  \vspace{-5mm}
   \begin{center}
   \begin{tabular}{{c@{ } c@{ } c@{ } c@{ }}}
    {\includegraphics[width=0.235\linewidth,height=0.17\linewidth]{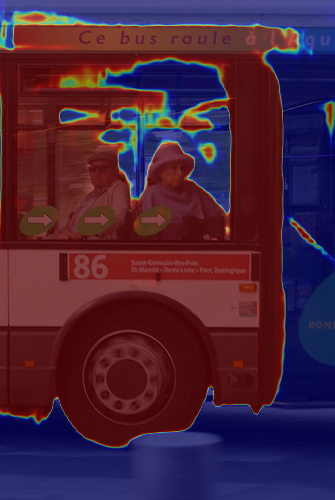}}&
   {\includegraphics[width=0.235\linewidth,height=0.17\linewidth]{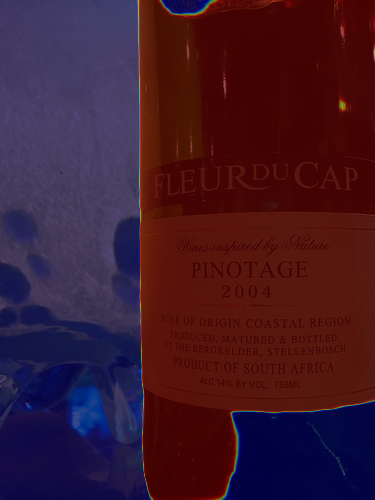}}&
   {\includegraphics[width=0.235\linewidth,height=0.17\linewidth]{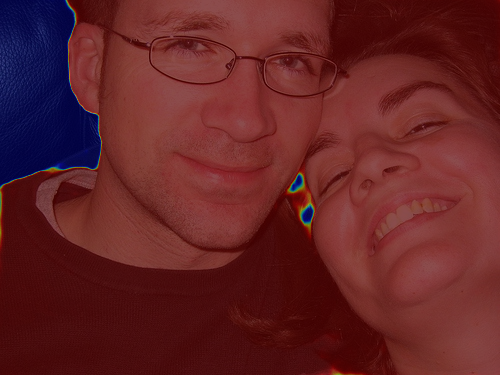}}&
   {\includegraphics[width=0.235\linewidth,height=0.17\linewidth]{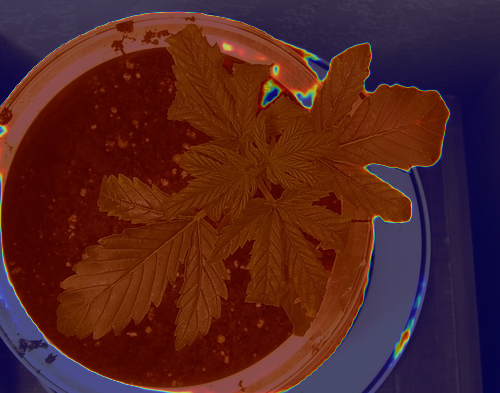}}
    \\
   {\includegraphics[width=0.235\linewidth,height=0.17\linewidth]{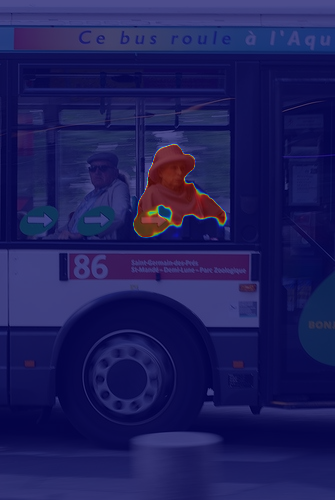}}&
   {\includegraphics[width=0.235\linewidth,height=0.17\linewidth]{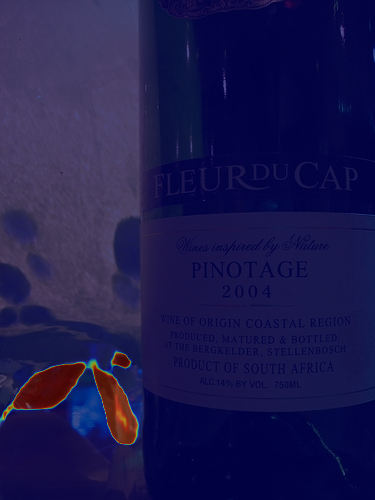}}&
   {\includegraphics[width=0.235\linewidth,height=0.17\linewidth]{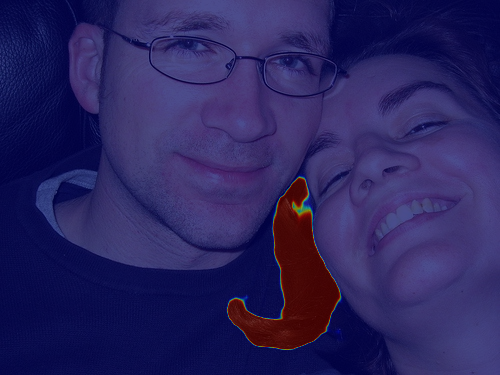}}&
   {\includegraphics[width=0.235\linewidth,height=0.17\linewidth]{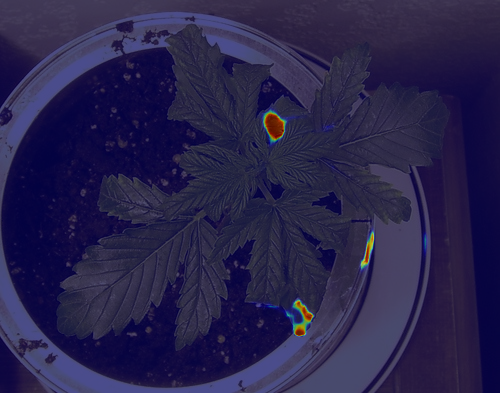}}
\\
   \end{tabular}
   \end{center}
    \caption{Detected regions from the saliency encoder branch ($1^{st}$ row) and the camouflage encoder branch ($2^{nd}$ row).
    % maps of the same image (a) from the saliency encoder (b) and camouflage encoder (c).
    } 
    \label{fig:activation_maps_cod_sod}
\end{figure}

\noindent\textbf{Prediction Decoder:}
% \Jing{I will come back to this part!}\\
% With the \enquote{Feature Encoder} for feature extraction,
% of each task, 
% and \enquote{Contrastive Module}  to constrain feature of each task based on the task \enquote{contradicting} attribute,
% the \enquote{Prediction Decoder} is designed to produce a prediction for each task.
As shown in Fig.~\ref{fig:network_overview}, we design a shared decoder structure for our joint learning framework.
% joint SOD and COD learning. 
% We argue that the different \enquote{Feature Encoder} modules can generate task-specific features.
% % for COD images and SOD images. 
% Then the \enquote{Prediction decoder} module aims to aggregate
% % integrate 
% the task-specific feature
% % with their corresponding lower level feature 
% to produce prediction for each task. 
% Specifically, given task-specific feature $F_{\alpha_s}=\{f^s_1,f^s_2,f^s_3,f^s_4\}$ and $F_{\alpha_c}=\{f^c_1,f^c_2,f^c_3,f^c_4\}$ from the saliency encoder and camouflage encoder respectively, the prediction decoder produces saliency map $G_{\beta}(F_{\alpha_s})$ and camouflage map $G_{\beta}(F_{\alpha_c})$, where $\beta$ is parameter set of the prediction decoder module. 
% Specifically, w
To reduce the computational burden, also to achieve feature with larger receptive field, we first attach a multi-scale dilated convolution~\cite{deeplabv3} of output channel size $C=32$ to each backbone feature to generate the new backbone features $F'_{\alpha_s}=\{f^{cs}_k\}_{k=1}^4$ and $F'_{\alpha_c}=\{f^{cc}_k\}_{k=1}^4$ for each specific task from $F_{\alpha_s}$ and $F_{\alpha_c}$. Then, we adopt the residual attention based feature fusion strategy from~\cite{midas_pami} to achieve high/low level feature aggregation. Specifically, the lower-level features are fed to a residual connection module~\cite{zhang2018image} with two $3\times 3$ convolutional layers, which is then added to the higher level feature. The sum of the high/low level feature is then fed to another residual connection block of the same structure as above to generate the fused feature. We perform the above feature fusion operation until we reach the lowest level feature, \eg~$f^{cc}_1$ or $f^{cs}_1$. To generate the prediction for each task, we design a classifier module, which is composed of three cascaded convolutional layers, where the kernel size of the first two convolutional layers is $3\times 3$, and that of the last convolutional layer is $1\times 1$. 
After generating initial predictions, we used the holistic attention module~\cite{cpd_sal} for feature optimization to obtain further improved predictions, as the final predictions. To simplify the explanation, we only use prediction after the holistic attention module as the decoder output. 
% We used the holistic attention module~\cite{cpd_sal} to optimize the features and obtain final predictions from the initial predictions. For simplicity, we only used one of these predictions to represent the decoder output.
We then define prediction of each task as: $G_{\beta}(F_{\alpha_s})$ for SOD and $G_{\beta}(F_{\alpha_c})$ for COD, where $\beta$ represents the parameter set of the shared prediction decoder.

\noindent\textbf{Contrastive Module:}
% Given the \enquote{contradicting} attribute of the two tasks, t
The \enquote{Contrastive Module} $\mathit{Ctrs}_\theta$ aims to enhance the identity of each task with the feature of other tasks as guidance. Specifically, it takes image $x^p$ from the connection modeling dataset $D_p=\{x_i^p\}_{i=1}^{N_p}$ as input to model the feature correlation of SOD and COD, where $\theta$ is parameter set of the contrastive module.

% Given the saliency encoder $E_{\alpha_s}$ and camouflage encoder $E_{\alpha_c}$, the saliency feature and camouflage feature of images 
For image $x^p$ from the connection modeling dataset, its saliency and camouflage features
% from $D_p$ encoded by SOD encoder and COD encoder 
are $F^p_{\alpha_s}=\{f^{p}_{sk}\}_{k=1}^4$ and $F^p_{\alpha_c}=\{f^{p}_{ck}\}_{k=1}^4$, respectively.
% , and w
With the shared decoder $G_{\beta}$, the prediction map are $G_{\beta}(F^p_{\alpha_s})$ indicating the saliency map and
% for SOD and 
$G_{\beta}(F^p_{\alpha_c})$ as the camouflage map. The contrastive module decides positive/negative pairs based on contrast information, where regions of similar contrast are defined as positive pairs and the different contrast regions are defined as negative pairs.
The intuition behind this is that COD and SOD are both contrast based class-agnostic binary segmentation tasks, making conventional category-aware contrastive learning infeasible to work in this scenario. Considering the goal of building the positive/negative pairs for contrastive learning is to learn representative features via exploring the inherent data correlation, \ie~the category information, we argue the inherent correlation in our scenario is the \enquote{contrast} information. For SOD, the foreground shows higher contrast compared with the background, indicating the different contrast level. For COD, the contrast levels of foreground and background are similar. Thus given the same input image $x^p$, we decide positive/negative pairs based on the contrast information of the activated regions.

In Fig.~\ref{fig:activation_maps_cod_sod}, we show the activation region (the processed predictions) of the same image from both the saliency encoder (first row) and camouflage encoder (second row). Specifically, given same image $x^p$, we compute its camouflage map and saliency map, and highlight the detected foreground region in red. 
% Specifically, we add one global average pooling on top of $F_{\alpha_s}^p$ and $F_{\alpha_c}^p$ respective to obtain a scalar for each channel. Then we feed it to Sigmoid activation function and obtain the normalized weight of each channel. We define the sum of the weighted feature of $F_{\alpha_s}^p$ or $F_{\alpha_c}^p$ as activation map for saliency and camouflage respectively.
Fig.~\ref{fig:activation_maps_cod_sod}  shows that the two encoders focus on different regions of the image, where the saliency encoder pays more attention to the region that stands out from the context.
% human faces that attract attention (Fig.~\ref{fig:activation_maps_cod_sod} (b)),
The camouflage encoder focuses more on the hidden object with similar color or structure as the background,
% dog of similar color with the background (Fig.~\ref{fig:activation_maps_cod_sod} (c)), 
which is consistent with our assumption that these two tasks are contradicting with each other in general.

% for COD respectively.
% To obtain the variability of SOD and COD features, 

\noindent\textit{Feature definition:} Following the conventional practice of contrastive learning, our contrastive module $Ctrs_{\theta}$ maps image features, \ie~$F^p_{\alpha_s}$ and $F^p_{\alpha_c}$ for the connection modeling data $x^p$, to the lower dimensional feature space via
% ontains 
four spectral normed convolutional layers (SNconv)~\cite{miyato2018spectral}, which is proven effective in preserving the geometric distance in the compressed space.
% to map  $F^p_{\alpha_s}$ and $F^p_{\alpha_c}$ to the lower dimensional feature space.
We then compute saliency and camouflage features of the same image:
\begin{equation}
    \label{fore_back_fea}
    \begin{aligned}
    F^p_{sf}&=S(G_{\beta}(F^p_{\alpha_s}),Ctrs_{\theta}(F^p_{\alpha_s})),\\
    F^p_{sb}&=S((1-G_{\beta}(F^p_{\alpha_s})),Ctrs_{\theta}(F^p_{\alpha_s})),\\
    F^p_{\mathit{cf}}&=S(G_{\beta}(F^p_{\alpha_c}),Ctrs_{\theta}(F^p_{\alpha_c})),\\
    F^p_{cb}&=S((1-G_{\beta}(F^p_{\alpha_c})),Ctrs_{\theta}(F^p_{\alpha_c})),
    \end{aligned}
\end{equation}
where $S(\cdot,\cdot)$ computes the region feature via matrix multiplication~\cite{xie2022c2am}, where the feature maps, \ie~$Ctrs_{\theta}(F^p_{\alpha_s})$, are scaled to be the same spatial size as the activation map, \ie~$G_{\beta}(F^p_{\alpha_s})$.
% define region features based on the activation regions~\cite{xie2022c2am}, \ie~model predictions and , as:
% After scaling features and prediction maps to a uniform size, inspired by \cite{xie2022c2am}, the region features can be calculated as:
% \begin{equation}
%     \label{fore_back_cal}
%     % \begin{aligned}
%     Reg(pred^{1\times H\times W},fea^{C\times H\times W}) = pred^{1\times HW} \times fea^{HW\times C},
%     % \end{aligned}
% \end{equation}
% And the foreground and background features of SOD and COD are:
% \begin{equation}
%     \label{fore_back_fea}
%     \begin{aligned}
%     F^p_{sf}&=Reg(G_{\beta}(F^p_{\alpha_s}),Ctrs_{\theta}(F^p_{\alpha_s})),\\
%     F^p_{sb}&=Reg((1-G_{\beta}(F^p_{\alpha_s})),Ctrs_{\theta}(F^p_{\alpha_s})),\\
%     F^p_{\mathit{cf}}&=Reg(G_{\beta}(F^p_{\alpha_c}),Ctrs_{\theta}(F^p_{\alpha_c})),\\
%     F^p_{cb}&=Reg((1-G_{\beta}(F^p_{\alpha_c})),Ctrs_{\theta}(F^p_{\alpha_c})),
%     \end{aligned}
% \end{equation}
$F^p_{sf}\in\mathbb{R}^{1\times C}$ and $F^p_{sb}\in\mathbb{R}^{1\times C}$ in Eq.~\eqref{fore_back_fea} represent the SOD foreground and background features, and $F^p_{\mathit{cf}}$ and $F^p_{cb}$ are the COD foreground and background features, respectively. 

\noindent\textit{Positive/negative pair construction:} According to our previous discussion, we define three sets of positive pairs based on contrast similarity:
% for the above features: 
(1) The SOD background feature and COD background feature of the same image should be highly similar, indicating similar contrast information; 
(2) Due to the nature of the camouflaged object, the foreground and the background features of COD are similar as well as camouflaged object shares similar contrast with the background; (3) Similarly, the
% Based on the above two points, 
COD foreground feature and SOD background feature are also similar in contrast.
% positive pair.
On the other hand, the negative pair consists of SOD foreground feature and background feature. 

\noindent\textit{Contrastive loss:} Given the positive/negative pairs, we follow~\cite{oord2018representation} and define the contrastive loss as:
% To encourage contradicting between saliency and camouflage, we follow \cite{oord2018representation} and define the task feature difference loss as:
\begin{equation}
    \label{latent_loss}
   \mathcal{L}_{ctrs}=-\log\frac {\textstyle \sum_{pos}} {{\textstyle \sum_{pos}}+\exp(c(F^p_{sf},F^p_{sb}))},
\end{equation}
where $c(\cdot )$ measures the cosine similarity of the normalized vectors.
% , which is defined as the cosine similarity in this paper.
% defined as $S(A,B)=AB^\top$, and 
$ \textstyle \sum_{pos}$ represents the similarity of positive pairs, which is defined as: 
\begin{equation}
    {\textstyle \sum_{pos}} =  \exp(c(F^p_{cf},F^p_{cb}))+\exp(c(F^p_{sb},F^p_{cb}))+\exp(c(F^p_{sb},F^p_{cf})).
\end{equation}

\begin{figure}[tp]
%  \vspace{-5mm}
   \begin{center}
   \begin{tabular}{{c@{ } c@{ } c@{ } c@{ }}}
{\includegraphics[width=0.235\linewidth,height=0.1434\linewidth]{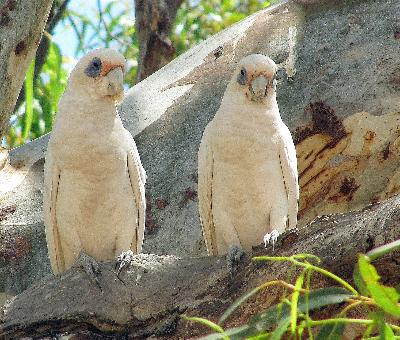}}&
{\includegraphics[width=0.235\linewidth,height=0.1434\linewidth]{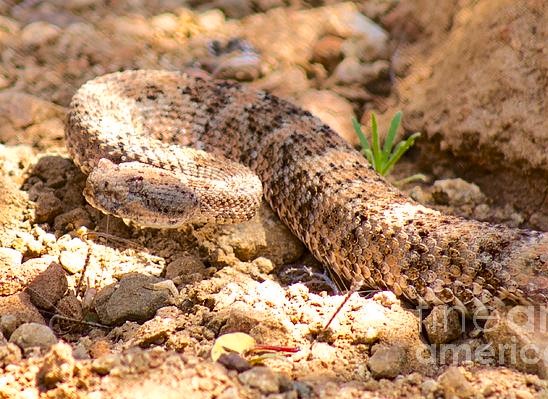}}&
{\includegraphics[width=0.235\linewidth,height=0.1434\linewidth]{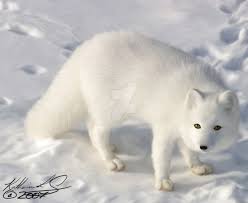}}&
{\includegraphics[width=0.235\linewidth,height=0.1434\linewidth]{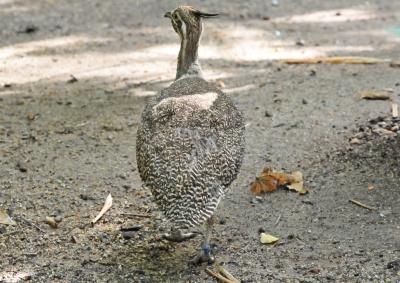}}\\
{\includegraphics[width=0.235\linewidth,height=0.1434\linewidth]{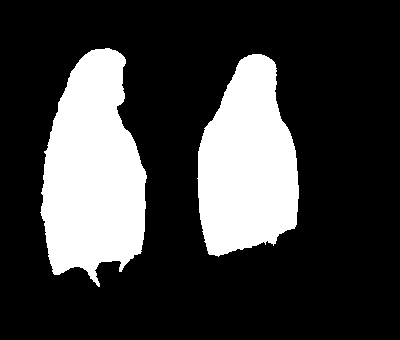}}&
{\includegraphics[width=0.235\linewidth,height=0.1434\linewidth]{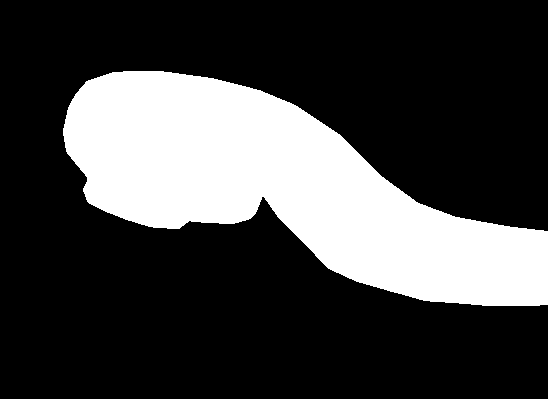}}&
{\includegraphics[width=0.235\linewidth,height=0.1434\linewidth]{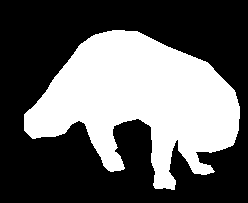}}&
{\includegraphics[width=0.235\linewidth,height=0.1434\linewidth]{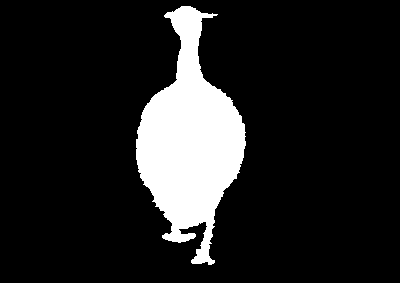}}\\

   \end{tabular}
   \end{center}
    \caption{Selected easy samples from the COD training dataset.} 
    \vspace{-2mm}
    \label{fig:selected_easy_samples_from_cod}
\end{figure}

\subsubsection{Data Interaction}
\label{subsub:data_interaction}
 In Section \ref{sub:contradicting_modeling}, we discuss the contradicting modeling strategy to model the two tasks from the model correlation perspective. In this section, we further explore the task relationships from dataset perspective, and introduce data interaction as data augmentation.

% \Jing{maybe an more efficient way for data interaction. currently, we select easy samples from cod, and add them to sod dataset. what about an automatic sample selection solution?}
\noindent\textit{Sample selection principle:} As shown in Fig.~\ref{fig:relationship_sod_cod}, saliency and camouflage are two properties that can transfer from each other. We find that there exist samples in the COD dataset that are both salient and camouflaged. We argue that those samples can be treated as hard samples for SOD to achieve robust learning. The main requirement is that the activation of those samples for SOD and COD should be similar.
In other words, the predictions of the selected images for both tasks need to be similar.
% consistently predicted on the SOD and COD models. 
To select those samples from the COD dataset, we resort to weighted Mean Absolute Error ($\mathit{wMAE}$), and select samples in the COD dataset~\cite{fan2020camouflaged} which achieve the smallest $\mathit{wMAE}$ by testing it using a trained SOD model.

The weighted mean absolute error $\mathit{wMAE}$ is defined as :
% \begin{equation}
% \label{wmae}
%    \mathit{wMAE}=\frac{\sum_{m}^{x^c} \left ( p^c(m)-y^c(m) \right ) }{\sum_{m}^{x^c} y^c(m)},
% \end{equation}
\begin{equation}
    \mathit{wMAE} = \frac{\sum_{u=1}^W\sum_{v=1}^H |y^{u, v} - p^{u,v} |}{\sum_{u=1}^W\sum_{v=1}^H y^{u, v}},
    \label{wmae}
\end{equation}
where
% $W$ and $H$ indicate size of the ground-truth and image. 
${u,v}$ is the pixel index, $p$ represents the model prediction, $y$ is the corresponding ground-truth, and $W$ and $H$ indicate size of $y$.
% As we adjust the images to a consistent size during the training phase, normalization of image size was omitted in $\mathit{wMAE}$. 
Compared with mean absolute error, $\mathit{wMAE}$ avoids the biased selection caused by different sizes of the foreground object(s).
% filters out samples with poor predictions of smaller MAE due to small object and keeps samples with good predictions and large MAEs due to large object.
% Specifically, f

\noindent\textit{Data interaction:} For the COD training dataset $D_c =\{x_i^c, y_i^c\}_{i=1}^{N_c}$ and
% , where $i$ indexes the images, and $N_c$ is the size of camouflaged object detection training set. We defined 
the trained SOD model $M_{\theta_s}$,
% Then 
% we obtain saliency prediction of the images in $D_c$ as $P^c_s=M_{\theta_s}(\{x^c\})=\{p^c_i\}_{i=1}^{N_c}$, where $p_i^c$ is the saliency prediction of the COD training dataset. We assume that easy samples for COD can be treated as hard samples for SOD as shown in Fig.~\ref{fig:relationship_sod_cod}. Then we select $M=403$ samples $D_c^M$ with the smallest $\mathit{wMAE}$ in $D_c$ via \eqref{wmae}, and replace it with $M=403$ randomly selected samples in our SOD training dataset~\cite{wang2017learning} as a data augmentation technique. Note that we replace the samples instead of adding the samples directly to SOD training dataset to keep the training dataset size unchanged.
we obtain saliency prediction of the images in $D_c$ as $P^c_s=M_{\theta_s}(\{x^c\})=\{p^c_i\}_{i=1}^{N_c}$, where $p_i^c$ is the saliency prediction of the COD training dataset. We assume that easy samples for COD can be treated as hard samples for SOD as shown in Fig.~\ref{fig:relationship_sod_cod}. Then we select $M=403$ samples $D_c^M$ with the smallest $\mathit{wMAE}$ in $D_c$ via Eq.~\eqref{wmae}, and add in our SOD training dataset~\cite{wang2017learning} as a data augmentation technique.
% As the model learns the data distribution implicitly under joint training, we consider it fair to explicitly perform about $3.8\%$ of the data interactions.
We show the selected samples in Fig.~\ref{fig:selected_easy_samples_from_cod}, which clearly illustrates the partially positive connection of the two tasks at the dataset level.
% consistency of clear saliency in those easy camouflaged objects.

% \noindent\textbf{Random-sampling based Foreground Cropping:}
\subsubsection{Foreground Cropping as Data Augmentation:}
% \label{random_sampling_cropping}
Considering the real-life scenarios, camouflaged objects can appear in different sizes,
% and localize in different regions of the image, 
we introduce foreground cropping to achieve context-aware data augmentation.
% for
% % avoid the \enquote{center bias} and \enquote{size bias} of 
% the camouflage datasets. 
Note that we only perform foreground cropping for COD as the prediction of SOD is relatively stable with different sizes of the foreground object(s).
% that the foreground cropping strategy is especially effective for COD than for SOD, indicating the closer dependency of COD with the foreground context.
% most of the  of the training dataset is not effective  perform random sampling for COD as the above two types of biases are acceptable for SOD while not suitable for COD.
% As shown in Fig.~\ref{fig:foreground_cropping_performance}, COD is related to the content of the picture. And in contrast to SOD, which focuses more on global comparisons, COD tends to distinguish between camouflaged targets and subtle changes in the local environment. Therefore, we propose to use 
Specifically, we first define the largest bounding box region that covers all the camouflaged objects as the compact cropping (\enquote{CCrop}). Then, we obtain the median cropping (\enquote{MCrop}) and loose cropping (\enquote{LCrop}) by randomly extending $0-80$ and $0-150$ pixels respectively randomly outward along the compact bounding box. We perform cropping on the raw images and resize the cropped image back to the pre-defined training image size for training.

\subsection{Uncertainty-aware Learning}
\label{sub:adversarial_learning}

In Section \ref{sub:task_analysis}, we discussed that both SOD and COD have inherent uncertainty, where the \enquote{subjective nature} of SOD poses serious \enquote{model uncertainty} \cite{what_uncertainty_kendall} for SOD and difficulty of labeling introduces \enquote{data uncertainty} \cite{what_uncertainty_kendall} for COD.
% \subsection{Uncertainty Estimation}
% \Jing{ground truth for the confidence estimation module is 0 or 1. what if we define it's ground truth as dynamic residual of prediction: y*(1-p)+(1-y)*p? Meanwhile, how about define output of confidence estimation module as hard sample identification module, and put more emphasis on those hard samples in the final loss function similar to focal loss?}
% The objective of machine learning methods is to minimize the expected loss function:
% \begin{equation}
% \begin{aligned}
% \min_\theta &\mathbb{E}_{x,y}[\mathcal{L}(f(x;\theta),y)]=\int\mathcal{L}(f(x;\theta),y)d p(x,y)\\
% &\approx\frac{1}{N}\sum_{i=1}^N\mathcal{L}(f(x_i;\theta),y_i), \quad (x_i,y_i)\sim p(x,y),
% \end{aligned}
% \label{uncertainty_formulation}
% \end{equation}
% where $x,y$ are input and output variables, $\theta$ is the learned model parameter set, $p(x,y)$ is the joint data distribution and $x_i,y_i$ represent sample from the joint data distribution $p(x,y)$.
As shown in Fig.~\ref{fig:task_uncertainty}, for the SOD dataset, the uncertainty comes from the ambiguity of saliency.
% \AX{(majority voting process)}
% \sout{(the representativeness of the selected saliency training dataset $D_s$ as the joint data distribution $p(x,y)$, or the generalization ability of the learned parameter set $\theta$).}
For the COD dataset, the uncertainty mainly comes from the difficulty of labeling (the accuracy of $y_i$).
% \eg the ball in the orange rectangle (a) can be defined as salient, but it's background in (b). The orange region in (c) belongs to the camouflaged object, while it's too similar to the background, making it very difficult to create the accurate annotation.
% \AX{In addition, the generalization ability of the learned parameter set $\theta$ also introduces uncertainty.}
To model the uncertainty of both tasks for reliable model generation, we introduce an uncertainty-aware adversarial training strategy to model the task-specific uncertainty in our joint learning framework.
\noindent\textit{Adversarial learning framework:} Following the conventional practice of generative adversarial network (GAN) \cite{gan_raw}, we design a fully convolutional discriminator network to evaluate confidence of the predictions.
% from the \enquote{Prediction Decoder} module.
The fully convolutional discriminator network $D_\gamma$ consists of five SNconv layers \cite{miyato2018spectral} of kernel size $3\times 3$.
% as shown in Table~\ref{tab:discriminator_gan}.
As a conditional generation task, the fully convolutional discriminator takes the prediction/ground truth and the conditional variable, \ie~the RGB image, as input, and produces a one-channel confidence map, where $\gamma$ is the network parameter set. Note that we have batch normalization and leaky relu layers after the first four convolutional layers. $D_\gamma$ aims to distinguish areas of uncertainty, which produce all-zero output with ground truth $y$ as input, and produce $|p-y|$ output with prediction map $p$ as input.
% model predictions and the ground truth maps, which produce all-zero output with prediction as input, and all-one matrix with ground truth as input. 
% Note that, as a conditional task with the input image as a conditional variable, the input of the discriminator is the concatenation of the input image and the corresponding prediction/ground truth map.

% \begin{table}[!ht]
%   \centering
%   \footnotesize
%   \renewcommand{\arraystretch}{1.1}
%   \renewcommand{\tabcolsep}{1.5mm}
%   \caption{Network structure of the discriminator network.}
%   \begin{tabular}{c|c|c|c|c}
%   \hline
% %   \toprule
%   \multicolumn{1}{c|}{Input Channel}&\multicolumn{1}{c|}{Output Channel}&\multicolumn{1}{c|}{kernel size}&\multicolumn{1}{c|}{Stride}&\multicolumn{1}{c}{Padding} \\
%   \hline
%   4 & 64 & 3 & 2 & 1 \\ 
%   64 & 64 & 3 & 1 & 1 \\
%   64 & 64 & 3 & 2 & 1 \\
%   64 & 64 & 3 & 1 & 1 \\
%   64 & 1 & 3 & 2 & 1 \\
%   \hline
%   \end{tabular}
%   \label{tab:discriminator_gan}
% %   \vspace{-2mm}
% \end{table}

In our case, the fully convolutional discriminator aims to discover the hard (or uncertain) regions of the input image.
% \AX{\sout{ For SOD and COD,} }
We use the same structure of discriminators
% in Table~\ref{tab:discriminator_gan}
with parameter sets $\gamma_s$ and $\gamma_c$ for SOD and COD respectively, to identify the two types of challenging regions, \ie~the \enquote{subjective area} for SOD, and the \enquote{ambiguous regions} for COD.
\noindent\textit{Uncertainty-aware learning:}
% \Jing{I will come back to this part!}\\
% \AX{the code is like hard mining}
For the prediction decoder module, we first have the task-specific loss function to learn each task. Specifically, we adopt the structure-aware loss function~\cite{wei2020f3net} for both SOD and COD, and define the loss function as:
% The structure loss is weighted by BCE Loss and IoU loss to both predictions:
\begin{equation}
\label{structure_loss}
   \mathcal{L}_{str}(p,y)=\omega*\mathcal{L}_{ce}(p,y)+\mathcal{L}_{iou}^\omega(p,y), 
\end{equation}
where $\omega$ is the edge-aware weight, which is defined as $\omega=1+5*\left | (avg\_pool(y)-y)\right |$, $y$ is task-specific ground truth, $\mathcal{L}_{ce}$ is the binary cross-entropy loss,  $\mathcal{L}_{iou}^\omega$ is the weighted boundary-IOU loss~\cite{nldf_sal}.
% which is defined as:
% \begin{equation}
% \label{iou_loss}
%    \mathcal{L}_{iou}= 1-\frac{\omega*inter+1}{\omega*union - \omega*inter+1},
% \end{equation}
% where $inter$ is the intersection of prediction and ground truth, and $union$ is their union.
In this way, the task specific loss functions $\mathcal{L}_{str}^s$ and $\mathcal{L}_{str}^c$ for SOD and COD are defined as:
% And we have the structure-aware loss function for SOD and COD as:
% \begin{equation}
% \begin{aligned}
% \label{structure_loss_sod_cod}
%   \mathcal{L}_{str}^s=0.5*[{L}_{str}(G_{init}(F_{\alpha_s}),Y^s)+{L}_{str}(G_{\beta}(F_{\alpha_s}),Y^s)], \\
%   \mathcal{L}_{str}^c=0.5*[{L}_{str}(G_{init}(F_{\alpha_c}),Y^c)+{L}_{str}(G_{\beta}(F_{\alpha_c}),Y^c)], 
%   \end{aligned}
% \end{equation}
% \begin{equation}
% \begin{aligned}
% \label{structure_loss_sod_cod}
%   \mathcal{L}_{str}^s=\mathcal{L}_{str}(G_{\beta}(F_{\alpha_s}),y^s), \\
%   \mathcal{L}_{str}^c=\mathcal{L}_{str}(G_{\beta}(F_{\alpha_c}),y^c).
%   \end{aligned}
% \end{equation}
\begin{gather}
\label{structure_loss_sod}
  \mathcal{L}_{str}^s=\mathcal{L}_{str}(G_{\beta}(F_{\alpha_s}),y^s), \\
  \label{structure_loss_cod}
   \mathcal{L}_{str}^c=\mathcal{L}_{str}(G_{\beta}(F_{\alpha_c}),y^c),
\end{gather}
% where $\mathcal{L}_{str}^s$ and $\mathcal{L}_{str}^c$ are task-specific loss function for SOD and COD respectively.
% Similarly, we have the structure-aware loss function for COD as:
% \begin{equation}
% \label{structure_loss_cod}
%   \mathcal{L}_{str}^c=0.5*[{L}_{str}(G_{init}(F_{\alpha_c}),Y^c)+{L}_{str}(G_{\beta}(F_{\alpha_c}),Y^c)]. 
% \end{equation}

% To achieve adversarial learning, following \cite{hung2018adversarial}, we assume that the clean ground-truth have no areas of uncertainty,\eg~$\mathbf{0}$. And the uncertainty of prediction map is a residual of prediction with ground-truth, defined as $|Pred-y|$. We further introduce adversarial loss function to both SOD and COD predictors, which is defined as cross-entropy loss between image based network prediction and the pre-defined uncertainty indicator, aiming to fool the discriminators that the prediction of SOD or COD is the actual ground truth. The adversarial loss functions ($\mathcal{L}_{adv}^s$ and $\mathcal{L}_{adv}^c$) for SOD and COD, respectively, are defined as:

To achieve adversarial learning, following \cite{hung2018adversarial}, we further introduce adversarial loss function to both SOD and COD predictors, which is defined as a consistency loss between discriminators prediction of prediction map and discriminators prediction of ground-truth, aiming to fool the discriminators that the prediction of SOD or COD is the actual ground truth.
The adversarial loss functions ($\mathcal{L}_{adv}^s$ and $\mathcal{L}_{adv}^c$) for SOD and COD, respectively, are defined as:
% \begin{equation}
% \begin{aligned}
%     \label{adv_pred_sod_cod}
%     \mathcal{L}_{adv}^s = 0.5*(& \mathcal{L}_{ce}(D_{\gamma_{s}}(x^s,G_{init}(F_{\alpha_s})), \mathbf{1})\\
%     &+\mathcal{L}_{ce}(D_\gamma^f(x^s,G_\beta(F_{\alpha_s})), \mathbf{1})),\\
%     \mathcal{L}_{adv}^c = 0.5*(& \mathcal{L}_{ce}(D_{\gamma_{c}}(x^c,G_{init}(F_{\alpha_c})), \mathbf{1})\\
%     &+\mathcal{L}_{ce}(D_\gamma^f(x^c,G_\beta(F_{\alpha_c})), \mathbf{1})),\\
%     \end{aligned}
% \end{equation}
% \begin{equation}
% \begin{aligned}
%     \label{adv_pred_sod_cod}
%     \mathcal{L}_{adv}^s &=0.5*( \mathcal{L}_{ce}(D_{\gamma_{s}}(x^s,G_{init}(F_{\alpha_s})), D_{\gamma_{s}}(x^s,y^s))\\
%     &+ \mathcal{L}_{ce}(D_{\gamma_{s}}(x^s,G_\beta(F_{\alpha_s})), D_{\gamma_{s}}(x^s,y^s))),\\
%     \mathcal{L}_{adv}^c &=0.5*(\mathcal{L}_{ce}(D_{\gamma_{c}}(x^c,G_{init}(F_{\alpha_c})), D_{\gamma_{c}}(x^c,y^c))\\
%     &+ \mathcal{L}_{ce}(D_{\gamma_{c}}(x^c,G_\beta(F_{\alpha_c})), D_{\gamma_{c}}(x^c,y^c))),\\
% \end{aligned}
% \end{equation}
\begin{gather}
    \label{adv_pred_sod}
    \mathcal{L}_{adv}^s = \mathcal{L}_{ce}(D_{\gamma_{s}}(x^s,G_\beta(F_{\alpha_s})), D_{\gamma_{s}}(x^s,y^s)),\\
    \label{adv_pred_cod}
    \mathcal{L}_{adv}^c =\mathcal{L}_{ce}(D_{\gamma_{c}}(x^c,G_\beta(F_{\alpha_c})), D_{\gamma_{c}}(x^c,y^c)),
\end{gather}

% and
% \begin{equation}
%     \label{adv_pred_cod}
%     \mathcal{L}_{adv}^c = \mathcal{L}_{ce}(D_\gamma^f(G_\beta(F_{\alpha_c})), \mathbf{1}),
% \end{equation}
% respectvely for each task, 
% where $\mathbf{1}$ is an all-one matrix. 
% Note that, we have two copies of fully convolutional discriminator $D_{\gamma_{s}}$ and $D_{\gamma_{c}}$ for SOD and COD respectively. They share the same network structure as shown in Table \ref{tab:discriminator_gan}, but not the weights.
% In this way, the discriminator takes model prediction as input, and tries to recognize it as real ground truth.

\begin{figure*}[h]
%  \vspace{-5mm}
   \begin{center}
   \begin{tabular}{{c@{ } c@{ } c@{ } c@{ } c@{ } c@{ } c@{ } c@{ }}}
    {\includegraphics[width=0.135\linewidth,height=0.0824\linewidth]{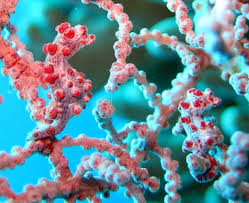}}&
    {\includegraphics[width=0.135\linewidth,height=0.0824\linewidth]{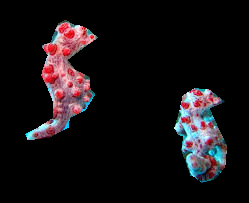}}&
    {\includegraphics[width=0.135\linewidth,height=0.0824\linewidth]{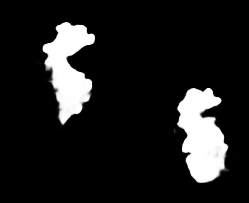}}&
    {\includegraphics[width=0.135\linewidth,height=0.0824\linewidth]{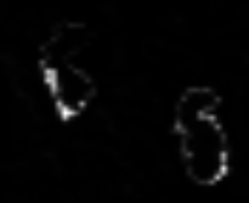}}&
    {\includegraphics[width=0.135\linewidth,height=0.0824\linewidth]{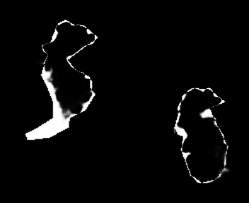}}&
    {\includegraphics[width=0.135\linewidth,height=0.0824\linewidth]{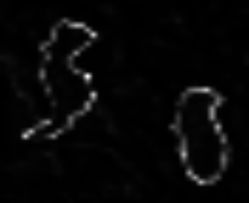}}&
    {\includegraphics[width=0.135\linewidth,height=0.0824\linewidth]{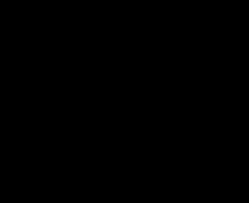}}\\
    
    {\includegraphics[width=0.135\linewidth,height=0.0824\linewidth]{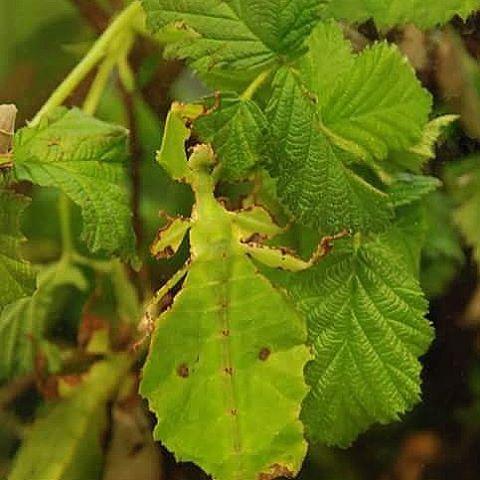}}&
    {\includegraphics[width=0.135\linewidth,height=0.0824\linewidth]{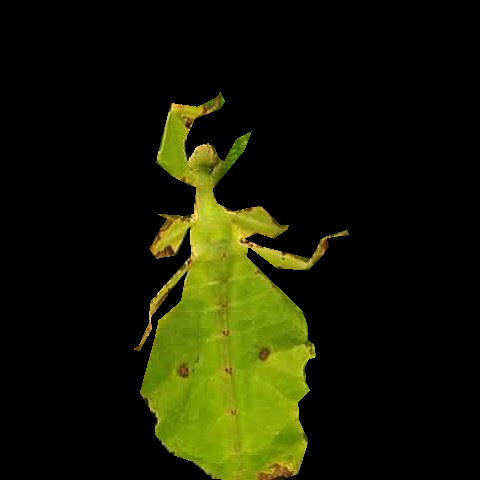}}&
    {\includegraphics[width=0.135\linewidth,height=0.0824\linewidth]{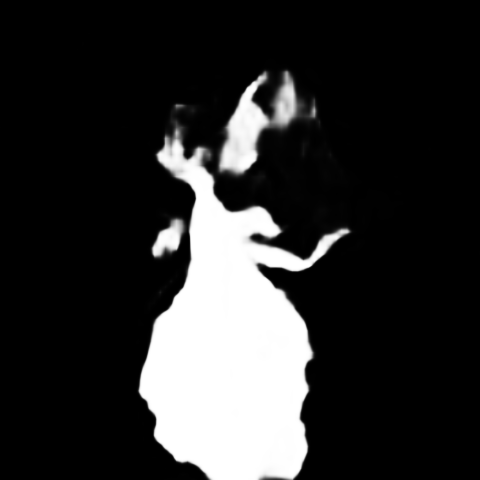}}&
    {\includegraphics[width=0.135\linewidth,height=0.0824\linewidth]{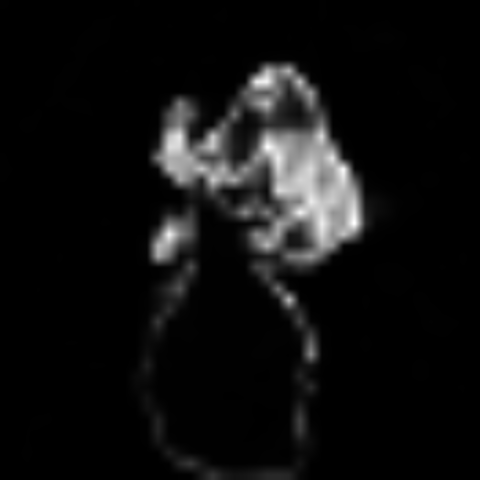}}&
    {\includegraphics[width=0.135\linewidth,height=0.0824\linewidth]{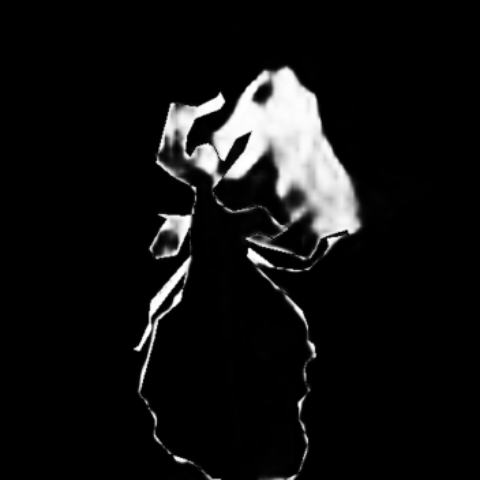}}&
    {\includegraphics[width=0.135\linewidth,height=0.0824\linewidth]{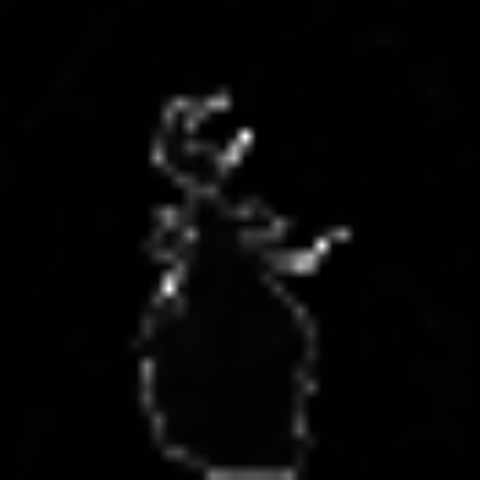}}&
    {\includegraphics[width=0.135\linewidth,height=0.0824\linewidth]{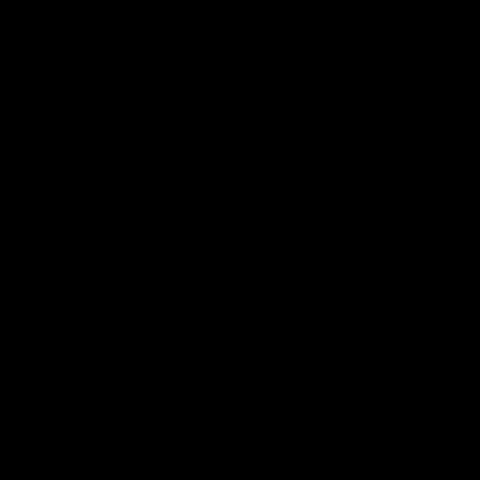}}\\
    
    {\includegraphics[width=0.135\linewidth,height=0.0824\linewidth]{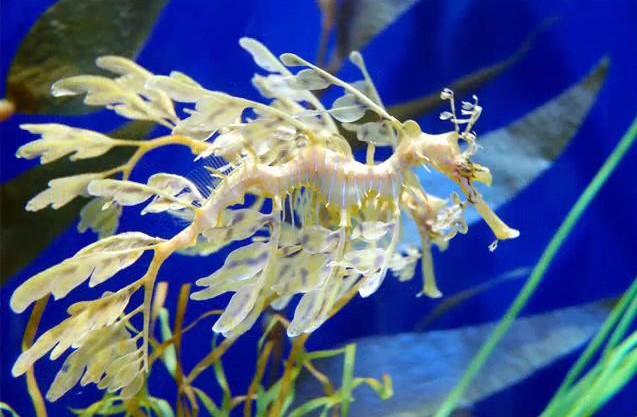}}&
    {\includegraphics[width=0.135\linewidth,height=0.0824\linewidth]{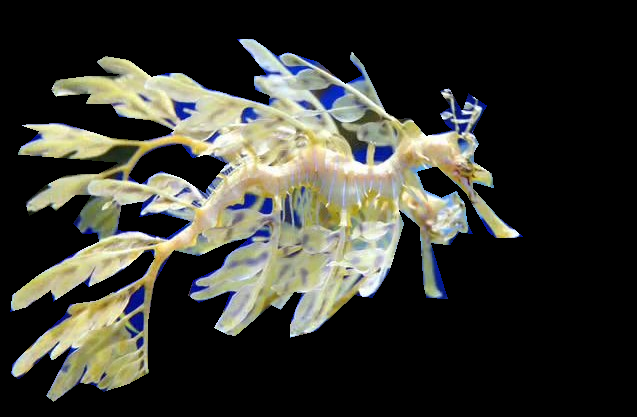}}&
    {\includegraphics[width=0.135\linewidth,height=0.0824\linewidth]{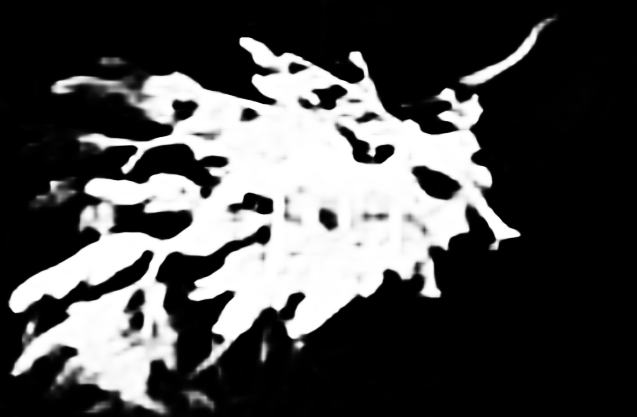}}&
    {\includegraphics[width=0.135\linewidth,height=0.0824\linewidth]{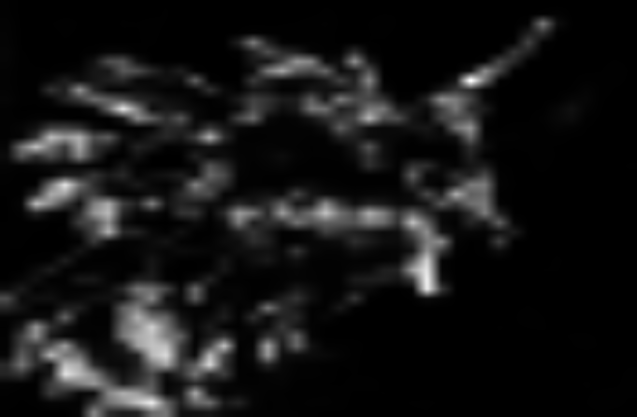}}&
    {\includegraphics[width=0.135\linewidth,height=0.0824\linewidth]{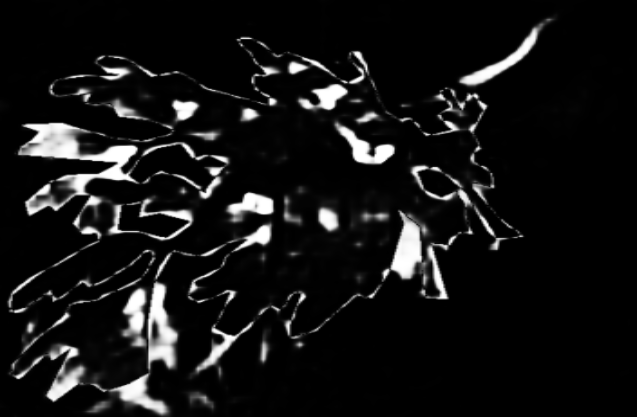}}&
    {\includegraphics[width=0.135\linewidth,height=0.0824\linewidth]{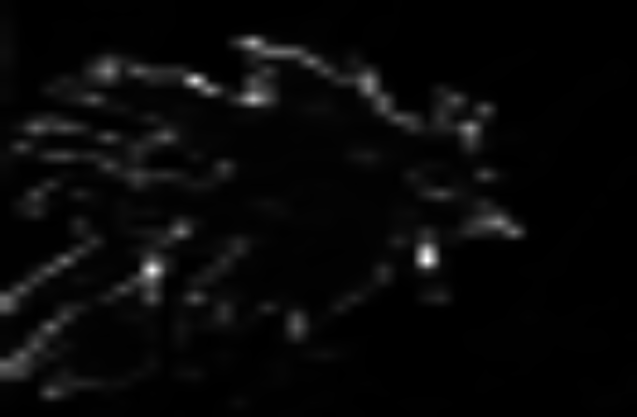}}&
    {\includegraphics[width=0.135\linewidth,height=0.0824\linewidth]{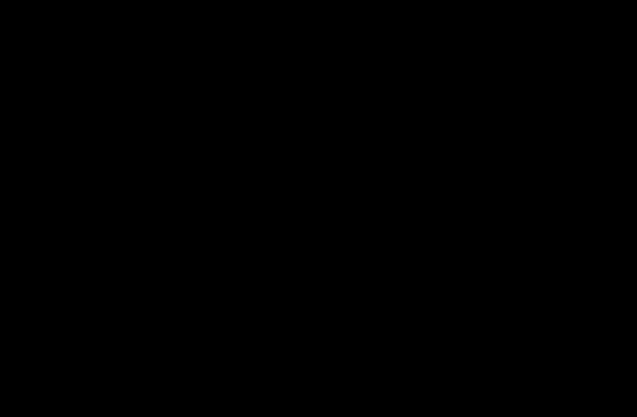}}\\

    \footnotesize{Image} & \footnotesize{GT} & \footnotesize{Pred} &\footnotesize{$D_{\gamma_{s}}(x^s,G_\beta(F_{\alpha_s}))$} &\footnotesize{$|G_\beta(F_{\alpha_s})-y^s|$} &\footnotesize{$D_{\gamma_{c}}(x^c,y^c)$} &\footnotesize{$\textbf{0}$}\\
    % \footnotesize{Image} & \footnotesize{GT} & \footnotesize{Pred} &\footnotesize{$\mathrm{ Dis\_Pred_{pred}}$} &\footnotesize{$\mathrm{ Dis\_GT_{pred}}$} &\footnotesize{$\mathrm{ Dis\_Pred_{gt}}$} &\footnotesize{$\mathrm{ Dis\_GT_{gt}}$}\\
   \end{tabular}
   \end{center}
     \caption{Illumination of the ground truth and predictions of the discriminator, where \enquote{Image}, \enquote{GT} and \enquote{Pred} are the RGB image, the camouflage ground truth map, and the camouflage prediction. The definition of other items is defined in Eq.~\eqref{dis_sod} and Eq.~\eqref{dis_cod}. The discriminator is designed to highlight the error predictions, leading to ground-truth aware uncertainty generation at the testing stage.}
     
     % adversarial learning framework. The predictions of the main network and the ground truth are input to the discriminator separately (concatenate with image). GT represents the ground truth of the Image. Pred represents the prediction of the main network. $\mathrm{ Dis\_Pred_{pred}}$ represents the uncertainty prediction obtained by inputting Pred to the discriminator. $\mathrm{ Dis\_GT_{pred}}$  represents the true value image corresponding to the uncertainty when inputting Pred to the discriminator(the residual image between Pred and the GT).
     % $\mathrm{ Dis\_Pred_{gt}}$ represents the uncertainty prediction obtained by inputting GT to the discriminator, and $\mathrm{ Dis\_GT_{gt}}$ represents the ground truth image corresponding to the uncertainty when inputting the GT to the discriminator (all-zero map).}
     \label{fig:ground_truth_adversarial_learning}
\end{figure*}  

Both the task specific loss in Eq.~\eqref{structure_loss_sod}, Eq.~\eqref{structure_loss_cod} and the adversarial loss in Eq.~\eqref{adv_pred_sod}, Eq.~\eqref{adv_pred_cod} are used to update the task-specific network (the generator). To update the discriminator, following the conventional GAN,
% For the confidence estimation module, similar to the typical definition of discriminator in GAN~\cite{gan_raw}, 
we want it to distinguish areas of uncertainty clearly. Due to the inherent uncertainty that cannot be directly described, the uncertainty in inputting the ground truth cannot be accurately represented. However, because the correctly annotated regions are dominant in the complete dataset, we believe that the network can perceive the areas that are difficult to learn. The adversarial learning mechanism makes it difficult for the discriminator to distinguish between predicted and ground truth maps, and it can differentiate between noisy ground truth images and areas where RGB images cannot be aligned. Therefore, the output of the discriminator when inputting ground truth is defined as an all-zero map. Additionally, it produces a residual output for the prediction map. The outputs corresponding to different inputs of the discriminator are shown in Fig.~\ref{fig:ground_truth_adversarial_learning}. Then, the discriminators ($D_{\gamma_{s}}$ and $D_{\gamma_{c}}$) are updated via:
% adversarial loss for the confidence estimation module of the SOD task is defined as:
\begin{gather}
\label{dis_sod}
    \mathcal{L}_{dis}^s=\mathcal{L}_{ce}(D_{\gamma_{s}}(x^s,G_\beta(F_{\alpha_s})), |G_\beta(F_{\alpha_s})-y^s|),\\
    + \mathcal{L}_{ce}(D_{\gamma_{s}}(x^s,y^s),\mathbf{0}), \notag \\
\label{dis_cod}
    \mathcal{L}_{dis}^c=\mathcal{L}_{ce}(D_{\gamma_{c}}(x^c,G_\beta(F_{\alpha_c})), |G_\beta(F_{\alpha_c})-y^c|),\\
    + \mathcal{L}_{ce}(D_{\gamma_{c}}(x^c,y^c),\mathbf{0}), \notag
\end{gather}
Note that the two discriminators are updated separately.

\subsection{Objective Function}
As a joint confidence-aware adversarial learning framework, we further introduce the objective functions in detail for better understanding of our learning pipeline.

Firstly, given a batch of images from the SOD training dataset $x^s$, we define the confidence-aware loss with contrastive modeling for the generator as:
\begin{equation}
    \label{generator_sod_conf}
    \mathcal{L}^s = \mathcal{L}_{str}^s +\lambda_{adv}*\mathcal{L}_{adv}^s+\lambda_{ctrs}*\mathcal{L}_{ctrs},  % \lambda_{adv1}*\mathcal{L}_{ctrs} + 
\end{equation}
% \begin{equation}
%     \label{generator_sod_conf}
%     \mathcal{L}^s = \mathcal{L}_{str}^s + \mathcal{L}_c^s + \mathcal{L}_{latent} + \lambda_{adv}*\mathcal{L}_{avd}^s,
% \end{equation}
where $\mathcal{L}_{str}^s$ is the task specific loss, defined in Eq.~\eqref{structure_loss_sod}, $\mathcal{L}_{avd}^s$ is the adversarial loss in Eq.~\eqref{adv_pred_sod}, and $\mathcal{L}_{ctrs}$ is the contrative loss in Eq.~\eqref{latent_loss}. The parameters $\lambda_{adv}=1,\lambda_{ctrs}=0.1$ are used to balance the contribution of adversarial loss/contrastive loss for robust training.
% , and similarly, $\lambda_{ctrs}=1$ is used to   %

Similarly, for image batch $x^c$ from the COD training dataset, the confidence-aware loss with contrastive modeling for the generator is defined as:
\begin{equation}
    \label{generator_cod_conf}
    \mathcal{L}^c = \mathcal{L}_{str}^c  +  \lambda_{adv}*\mathcal{L}_{adv}^c+\lambda_{ctrs}*\mathcal{L}_{ctrs}.
\end{equation}
% where $\lambda_{adv}=1$
% Then, t
The discriminators are optimized separately, where $D_{\gamma_{s}}$ and $D_{\gamma_{c}}$ are updated via Eq.~\eqref{dis_sod} and Eq.~\eqref{dis_cod}.
% $\mathcal{L}_{dis}^s$ and $\mathcal{L}_{dis}^c$ respectively in Eq.~\eqref{dis_sod_cod}.
Note that, we only introduce contrastive learning to our joint-task learning framework after every 5 steps, which is proven more effective in practice.
% And every $5$ steps, We model the contrasting properties of the two tasks by the contradicting modeling loss $\mathcal{L}_ctrs$ , defined in Eq.~\eqref{latent_loss}, and we apply a weight of value $0.1$ to $\mathcal{L}_ctrs$. 
We show the training pipeline of our framework in Algorithm \ref{our_alg} for better understanding of the implementation details.

\begin{algorithm}[!t]
\small
\caption{Uncertainty-aware Joint Salient Object Detection and Camouflaged Object Detection}
\textbf{Input}: Training image sets: $D_s = \{x^s,y^s\}$, $D_c = \{x^c,y^c\}$ and $D_p = \{x^p\}$; Maximal number of learning iterations $T$. \\
\textbf{Output}: 
$\alpha_s$, $\alpha_c$ for feature encoder, $\theta$ for contrastive module, $\beta$ for prediction decoder, and $\gamma_s$ and $\gamma_c$ for the two fully convolutional discriminators; 
\begin{algorithmic}[1]
\State Initialize $\alpha_s$, $\alpha_c$ with ResNet50~\cite{he2016deep}, and $\theta$, $\beta$, $\gamma_s$, $\gamma_c$ by default.
\For{$t \leftarrow  1$ to $T$} 

% \If{$t \% 5 =0$}
% $\lambda_{ctrs}=0.1$
% Otherwise,
% $\lambda_{ctrs}=0$ in Eq.~\ref{generator_sod_conf} and Eq.~\ref{generator_cod_conf}.
% % \State Compute $\mathcal{L}_{ctrs}$ for contradicting modeling with $D_p$ and update $\alpha_s$, $\alpha_c$, $\theta$  via loss function in Eq.~\eqref{latent_loss};
% \EndIf

% \AX{
\If{$t \textbf{ mod } 5 = 0$}
\State Compute contrastive loss $\mathcal{L}^{ctrs}$ via Eq.\eqref{latent_loss} with $\lambda_{ctrs} = 0.1$;
\Else
\State $\lambda_{ctrs} = 0$ in Eq.\eqref{generator_sod_conf} and Eq.~\eqref{generator_cod_conf}.
\EndIf
% }

\State Compute generator loss $\mathcal{L}^s$
% , and confidence-aware adversarial loss $\mathcal{L}_{adv}^s$ for the salient object detection task with $D_s$, 
and update $\alpha_s$,  $\beta$ via loss function in Eq.~\eqref{generator_sod_conf};
\State Compute generator loss $\mathcal{L}^c$
% , and confidence-aware adversarial loss $\mathcal{L}_{adv}^c$ for the camouflaged object detection task with $D_c$, 
and update $\alpha_c$,  $\beta$ via loss function in Eq.~\eqref{generator_cod_conf};
\State Compute discriminator loss $\mathcal{L}_{dis}^s$ and update  $\gamma_s$ via Eq.~\eqref{dis_sod};
\State Compute discriminator loss $\mathcal{L}_{dis}^c$ and update  $\gamma_c$ via Eq.~\eqref{dis_cod};
% \If{$t \% 5 =0$} 
% \State Compute $\mathcal{L}_{ctrs}$ for contradicting modeling with $D_p$ and update $\alpha_s$, $\alpha_c$, $\theta$  via loss function in Eq.~\eqref{latent_loss};
\EndFor
\end{algorithmic} 
\label{our_alg}
% \vspace{5mm}
\end{algorithm}

\begin{table*}[t!]
  \centering
  \scriptsize
  \renewcommand{\arraystretch}{1.4}
  \renewcommand{\tabcolsep}{0.7mm}
  \caption{
%   Performance comparison with benchmark saliency detection models.
Performance comparison
%   benchmark
  with
%   re-implemented 
%   COD models from SOD and 
  benchmark SOD models, where \enquote{BkB} indicates the backbone model. \enquote{V16} and \enquote{R50} are VGG16Net and Res50Net respectively. \enquote{RV} represents that the model use both VGG16Net and Res50Net backbones. \Firstre{\textbf{Red}} and \Secondre{\textbf{blue}} denote the best and the second-best results, respectively.
  }
  \begin{tabular}{l|l|l|cccc|cccc|cccc|cccc|cccc|cccc}
  % \hline
\toprule
 && &\multicolumn{4}{c|}{DUTS~\cite{wang2017learning}}&\multicolumn{4}{c|}{ECSSD~\cite{yan2013hierarchical}}&\multicolumn{4}{c|}{DUT~\cite{Manifold-Ranking:CVPR-2013}}&\multicolumn{4}{c|}{HKU-IS~\cite{li2015visual}}&\multicolumn{4}{c|}{PASCAL-S~\cite{pascal_s_dataset}}&\multicolumn{4}{c}{SOD~\cite{sod_dataset}} \\
    Method &Year&BkB& $S_{\alpha}\uparrow$&$F_{\beta}\uparrow$&$E_{\xi}\uparrow$&$\mathcal{M}\downarrow$& $S_{\alpha}\uparrow$&$F_{\beta}\uparrow$&$E_{\xi}\uparrow$&$\mathcal{M}\downarrow$& $S_{\alpha}\uparrow$&$F_{\beta}\uparrow$&$E_{\xi}\uparrow$&$\mathcal{M}\downarrow$& $S_{\alpha}\uparrow$&$F_{\beta}\uparrow$&$E_{\xi}\uparrow$&$\mathcal{M}\downarrow$& $S_{\alpha}\uparrow$&$F_{\beta}\uparrow$&$E_{\xi}\uparrow$&$\mathcal{M}\downarrow$& $S_{\alpha}\uparrow$&$F_{\beta}\uparrow$&$E_{\xi}\uparrow$&$\mathcal{M}\downarrow$ \\ \hline
%   DGRL & .846 & .790 & .887 & .051 & .902 & .898 & .934 & .045 & .809 & .726 & .845 & .063 & .897 & .884 & .939 & .037 & .816 & .727 & .838 & .077 & - & - & - & -  \\
%     NLDF~\cite{nldf_sal} &2017&V16& .816 & .757 & .851 & .065 & .870 & .871 & .896 & .066 & .770 & .683 & .798 & .080 & .879 & .871 & .914 & .048 & .802 & .794 & .841 & .101  & .784 & .782 & .812 & .103\\ 
%   PiCANet~\cite{picanet} & 2018&V16&.842 & .757 & .853 & .062 & .898 & .872 & .909 & .054 & .817 & .711 & .823 & .072 & .895 & .854 & .910 & .046 & .844 & .844 & .892 & .067  & .808 & .807 & .844 & .076 \\ 
   CPD~\cite{cpd_sal} &2019&R50& .869 & .821 & .898 & .043 & .913 & .909 & .937 & .040 & .825 & .742 & .847 & .056 & .906 & .892 & .938 & .034 & .848 & .819 & .882 & .071 & .799 & .779 & .811 & .088  \\
   SCRN~\cite{scrn_sal} &2019&R50& .885 & .833 & .900 & .040 & .920 & .910 & .933 & .041 & .837 & .749 & .847 & .056 & .916 & .894 & .935 & .034 & .869 & .833 & .892 & .063 & .817 & .790 & .829 & .087\\ 
   PoolNet~\cite{Liu19PoolNet} &2019&R50& .887 & .840 & .910 & .037 & .919 & .913 & .938 & .038 & .831 & .748 & .848 & .054 & .919 & .903 & .945 & .030 & .865 & .835 & .896 & .065 & .820 & .804 & .834 & .084 \\ 
    BASNet~\cite{basnet_sal} &2019&R34& .876 & .823 & .896 & .048 & .910 & .913 & .938 & .040 & .836 & .767 & .865 & .057 & .909 & .903 & .943 & .032 & .838 & .818 & .879 & .076 & .798 & .792 & .827 & .094\\ 
   EGNet~\cite{zhao2019EGNet} &2019&R50& .878 & .824 & .898 & .043 & .914 & .906 & .933 & .043 & .840 & .755 & .855 & .054 & .917 & .900 & .943 & .031 & .852 & .823 & .881 & .074 & .824 & .811 & .843 & .081 \\
     AFNet~\cite{feng2019attentive} &2019&V16& .867 & .812 & .893 & .046 & .907 & .901 & .929 & .045 & .826 & .743 & .846 & .057 & .905 & .888 & .934 & .036 & .850 & .837 & .886 & .071  & . & . & . & .\\ 
     F3Net~\cite{wei2020f3net} &2019&R50& .888 & .852 & .920 & .035 & .919 & .921 & .943 & .036 & .839 &  .766 & .864 & .053 & .917 & .910 & .952 & .028 & .861 & .835 & .898 & \Secondre{\textbf{.062}} & .824 & .814 & .850 & .077\\
      CSNet~\cite{gao2020highly} &2020&R50& .884 & .834 & .907 & .040 & .920 & .911 & .940 & .038 & .836 & .750 & .852 & .055 & .918 & .900 & .944 & .031 & .810 & .752 & .810 & .107  & .770 & .688 & .750 & .120 \\
   
   ITSD~\cite{zhou2020interactive} &2020&R50& .886 & .841 & .917 & .039 & .920 & .916 & .943 & .037 & .842 & .767 & .867 & .056 & .921 & .906 & .950 & .030 & .860 & .830 & .894 & .066 & .836 & .829 & .867 & .076\\

  LDF \cite{wei2020label} &2020&R50& .892 & .861 & .925 & .034 & .919 & .923 & .943 & .036 & .839 & .770 & .865 & .052 & .920 & .913 & .953 & .028 & .860 & .856 & .901 & .063  & .826 & .822 & .852 & .075\\ 

  GateNet \cite{GateNet} &2020&R50&  .885 & .833 & .902 & .040 & .920 & .910 & .934 & .041 &.837 & .752 & .851 & .055 & .915 & .896 & .937 & .034 & .654 & .835 & .879 & .072  & .828 & .809 & .844 & .081 \\ 
%   Best & .888 & .852 & .920 & .035 & .920 & .921 & .943 & .036 & .842 & .767 & .867 & .053 & .921 & .910 & .952 & .028 & .845 & .761 & .858 & .066  & .858 & .363 & .873 & .078\\ \hline 
UIS \cite{zhang2020uncertainty}  &2021&R50& .888 & .860 & .927 & .034 & .921 & .926 & .947 & .035 & .839 & .773 & .869 & {.051} & .921 & .919 & .957 & \Firstre{\textbf{.026}} & .848 & .836 & .899 & .063  & .808 & .808 & .847 & .079 \\ 
PAKRN \cite{xu2021locate}&2021&R50 & \Firstre{\textbf{.900}} & \Firstre{\textbf{.876}} & \Secondre{\textbf{.935}} & .033  & .928 &  {.930} & .951 & .032 & \Firstre{\textbf{.853}}&  \Firstre{\textbf{.796}} &  \Firstre{\textbf{.888}} & \Firstre{ \textbf{.050}} &\Secondre{ \textbf{.923}} & .919 & .955 & .028 & .859 & .856 & .898 & .068  & .833 & .836 & .866 & .074 \\ 
% DCN ~\cite{DCN_TIP} &2021&R50& .892 & .859 & .924 & .035  & .928 & .931 & .954 & .032 & .846 & .779 & .875 & .051 & .922 & .916 & .957 & .027 & . & . & . & . & . & . & . & .  \\ 
MSFNet \cite{Miao_2021_ACM_MM} &2021&R50& .877 & .855 & .927 & .034  & .915 & .927 & .951 & .033 & .832 & .772 & .873 &  \Firstre{\textbf{.050}} & .909 & .913 & .957 & .027 & .849 & .855 & .900 & .064  & .813 & .822 & .852 & .077  \\ 
MMDF\cite{wu2022recursive}&2022&RV& .849 & .819 & .891 & .050  & .902 & .915 & .931 & .045 & .831 & .775 & .870 & .053 & .901 & .910 & .942 & .034 & .837 & .825 & .875 & .077 & .786 & .798 & .814 & .098  \\
EDN\cite{wu2022edn} &2022&R50& .892 & .860 & .922 & .035  & .927 & .927 & .948 & .033 & \Secondre{ \textbf{.850}} & \Secondre{\textbf{.785}} & .874 &  \Firstre{\textbf{.050}} & .922 & .913 & .951 & .028 & .861 & \Secondre{\textbf{.857}} & .895 & .066 & .828 & .826 & .855 & .079  \\
BiconNets\cite{yang2022biconnet}&2022&R50& .892 & .859 & .917 & .035  & .923 & .923 & .940 & .037 & .841 & .771 & .862 & .052 & .922 & .913 & .946 & .029 & .863 & .855 & .895 & .067 & .813 & .808 & .833 & .082  \\
% \textbf{SOTA} &2021&R50& \textbf{.900} & \textbf{.876} & \textbf{.937} & \textbf{.032}  & \textbf{.933} & \textbf{.935} & \textbf{.960} & \textbf{.030} & \textbf{.853} & \textbf{.796} & \textbf{.888} & \textbf{.050} & \textbf{.931} & \textbf{.924} & \textbf{.967} & \textbf{.026} & \textbf{.869} & \textbf{.856} & \textbf{.902} & \textbf{.062}  & \textbf{.836} & \textbf{.836} & \textbf{.867} & \textbf{.067}  \\ 
\hline
UJSC \cite{aixuan_cod_sod21} &2021&R50& \Secondre{\textbf{.899}} & .866 & \Firstre{\textbf{.937}} & \Secondre{\textbf{.032}} & \Firstre{\textbf{.933}} &  \Firstre{\textbf{.935}} & \Firstre{\textbf{.960}} & \Secondre{\textbf{.030}} & \Secondre{\textbf{.850}} & .782 & \Secondre{\textbf{.884}} & .051 &  \Firstre{\textbf{.931}} &  \Firstre{\textbf{.924}} & \Firstre{\textbf{.967}} &  \Firstre{\textbf{.026}} &  \Secondre{\textbf{.864}}& .841 &  \Secondre{\textbf{.902}} & \Secondre{\textbf{.062}}  &  \Firstre{\textbf{.840}} & \Secondre{\textbf{.831}} & \Secondre{\textbf{.867}} & \Firstre{\textbf{.067}}  \\ 
% Ours &2021&No& . & . & . & .  & . & . & . & . & . & . & . & . & . & . & . & . & . & . & . & . & . & . & . & .  \\
%   Ours &2021&V16& . & . & . & .  & . & . & . & . & . & . & . & . & . & . & . & . & . & . & . & . & . & . & . & .  \\ 
  Ours &2023&R50& \Firstre{\textbf{.900}} & \Secondre{\textbf{.875}} & \Firstre{\textbf{.937}} &  \Firstre{\textbf{.030}} &  \Secondre{\textbf{.929}} &  \Firstre{\textbf{.935}} & \Secondre{\textbf{.955}} &  \Firstre{\textbf{.029}} & .841 & .777 & .876 &  \Firstre{\textbf{.050}} & .921 & \Secondre{\textbf{.920}} & \Secondre{\textbf{.958}} &  \Firstre{\textbf{.026} } &  \Firstre{\textbf{.866}} & \Firstre{\textbf{.867}} &  \Firstre{\textbf{.910}} & \Firstre{\textbf{.058}} & \Secondre{\textbf{.835}} &  \Firstre{\textbf{.839}} & \Firstre{\textbf{.871}} &\Secondre{\textbf{.072}} \\
%   Ours &2021&R250& . & . & . & .  & . & . & . & . & . & . & . & . & . & . & . & . & . & . & . & . & . & . & . & .  \\
%   Ours &2021&VIT& . & . & . & .  & . & . & . & . & . & . & . & . & . & . & . & . & . & . & . & . & . & . & . & .  \\
%   \hline 
%   Just\_SOD & . & . & . & .0353  & . & . & . & .0290 & . & . & . & .0526 & . & . & . & .0266 & . & . & . & .0685 & . & . & . & .1175  \\
%   Just\_SOD & .890 & .866 & .930 & .033  & .921 & .930 & .950 & .033 & .839 & .776 & .870 & .049 & .922 & .921 & .958 & .026 & .842 & .772 & .867 & .062 & .863 & .371 & .887 & .074  \\
%   Joint & . & . & . & .0413  & . & . & . & .0391 & . & . & . & .0607 & . & . & . & .0317 & . & . & . & .0663 & . & . & . & .0971  \\ 
   % \hline 
\bottomrule
  \end{tabular}
  \label{tab:benchmark_sod_model_comparison}
%   \vspace{-5mm}
\end{table*}

\section{Experimental Results}
\subsection{Setting:}
\noindent\textbf{Dataset:} For salient object detection, we train our model using the augmented DUTS training dataset~\cite{wang2017learning} via data interaction (see Sec.~\ref{subsub:data_interaction}), and testing on six other testing dataset, including the DUTS testing datasets, ECSSD~\cite{yan2013hierarchical}, DUT~\cite{Manifold-Ranking:CVPR-2013}, HKU-IS~\cite{li2015visual}, PASCAL-S dataset~\cite{pascal_s_dataset} and SOD dataset~\cite{sod_dataset}.
% ECSSD~\cite{yan2013hierarchical} contains 1,000 images with multiple salient objects. DUT~\cite{Manifold-Ranking:CVPR-2013} dataset contains 5,168 carefully labelled images by five users, and background of images in this dataset is much more complex.
For camouflaged object detection, we train our model using the benchmark COD training dataset, which is a combination of COD10K training set~\cite{fan2020camouflaged} and CAMO training dataset~\cite{le2019anabranch}, and test on four camouflaged object detection testing sets, including the CAMO testing dataset \cite{le2019anabranch}, CHAMELEON \cite{Chameleon2018}, COD10K testing dataset \cite{fan2020camouflaged} and NC4K dataset \cite{yunqiu_cod21}.

% \textbf{CAMO \cite{le2019anabranch}} dataset is split into a training set and a testing set containing 1,000 and 250 images respectively. It has 8 main categories of camouflaged objects including \textit{Mammal}, \textit{Insect}, \textit{Bird}, \textit{Underwater}, \textit{Reptile}, \textit{Amphibian}, \textit{Body Art} and \textit{Soldier}.
% \textbf{CHAMELEON \cite{Chameleon2018}} dataset contains 76 images of camouflaged animal. It is collected for evaluation purpose.
% \textbf{COD10K \cite{fan2020camouflaged}} dataset contains images of both natural and artificial camouflaged objects. It is split into a training set of 3,040 images and a testing set of 2,026 images. \textbf{NC4K \cite{yunqiu_cod21}} dataset contains 4,121 images of camouflaged objects downloaded from the internet. It is collected for evaluation purpose.

\noindent\textbf{Evaluation Metrics:}
We use four evaluation metrics to evaluate the performance of the salient object detection models and the camouflaged object detection models, including Mean Absolute Error ($\mathcal{M}$), Mean F-measure ($F_\beta$), Mean E-measure \cite{fan2018enhanced} ($E_\xi$) and S-measure \cite{fan2017structure} ($S_{\alpha}$).
% A detailed introduction to those metrics appears in the supplementary materials.

\noindent\textit{Mean Absolute Error ($\mathcal{M}$):} measures the pixel-level pairwise errors between the prediction $s$
% camouflage map $f_{\theta}(x)$ 
and the ground-truth map $y$, which is defined as:
\begin{equation}
    \mathcal{M} = \frac{\sum_{u=1}^W\sum_{v=1}^H |y^{u, v} - s^{u,v} |}{W \times H},
    \label{eq: MAE}
\end{equation}
% \begin{equation}
%     \mathcal{M} = \frac{\sum_{n=1}^{N}\sum_{u, v}^{W, H} |y_{i}^{u, v} - s^{u,v} |}{N \times W \times H}
%     \label{eq: MAE}
% \end{equation}
% where $f_{\theta}(x)$ is the predicted camouflage map, $y$ is the ground-truth camouflage map, $N$ is the total number of samples in the testing dataset, 
where $W$ and $H$ indicate size of
% height and width of 
the ground-truth map.

\noindent\textit{Mean F-measure ($F_{\beta}$):} measures the precision and robustness of the model, which is
% camouflaged object detection model. It is 
defined as:
% in Eq.~\ref{eq: mean F-measure}:
\begin{equation}
    F_{\beta} = \frac{TP}{TP + \frac{1}{2}(FP + FN)},
    \label{eq:mean_F-measure}
\end{equation}
% \begin{equation}
%     F_{\beta} = \frac{1}{N} \sum_{i=1}^{N} \frac{TP_{i}}{TP_{i} + \frac{1}{2}(FP_{i} + FN_{i})}
%     \label{eq: mean F-measure}
% \end{equation}
where $TP$ denotes the number of true positives, $FP$ shows the false positives and $FN$ indicates the false negatives.
% , and $N$ is the total number of samples in the testing dataset.

\noindent\textit{Mean E-measure ($E_{\xi}$):} measures the pixel-level matching and image-level statistics of the prediction \cite{fan2018enhanced}, which is defined as:
\begin{equation}
\begin{aligned}
    E_{\xi} &= \frac{1}{W \times H} \sum_{u=1}^W\sum_{v=1}^H \phi_{p}(u, v),
\end{aligned}
\label{eq:mean_E-measure}
\end{equation}
where $\phi_{p}(u, v)$ is the alignment matrix \cite{fan2018enhanced}, measuring the alignment of model prediction and the ground truth.
% with
% \begin{equation}
% \begin{aligned}
%     \phi_{p} &= \frac{1}{4}(1 + \xi_{p})^{2},\\
%     \xi_{p} &= \frac{2 \varphi_{y} \circ \varphi_{p}}{\varphi_{y} \circ \varphi_{y} + \varphi_{p} \circ \varphi_{p}},\\
%     \varphi_{y} &= y - \mu_{y} \cdot \mathbb{A},\\
%     \varphi_{p} &= s - \mu_{s} \cdot \mathbb{A},
% \end{aligned}
% \end{equation}
% % where $Q_{p}$ is the mean E-measure, 
% where $\mu_s$ and $\mu_y$ are the
% % denotes the 
% mean of prediction $s$ and ground truth $y$ respectively,
% % the predicted camouflage map $f_{\theta}(x)$ or ground-truth camouflage map $y$, 
% $\mathbb{A}$ is a all-one matrix of the same size as $s$.
% denotes a map with corresponding size to those of the predicted camouflage map or ground-truth camouflage map. All elements in $\mathbb{A}$ have value $1$.

\begin{table*}[t!]
  \centering
  \scriptsize
  \renewcommand{\arraystretch}{1.40}
  \renewcommand{\tabcolsep}{1.6mm}
  \caption{Performance comparison
%   benchmark
  with
%   re-implemented 
%   COD models from SOD and 
  benchmark COD models, where \enquote{BkB} indicates the backbone models. \enquote{V16}, \enquote{R50}, \enquote{R34} and \enquote{R250} are VGG16Net, Res50Net, Res34Net and Res2Net50 \cite{res2net} respectively. \enquote{RL} indicates the resolution of the training/testing images.
  % \Firstre{\textbf{Red}} and \Secondre{\textbf{blue}} denote the best and the second-best results, respectively.
  }
  \begin{tabular}{l|l|l|l|cccc|cccc|cccc|cccc}
  % \hline
  \toprule
  &&&&\multicolumn{4}{c|}{CAMO~\cite{le2019anabranch}}&\multicolumn{4}{c|}{CHAMELEON~\cite{Chameleon2018}}&\multicolumn{4}{c}{COD10K~\cite{fan2020camouflaged}}&\multicolumn{4}{c}{NC4K~\cite{yunqiu_cod21}} \\
    Method &Year&BkB&RL& $S_{\alpha}\uparrow$&$F_{\beta}\uparrow$&$E_{\xi}\uparrow$&$\mathcal{M}\downarrow$& $S_{\alpha}\uparrow$&$F_{\beta}\uparrow$&$E_{\xi}\uparrow$&$\mathcal{M}\downarrow$ &  $S_{\alpha}\uparrow$ & $F_{\beta}\uparrow$ & $E_{\xi}\uparrow$ & $\mathcal{M}\downarrow$ &  $S_{\alpha}\uparrow$ & $F_{\beta}\uparrow$ & $E_{\xi}\uparrow$ & $\mathcal{M}\downarrow$  \\
  \hline
  SINet~\cite{fan2020camouflaged} &2020&R50&$352\times352$& .745 & .702 & .804 & .092 & .872 & .827 & .936 & .034 & .776 & .679 & .864 & .043 & .810 & .772 & .873 & .057  \\ 
  SINet-V2~\cite{fan2021concealed} &2021&R250&$352\times352$& \Firstre{\textbf{.820}} & \Firstre{\textbf{.782}} &  \Firstre{\textbf{.882}} &  \Firstre{\textbf{.070}} & .888 & .835 & .942 & \Secondre{\textbf{.030}} & .815 & .718 & .887 & .037 & \Secondre{\textbf{.847}} & .805& \Secondre{\textbf{.903}} & .048\\ 
  PFNet~\cite{mei2021Ming} &2021&R50&$416\times416$& .782 & .744 & .840 & .085 & .882 & .826 & .922 & .033 & .800 & .700 & .875 & .040 & .829 & .782 & .886 & .053 \\
  % LSR~\cite{yunqiu_cod21}  &2021&R50&$352\times352$& .793 & .725 & .826 & .085 & .893 & .839 & .938 & .033 & .793 & .685 & .868 & .041 & .839 & .779 & .883 & .053  \\
  MGL~\cite{zhai2021Mutual} &2021&R50&$473\times473$& .775 & .726 & .812 & .088 & .893 & .834 & .918 & \Secondre{\textbf{.030}} & .814 & .711 & .852 & .035 & .833 & .782 & .867 & .052\\
%   PraNet~\cite{fan2020pranet} & .769 & .711 & .825 & .094 & .860 & .790 & .908 & .044 & .790 & .672 & .861 & .045 & .822 & .763 & .877 & .059\\
  C2FNet \cite{sun2021c2fnet} & 2021 & R50 & $352\times352$   & .611 & .481 & .672 & .147 & .791 & .704 & .860 & .069 & .638 & .438 & .718 & .089 & .681 & .570 & .744 & .110  \\
  C2FNet-V2 \cite{sun2021c2fnet} & 2021 & R250 & $352\times352$  & .772 & .737 & .825 & .087 & .889 & .853 & .941 & .030 & .807 & .719 & .883 & .036 & .837 & .805 & .894 & .049  \\
  ERRNet \cite{ji2022fast} & 2022 & R50 & $352\times352$   & .690 & .599 & .730 & .112 & .825 & .756 & .888 & .047 & .715 & .572 & .795 & .053 & .764 & .697 & .833 & .071 \\
  ZoomNet \cite{ZoomNet_CVPR2022} &  2022 & R50  & $384\times384$ & .789 & .741 & .829 & .076 & .865 & .823 & .939 & .031 &  \Firstre{\textbf{.821}} & \Firstre{\textbf{.741}} & .866 & \Firstre{\textbf{.032}} & .839 & .796 & .867 & \Secondre{\textbf{.046}} \\
  LSR+~\cite{lv2023towards}  &2023&R50&$352\times352$& .789 & .751 & .840 & .079 & .878 & .828 & .929 & .034 & .805 & .711 & .880 & .037 & .840 & .801 & .896 & .048  \\
 \hline
%   \hline
%  &\multicolumn{16}{c}{Our Deep COD Models} \\ \hline
%   VGG16 & .694 & .639 & .748 & .113 & .859 & .809 & .926 & .042 & .718 & .583 & .812 & .056  & . & . & . & .  \\ 
%   ResNet50 & .739 & .696 & .802 & .093 & .867 & .811 & .924 & .038 & .757 & .640 & .841 & .046  & . & . & . & .  \\ 
%  Best  & .771 & .705 & .804 & .090 & .876 & .827 & .936 & .034 & .789 & .679 & .864 & .043   \\ \hline
%   Ours  &&& \textbf{.803} & \textbf{.759} & \textbf{.853} & \textbf{.076} & \textbf{.894} & \textbf{.848} & \textbf{.943} & \textbf{.030} & \textbf{.817} & \textbf{.726} & \textbf{.892} & \textbf{.035}    & . & . & . & .     \\
% Ours  &2021&No & . & . & . & .  & . & . & . & . & . & . & . & .  & . & . & . & .     \\ 
%   Ours  &2021&V16 & . & . & . & .  & . & . & . & . & . & . & . & .  & . & . & . & .     \\ 
UJSC~\cite{aixuan_cod_sod21}  &2021&R50&$352\times352$&\Secondre{\textbf{.803}} & .759 & .853 & .076 &  \Firstre{\textbf{.894}}&  \Firstre{\textbf{.848}} & \Secondre{\textbf{.943}} & \Secondre{\textbf{.030}} & \Secondre{\textbf{.817}} & .726 & \Secondre{\textbf{.892}} &  .035 &.842 &\Secondre{\textbf{.806}} &.898 &.047 \\
  Ours  &2023&R50& $352\times352$& \Secondre{\textbf{.803}} & \Secondre{\textbf{.768}}& \Secondre{\textbf{.858}} & \Secondre{\textbf{.071}} & \Secondre{\textbf{.892}} & \Firstre{\textbf{.848}} &  \Firstre{\textbf{.948}} & \Firstre{\textbf{.025}} &\Secondre{\textbf{.817}} & \Secondre{\textbf{.733}} & \Firstre{\textbf{.895}} & \Secondre{\textbf{.033}} &  \Firstre{\textbf{.856}} & \Firstre{\textbf{.824}} & \Firstre{\textbf{.913}} &  \Firstre{\textbf{.040}} \\ 
%   Ours  &2021&R250 & . & . & . & .  & . & . & . & . & . & . & . & .  & . & . & . & .     \\
%   Ours  &2021&VIT& . & . & . & .  & . & . & . & . & . & . & . & .  & . & . & . & .     \\ 
   % \hline
   \bottomrule
%   Just\_COD & .783 & .716 & .813 & .088  & .872 & .843 & .951 & .031 & .788 & .675 & .872 & .042  \\
%   Just\_COD & . & . & . & .0947  & . & . & . & .0403 & . & . & . & .0448   \\
%   Joint & . & . & . & .1059  & . & . & . & .0371 & . & . & . & .0453   \\
%   \hline 

  \end{tabular}
  \label{tab:benchmark_cod_model_comparison}
%   \vspace{-2mm}
\end{table*}

\noindent\textit{S-measure ($S_{\alpha}$):} measures the regional and global structural similarities between the prediction
% camouflage map 
and the ground-truth \cite{fan2017structure} as:
% which is defined as:
\begin{equation}
    S_{\alpha} = \alpha \cdot S_{o} + (1 - \alpha) \cdot S_{r}.
\label{eq:S-measure}
\end{equation}
% where $\alpha = 0.5$ is used in our implementation, complying with the setting in \cite{fan2017structure}, 
where $S_{o}$ measures the global structural similarity, in terms of the consistencies in the foreground and background predictions and contrast between the foreground and background predictions,
% as defined in Eq.~\ref{eq: s_o}. 
$S_{r}$
% divides the predicted and ground-truth camouflage maps into $K$ patches and 
measures the regional structure similarity, and $\alpha = 0.5$ balances the two similarity measures following \cite{fan2017structure}.

\noindent\textbf{Training details:}
We train our model in Pytorch with ResNet50~\cite{he2016deep} as backbone, as shown in Fig.~\ref{fig:network_overview}. Both the encoders for saliency and camouflage branches are initialized with ResNet50~\cite{he2016deep} trained on ImageNet, and other newly added layers are initialized by default. We resize all the images and ground truth to $352\times352$, and perform multi-scale training. The maximum step is 30000.
% and we iteratively update 3 times the saliency branch and then one time the camouflage branch. 
The initial learning rate are 2e-5, 2e-5 and 1.2e-5 with Adam optimizer for the generator, discriminators and contrastive module respectively.
% We adopt the \enquote{step} learning rate decay policy, and set the decay step as 24000 iteration, and decay rate as 0.1.
The whole training takes 26 hours with batch size 22 on an NVIDIA GeForce RTX 3090 GPU.

\begin{figure*}[tp]
%  \vspace{-5mm}
   \begin{center}
   \begin{tabular}{{c@{ } c@{ } c@{ } c@{ } c@{ } c@{ } c@{ } c@{ } c@{ } c@{ }}}
    {\includegraphics[width=0.093\linewidth,height=0.080\linewidth]{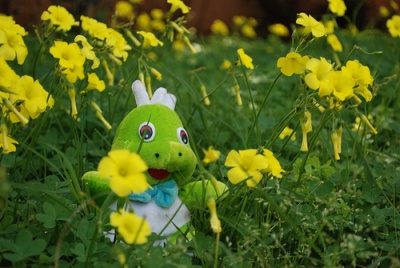}}&
    {\includegraphics[width=0.093\linewidth,height=0.080\linewidth]{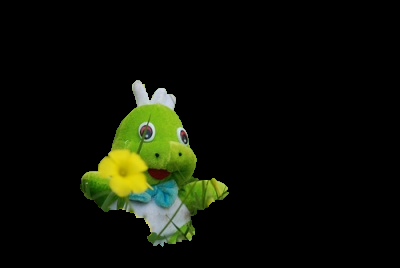}}&
    {\includegraphics[width=0.093\linewidth,height=0.080\linewidth]{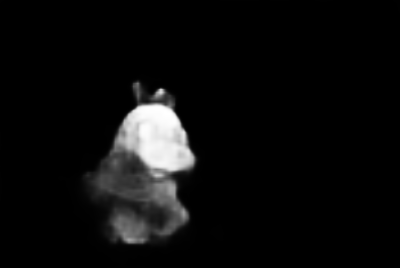}}&
    {\includegraphics[width=0.093\linewidth,height=0.080\linewidth]{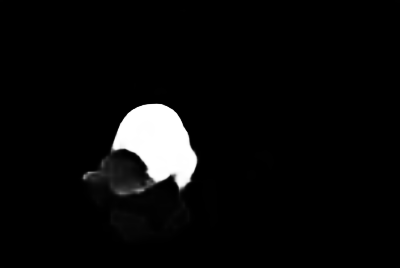}}&
    {\includegraphics[width=0.093\linewidth,height=0.080\linewidth]{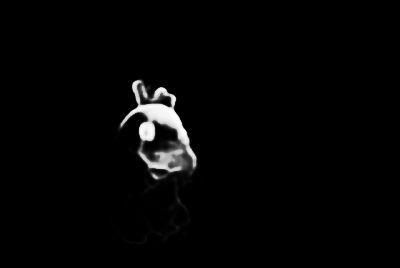}}&
    {\includegraphics[width=0.093\linewidth,height=0.080\linewidth]{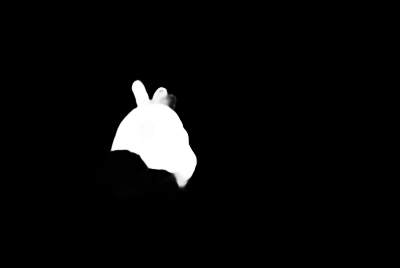}}&
    {\includegraphics[width=0.093\linewidth,height=0.080\linewidth]{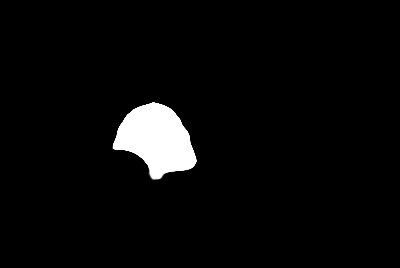}}&
    {\includegraphics[width=0.093\linewidth,height=0.080\linewidth]{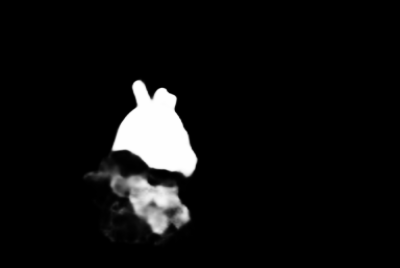}}&
    {\includegraphics[width=0.093\linewidth,height=0.080\linewidth]{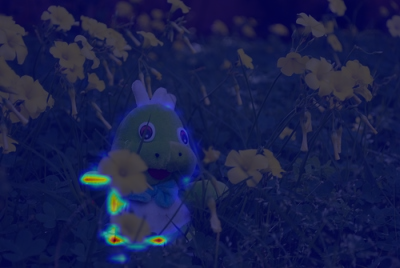}}&
    {\includegraphics[width=0.093\linewidth,height=0.080\linewidth]{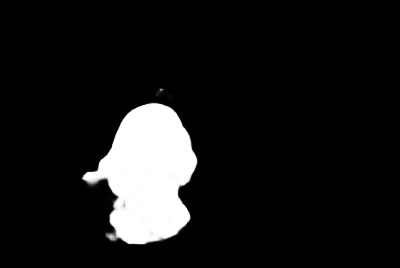}}\\
    
    {\includegraphics[width=0.093\linewidth,height=0.080\linewidth]{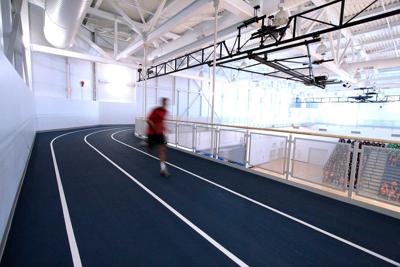}}&
    {\includegraphics[width=0.093\linewidth,height=0.080\linewidth]{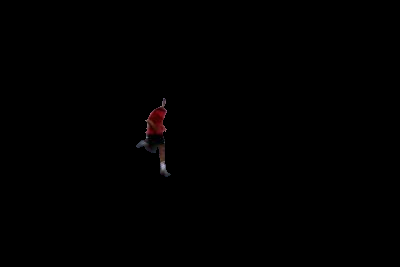}}&
    {\includegraphics[width=0.093\linewidth,height=0.080\linewidth]{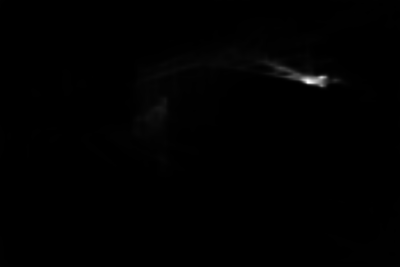}}&
    {\includegraphics[width=0.093\linewidth,height=0.080\linewidth]{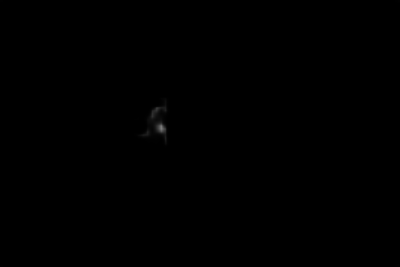}}&
    {\includegraphics[width=0.093\linewidth,height=0.080\linewidth]{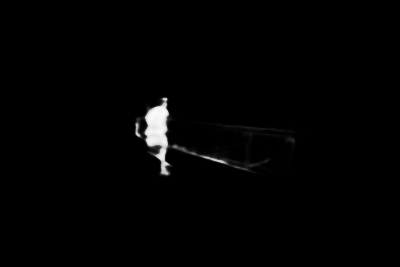}}&
    {\includegraphics[width=0.093\linewidth,height=0.080\linewidth]{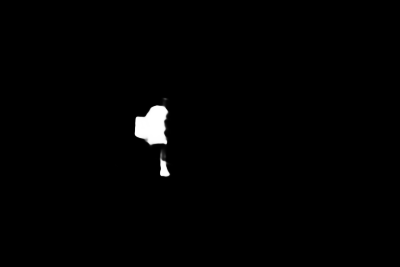}}&
    {\includegraphics[width=0.093\linewidth,height=0.080\linewidth]{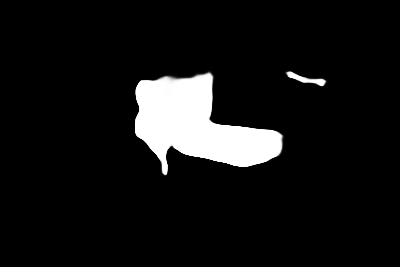}}&
    {\includegraphics[width=0.093\linewidth,height=0.080\linewidth]{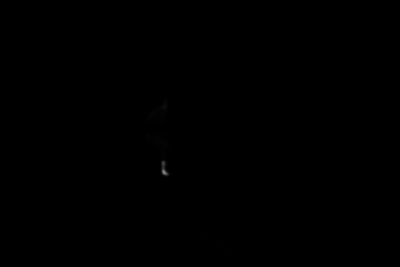}}&
    {\includegraphics[width=0.093\linewidth,height=0.080\linewidth]{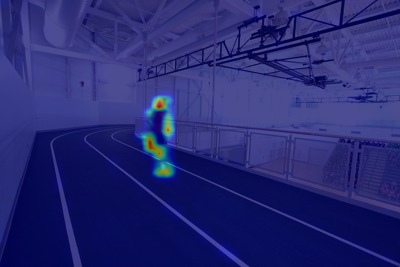}}&
    {\includegraphics[width=0.093\linewidth,height=0.080\linewidth]{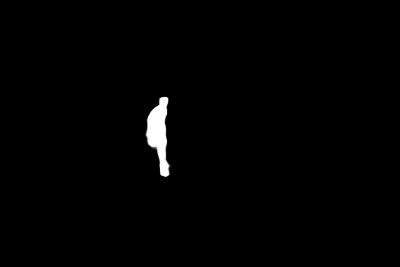}}\\
    
    {\includegraphics[width=0.093\linewidth,height=0.080\linewidth]{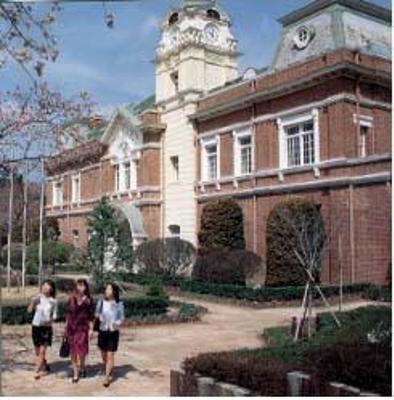}}&
    {\includegraphics[width=0.093\linewidth,height=0.080\linewidth]{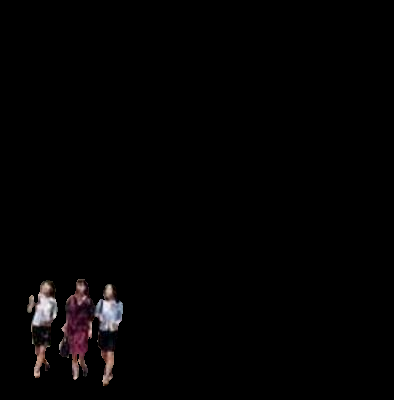}}&
    {\includegraphics[width=0.093\linewidth,height=0.080\linewidth]{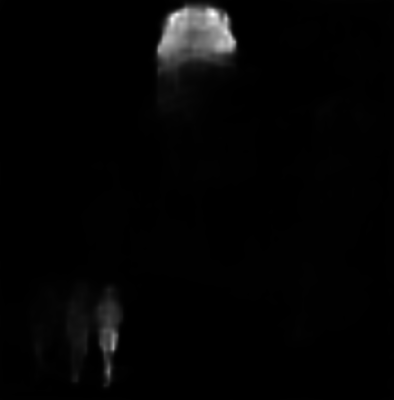}}&
    {\includegraphics[width=0.093\linewidth,height=0.080\linewidth]{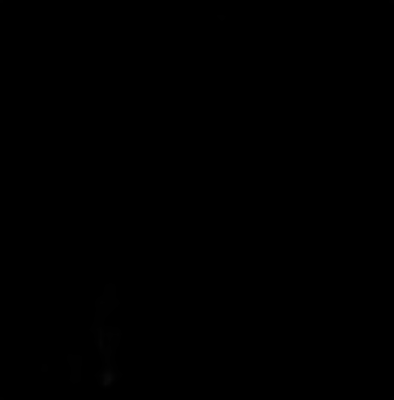}}&
    {\includegraphics[width=0.093\linewidth,height=0.080\linewidth]{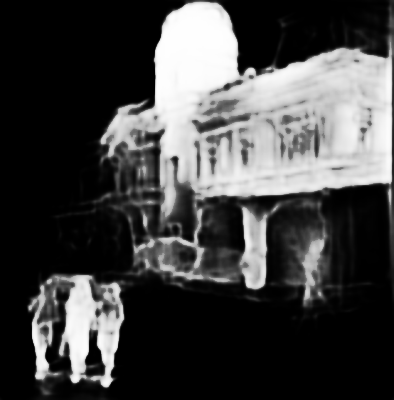}}&
    {\includegraphics[width=0.093\linewidth,height=0.080\linewidth]{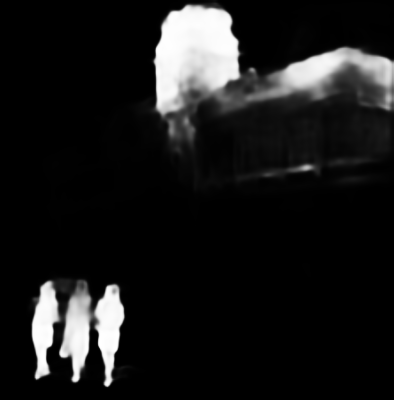}}&
    {\includegraphics[width=0.093\linewidth,height=0.080\linewidth]{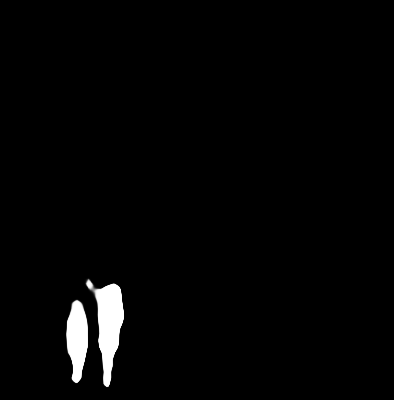}}&
    {\includegraphics[width=0.093\linewidth,height=0.080\linewidth]{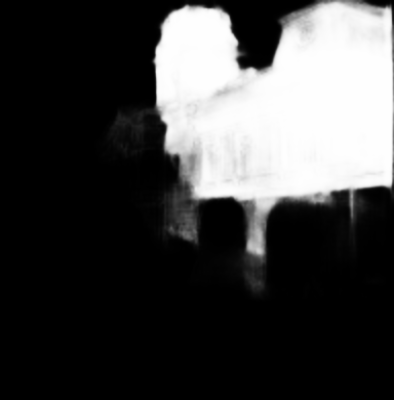}}&
    {\includegraphics[width=0.093\linewidth,height=0.080\linewidth]{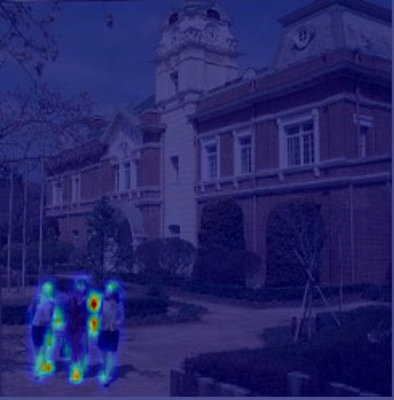}}&
    % {\includegraphics[width=0.093\linewidth,height=0.080\linewidth]{samples_sod/ILSVRC2012_test_00001410_uncertainty.png}}&
    {\includegraphics[width=0.093\linewidth,height=0.080\linewidth]{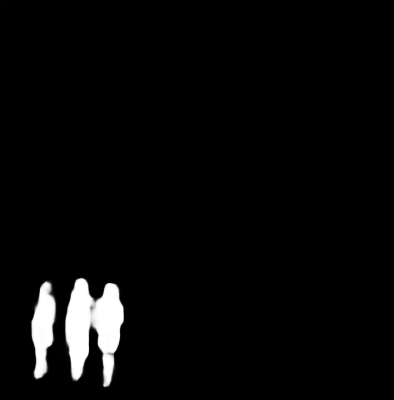}}\\

    {\includegraphics[width=0.093\linewidth,height=0.080\linewidth]{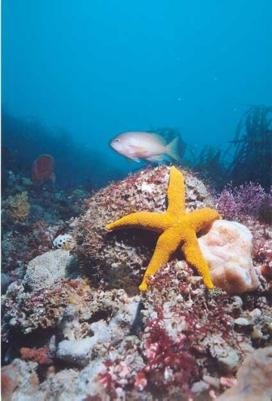}}&
    {\includegraphics[width=0.093\linewidth,height=0.080\linewidth]{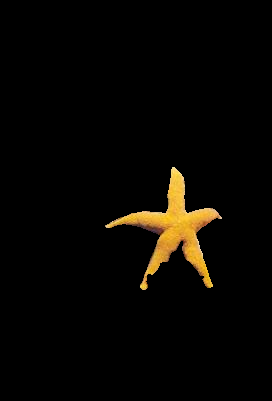}}&
    {\includegraphics[width=0.093\linewidth,height=0.080\linewidth]{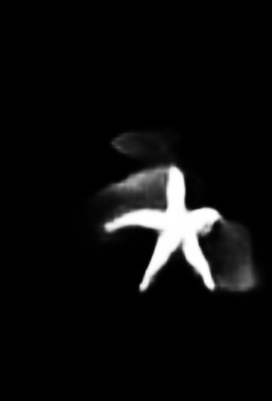}}&
    {\includegraphics[width=0.093\linewidth,height=0.080\linewidth]{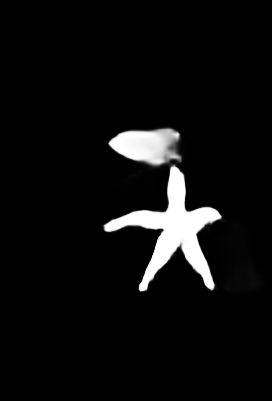}}&
    {\includegraphics[width=0.093\linewidth,height=0.080\linewidth]{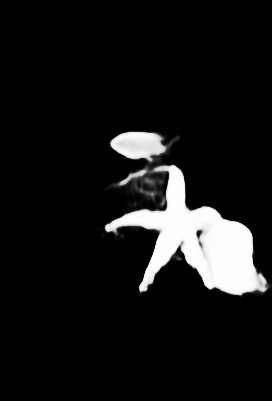}}&
    {\includegraphics[width=0.093\linewidth,height=0.080\linewidth]{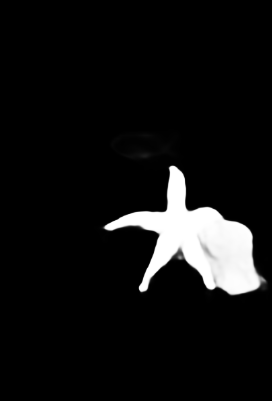}}&
    {\includegraphics[width=0.093\linewidth,height=0.080\linewidth]{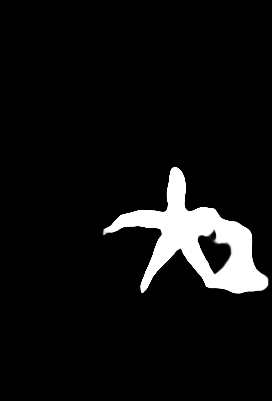}}&
    {\includegraphics[width=0.093\linewidth,height=0.080\linewidth]{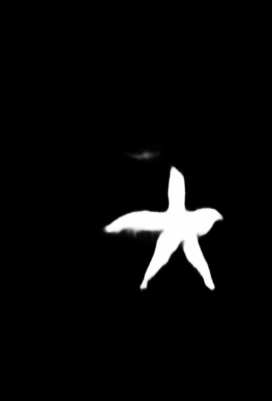}}&
    {\includegraphics[width=0.093\linewidth,height=0.080\linewidth]{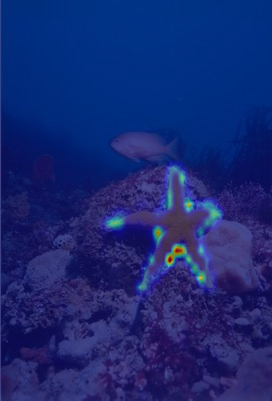}}&
    % {\includegraphics[width=0.093\linewidth,height=0.080\linewidth]{samples_sod/ILSVRC2012_test_00001410_uncertainty.png}}&
    {\includegraphics[width=0.093\linewidth,height=0.080\linewidth]{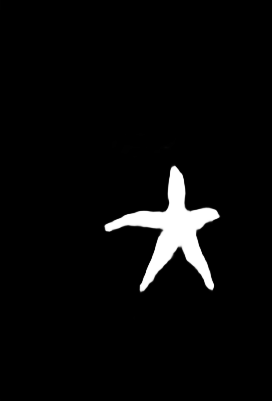}}\\
    
    % {\includegraphics[width=0.093\linewidth,height=0.080\linewidth]{pic/sod_compare/sun_agixetllrwxjrpdv.jpg}}&
    % {\includegraphics[width=0.093\linewidth,height=0.080\linewidth]{pic/sod_compare/sun_agixetllrwxjrpdv_mattiong_gt.png}}&
    % {\includegraphics[width=0.093\linewidth,height=0.080\linewidth]{pic/sod_compare/sun_agixetllrwxjrpdv_SCRN.png}}&
    % {\includegraphics[width=0.093\linewidth,height=0.080\linewidth]{pic/sod_compare/sun_agixetllrwxjrpdv_F3Net.png}}&
    % {\includegraphics[width=0.093\linewidth,height=0.080\linewidth]{pic/sod_compare/sun_agixetllrwxjrpdv_ITSD.png}}&
    % {\includegraphics[width=0.093\linewidth,height=0.080\linewidth]{pic/sod_compare/sun_agixetllrwxjrpdv_sal_fuse_PAKRN.png}}&
    % {\includegraphics[width=0.093\linewidth,height=0.080\linewidth]{pic/sod_compare/sun_agixetllrwxjrpdv_MSFNet.png}}&
    % {\includegraphics[width=0.093\linewidth,height=0.080\linewidth]{pic/sod_compare/sun_agixetllrwxjrpdv_EDN.png}}&
    % {\includegraphics[width=0.093\linewidth,height=0.080\linewidth]{pic/sod_compare/sun_agixetllrwxjrpdv_un_color.png}}&
    % % {\includegraphics[width=0.093\linewidth,height=0.080\linewidth]{samples_sod/ILSVRC2012_test_00001410_uncertainty.png}}&
    % {\includegraphics[width=0.093\linewidth,height=0.080\linewidth]{pic/sod_compare/sun_agixetllrwxjrpdv_JSCOD.png}}\\
        \footnotesize{Image} &
     \footnotesize{GT} &\footnotesize{SCRN~\cite{scrn_sal}} &\footnotesize{F3Net~\cite{wei2020f3net}}&\footnotesize{ITSD~\cite{zhou2020interactive}}&\footnotesize{PAKRN \cite{xu2021locate}} &\footnotesize{MSFNet \cite{Miao_2021_ACM_MM}}&\footnotesize{EDN\cite{wu2022edn}} &\footnotesize{Uncertainty} &\footnotesize{Ours}\\
   \end{tabular}
   \end{center}
    \caption{
    % Predictions of competing salient object detection models and ours.
    Visual comparison of our saliency predictions with the state-of-the-art models.
    % produced from our fully convolutional discriminator.
    } 
    \label{fig:sod_samples_comparison}
\end{figure*}

\begin{figure*}[tp]
%  \vspace{-5mm}
   \begin{center}
   \begin{tabular}{{c@{ } c@{ } c@{ } c@{ } c@{ } c@{ } c@{ } c@{ } c@{ } c@{ }}}
    {\includegraphics[width=0.093\linewidth,height=0.08\linewidth]{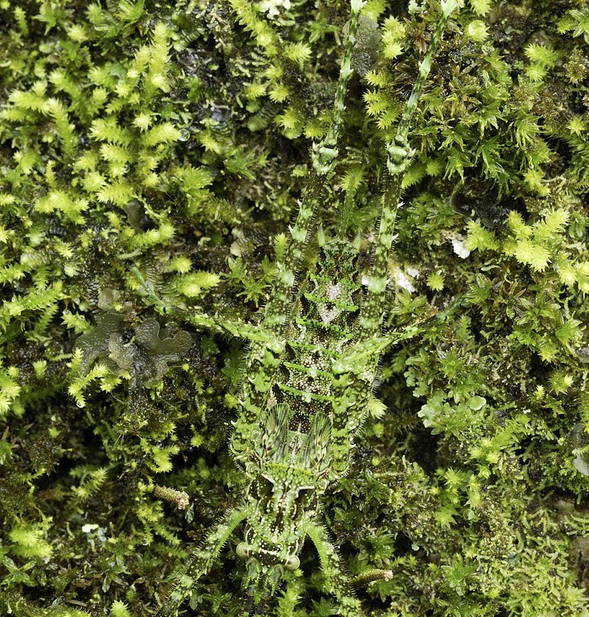}}&
    {\includegraphics[width=0.093\linewidth,height=0.08\linewidth]{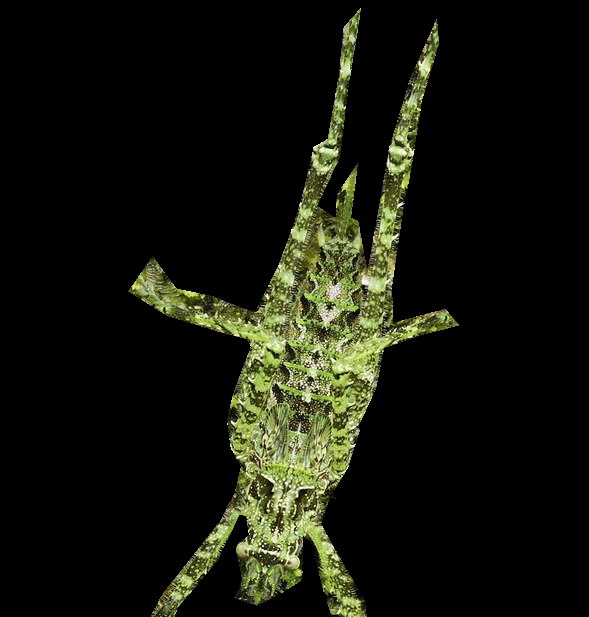}}&
    {\includegraphics[width=0.093\linewidth,height=0.08\linewidth]{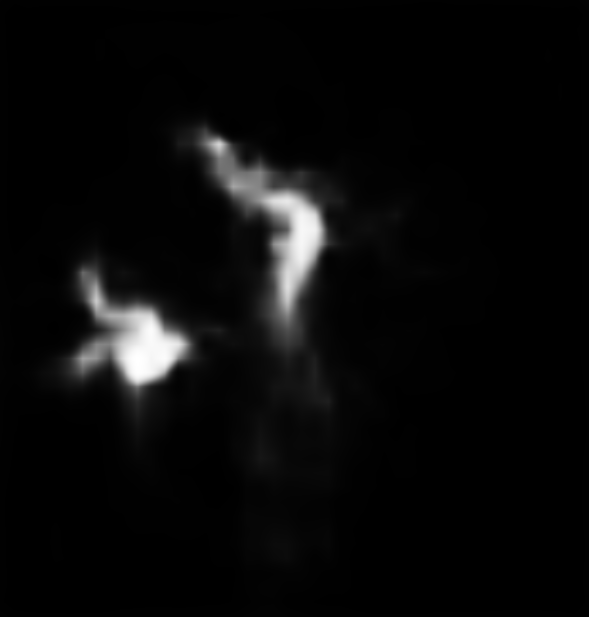}}&
    {\includegraphics[width=0.093\linewidth,height=0.08\linewidth]{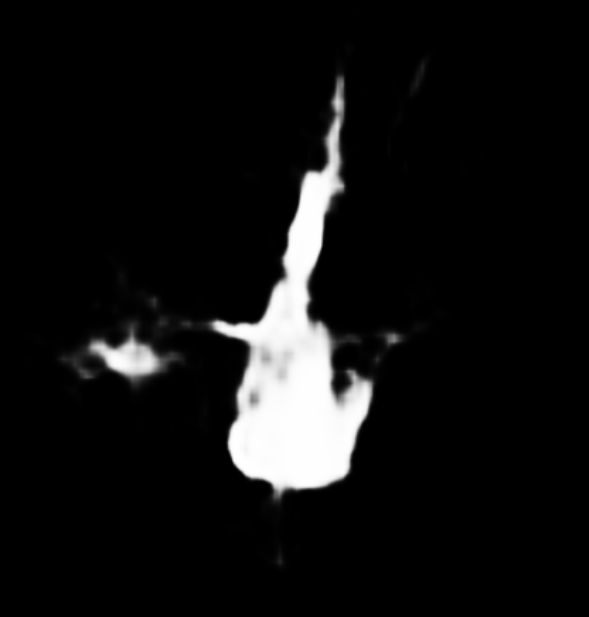}}&
    % {\includegraphics[width=0.093\linewidth,height=0.08\linewidth]{pic/cod_compare/camourflage_00064_LSR.png}}&
    {\includegraphics[width=0.093\linewidth,height=0.08\linewidth]{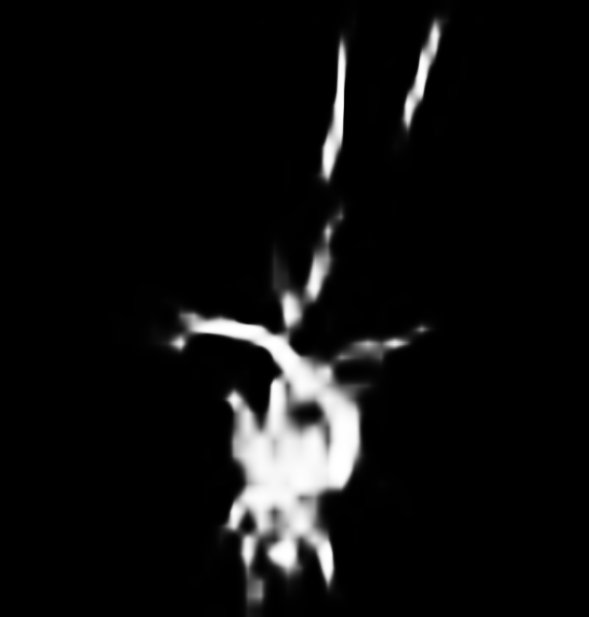}}&
    {\includegraphics[width=0.093\linewidth,height=0.08\linewidth]{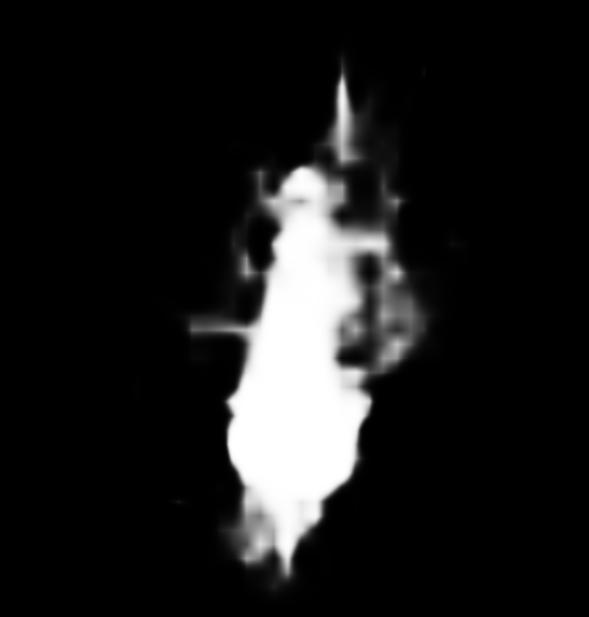}}&
    {\includegraphics[width=0.093\linewidth,height=0.08\linewidth]{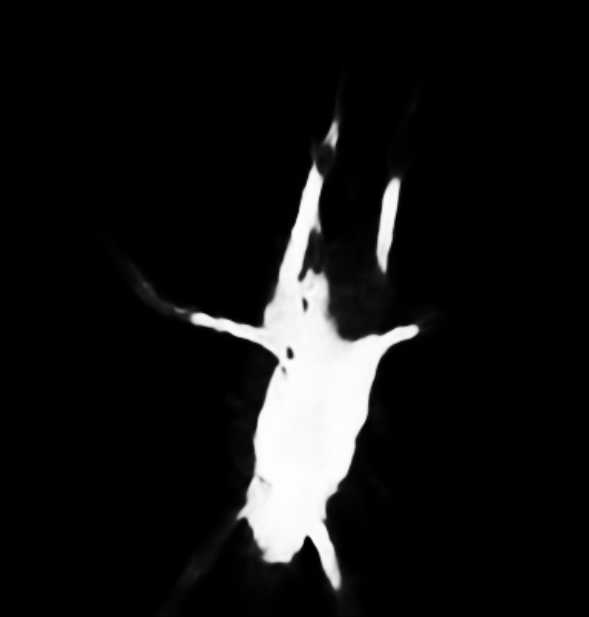}}&
    {\includegraphics[width=0.093\linewidth,height=0.08\linewidth]{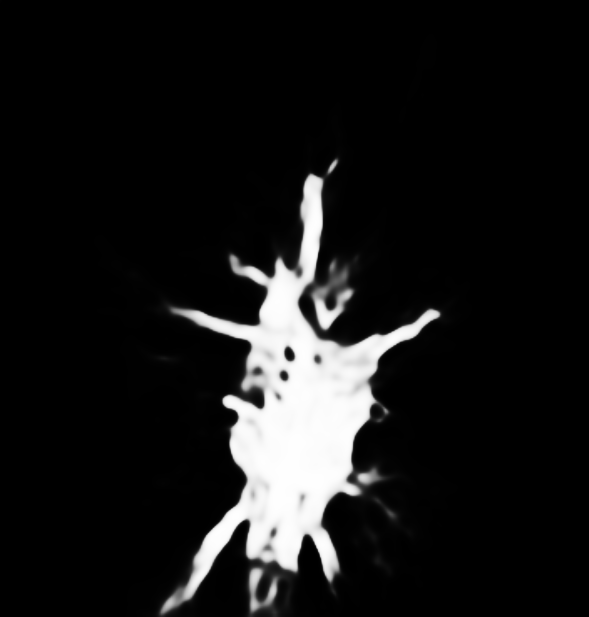}}&
    {\includegraphics[width=0.093\linewidth,height=0.08\linewidth]{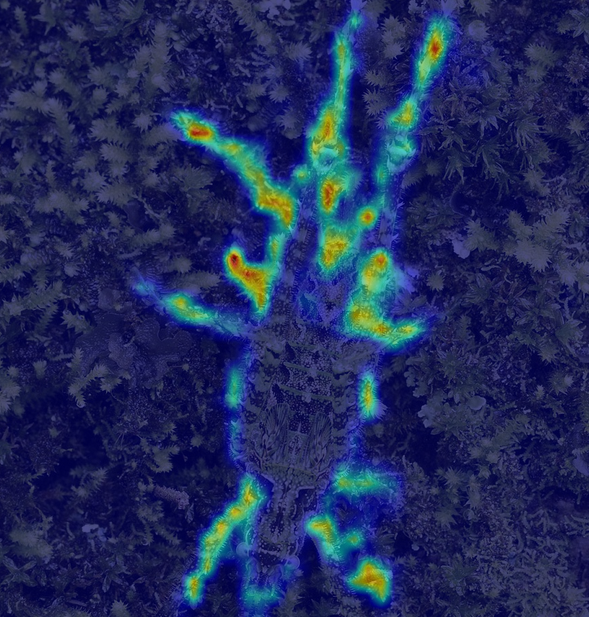}}&
    {\includegraphics[width=0.093\linewidth,height=0.08\linewidth]{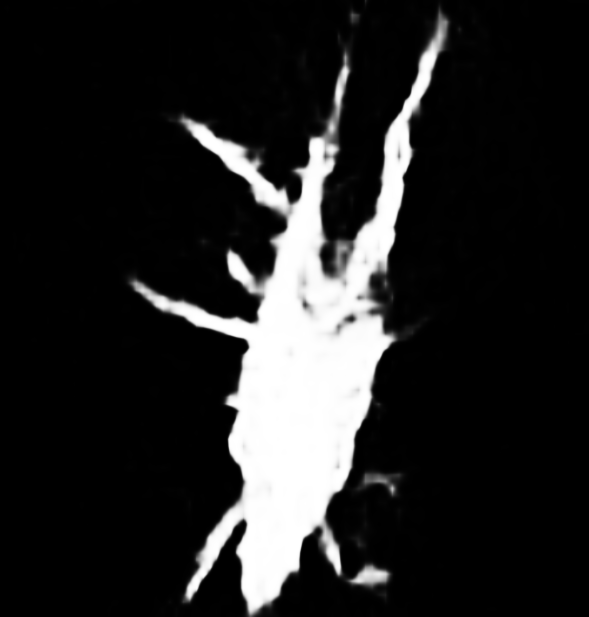}}\\

    {\includegraphics[width=0.093\linewidth,height=0.08\linewidth]{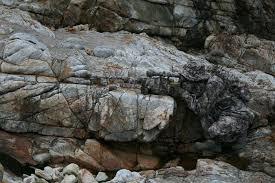}}&
    {\includegraphics[width=0.093\linewidth,height=0.08\linewidth]{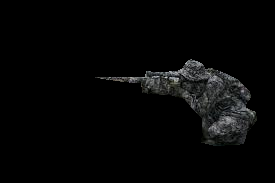}}&
    {\includegraphics[width=0.093\linewidth,height=0.08\linewidth]{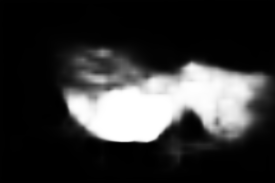}}&
    {\includegraphics[width=0.093\linewidth,height=0.08\linewidth]{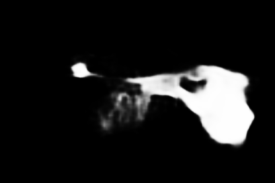}}&
    % {\includegraphics[width=0.093\linewidth,height=0.08\linewidth]{pic/cod_compare/camourflage_01090_LSR.png}}&
    {\includegraphics[width=0.093\linewidth,height=0.08\linewidth]{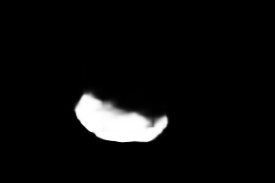}}&
    {\includegraphics[width=0.093\linewidth,height=0.08\linewidth]{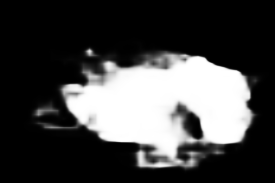}}&
    {\includegraphics[width=0.093\linewidth,height=0.08\linewidth]{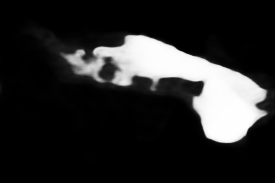}}&
    {\includegraphics[width=0.093\linewidth,height=0.08\linewidth]{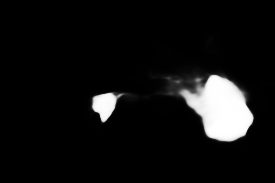}}&
    {\includegraphics[width=0.093\linewidth,height=0.08\linewidth]{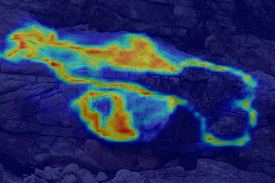}}&
    {\includegraphics[width=0.093\linewidth,height=0.08\linewidth]{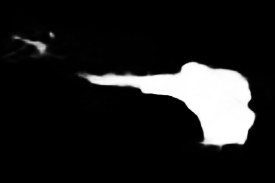}}\\
    
    % {\includegraphics[width=0.093\linewidth,height=0.08\linewidth]{pic/cod_compare/camourflage_00124.jpg}}&
    % {\includegraphics[width=0.093\linewidth,height=0.08\linewidth]{pic/cod_compare/camourflage_00124_mattiong_gt.png}}&
    % {\includegraphics[width=0.093\linewidth,height=0.08\linewidth]{pic/cod_compare/camourflage_00124_SINet.png}}&
    % {\includegraphics[width=0.093\linewidth,height=0.08\linewidth]{pic/cod_compare/camourflage_00124_PFNet.png}}&
    % % {\includegraphics[width=0.093\linewidth,height=0.08\linewidth]{pic/cod_compare/camourflage_00124_LSR.png}}&
    % {\includegraphics[width=0.093\linewidth,height=0.08\linewidth]{pic/cod_compare/camourflage_00124_C2FNet.png}}&
    % {\includegraphics[width=0.093\linewidth,height=0.08\linewidth]{pic/cod_compare/camourflage_00124_ERRNet.png}}&
    % {\includegraphics[width=0.093\linewidth,height=0.08\linewidth]{pic/cod_compare/camourflage_00124_ZoomNet.png}}&
    % {\includegraphics[width=0.093\linewidth,height=0.08\linewidth]{pic/cod_compare/camourflage_00124_LSR+.png}}&
    % {\includegraphics[width=0.093\linewidth,height=0.08\linewidth]{pic/cod_compare/camourflage_00124_un_color.png}}&
    % {\includegraphics[width=0.093\linewidth,height=0.08\linewidth]{pic/cod_compare/camourflage_00124_JSCOD.png}}\\
    
    {\includegraphics[width=0.093\linewidth,height=0.08\linewidth]{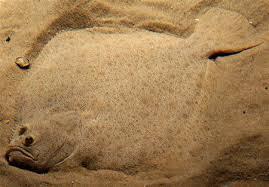}}&
    {\includegraphics[width=0.093\linewidth,height=0.08\linewidth]{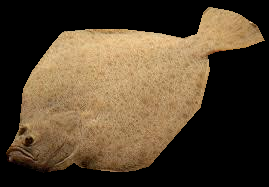}}&
    {\includegraphics[width=0.093\linewidth,height=0.08\linewidth]{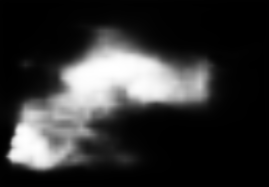}}&
    {\includegraphics[width=0.093\linewidth,height=0.08\linewidth]{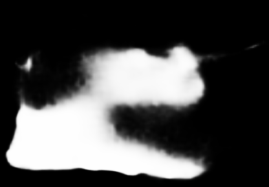}}&
    % {\includegraphics[width=0.093\linewidth,height=0.08\linewidth]{pic/cod_compare/camourflage_00310_LSR.png}}&
    {\includegraphics[width=0.093\linewidth,height=0.08\linewidth]{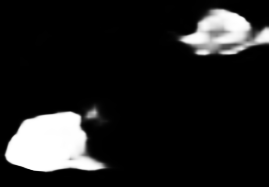}}&
    {\includegraphics[width=0.093\linewidth,height=0.08\linewidth]{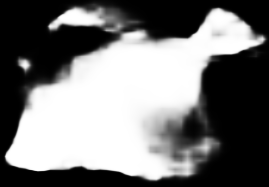}}&
    {\includegraphics[width=0.093\linewidth,height=0.08\linewidth]{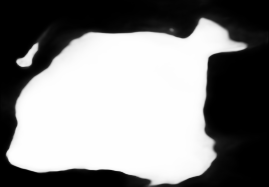}}&
    {\includegraphics[width=0.093\linewidth,height=0.08\linewidth]{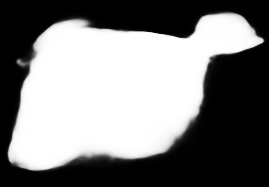}}&
    {\includegraphics[width=0.093\linewidth,height=0.08\linewidth]{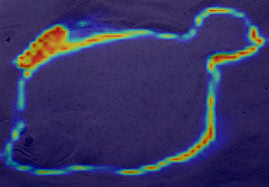}}&
    {\includegraphics[width=0.093\linewidth,height=0.08\linewidth]{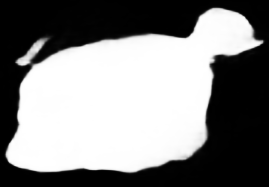}}\\
    
    % {\includegraphics[width=0.093\linewidth,height=0.08\linewidth]{pic/cod_compare/camourflage_00977.jpg}}&
    % {\includegraphics[width=0.093\linewidth,height=0.08\linewidth]{pic/cod_compare/camourflage_00977_mattiong_gt.png}}&
    % {\includegraphics[width=0.093\linewidth,height=0.08\linewidth]{pic/cod_compare/camourflage_00977_SINet.png}}&
    % {\includegraphics[width=0.093\linewidth,height=0.08\linewidth]{pic/cod_compare/camourflage_00977_PFNet.png}}&
    % % {\includegraphics[width=0.093\linewidth,height=0.08\linewidth]{pic/cod_compare/camourflage_00977_LSR.png}}&
    % {\includegraphics[width=0.093\linewidth,height=0.08\linewidth]{pic/cod_compare/camourflage_00977_C2FNet.png}}&
    % {\includegraphics[width=0.093\linewidth,height=0.08\linewidth]{pic/cod_compare/camourflage_00977_ERRNet.png}}&
    % {\includegraphics[width=0.093\linewidth,height=0.08\linewidth]{pic/cod_compare/camourflage_00977_ZoomNet.png}}&
    % {\includegraphics[width=0.093\linewidth,height=0.08\linewidth]{pic/cod_compare/camourflage_00977_LSR+.png}}&
    % {\includegraphics[width=0.093\linewidth,height=0.08\linewidth]{pic/cod_compare/camourflage_00977_un_color.png}}&
    % {\includegraphics[width=0.093\linewidth,height=0.08\linewidth]{pic/cod_compare/camourflage_00977_JSCOD.png}}\\

    {\includegraphics[width=0.093\linewidth,height=0.08\linewidth]{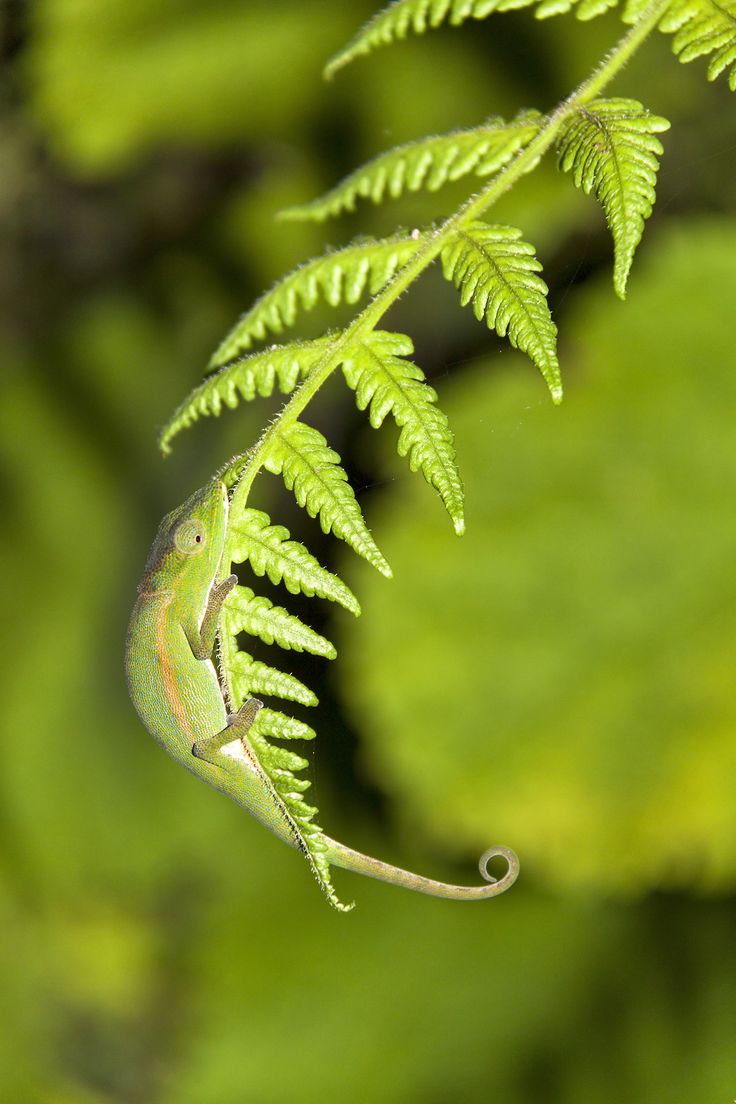}}&
    {\includegraphics[width=0.093\linewidth,height=0.08\linewidth]{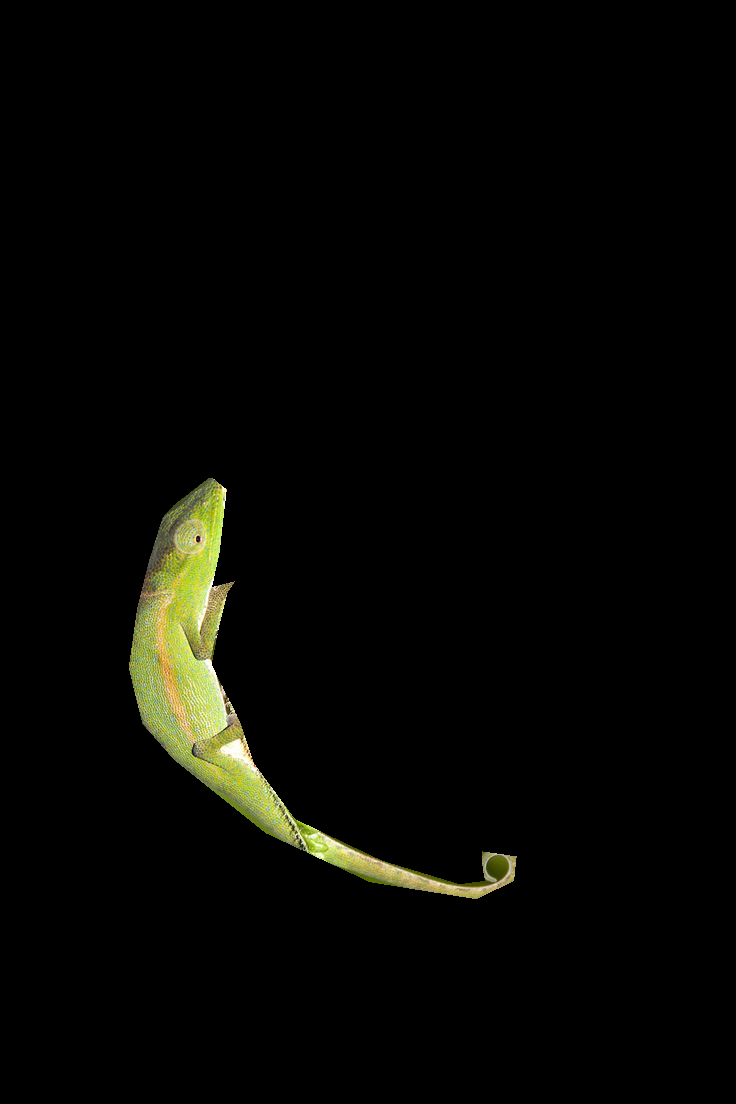}}&
    {\includegraphics[width=0.093\linewidth,height=0.08\linewidth]{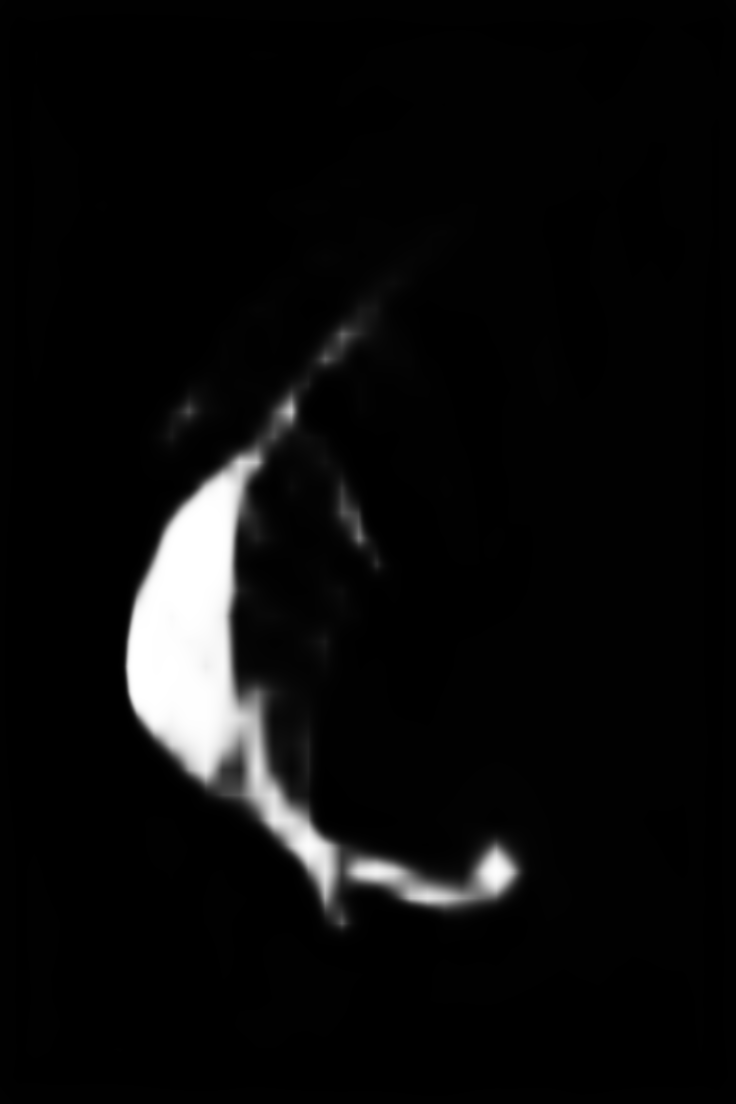}}&
    {\includegraphics[width=0.093\linewidth,height=0.08\linewidth]{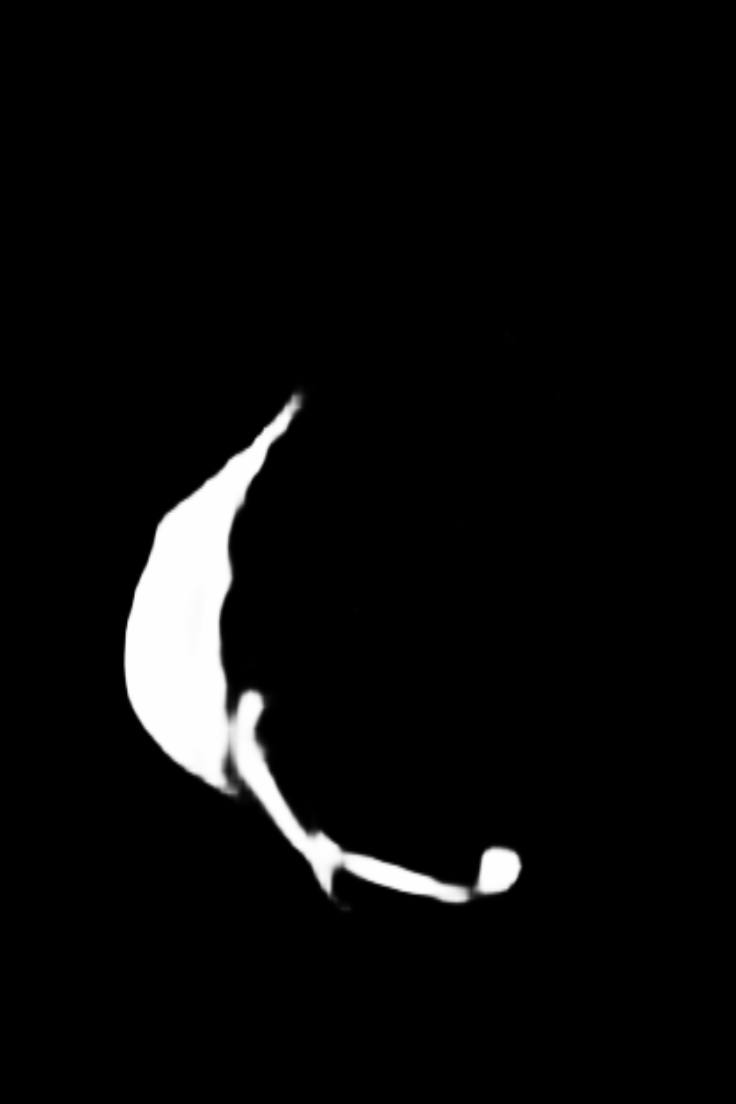}}&
    % {\includegraphics[width=0.093\linewidth,height=0.08\linewidth]{pic/cod_compare/camourflage_00903_LSR.png}}&
    {\includegraphics[width=0.093\linewidth,height=0.08\linewidth]{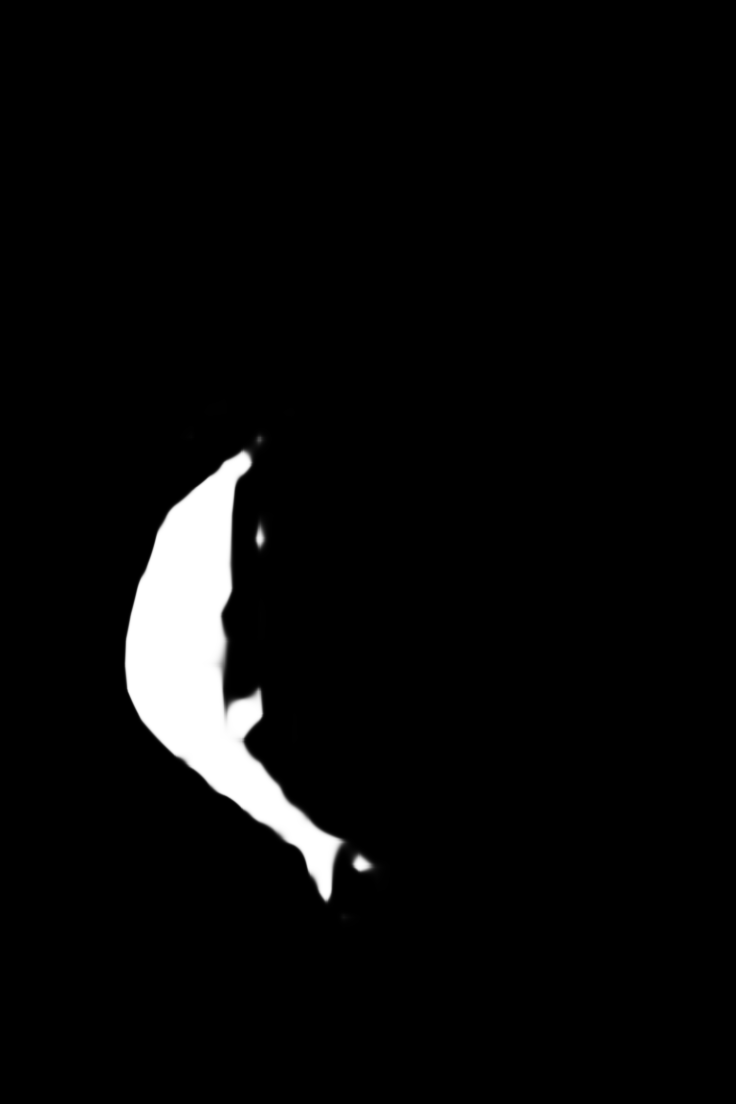}}&
    {\includegraphics[width=0.093\linewidth,height=0.08\linewidth]{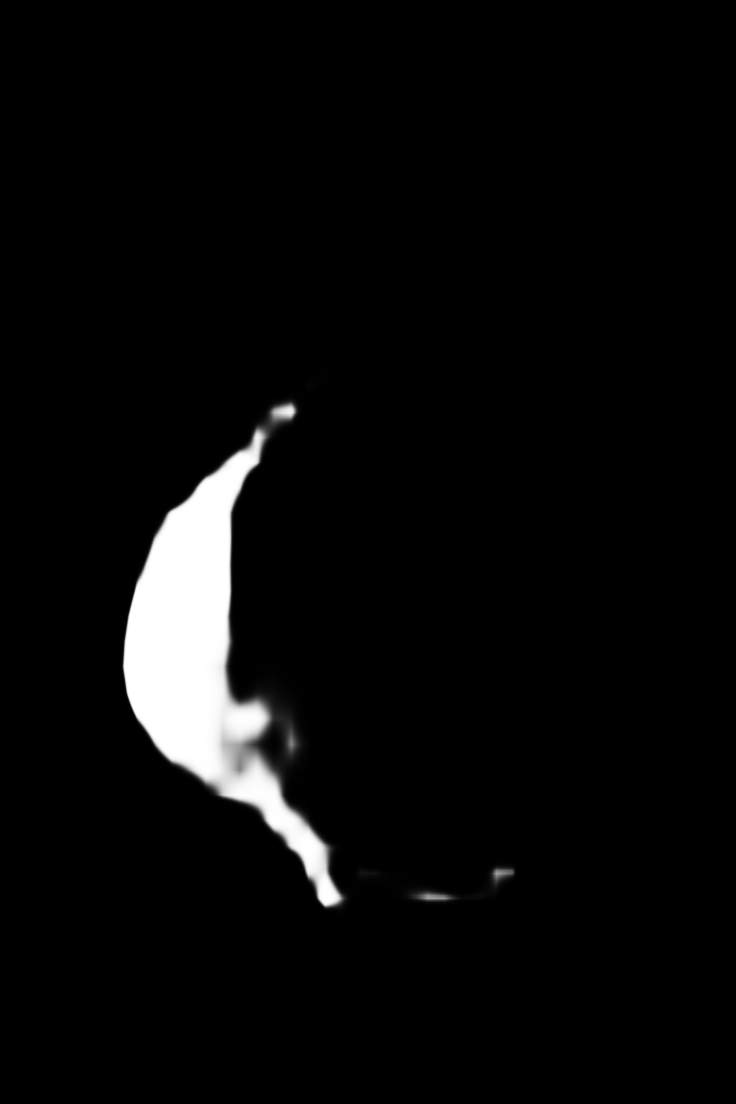}}&
    {\includegraphics[width=0.093\linewidth,height=0.08\linewidth]{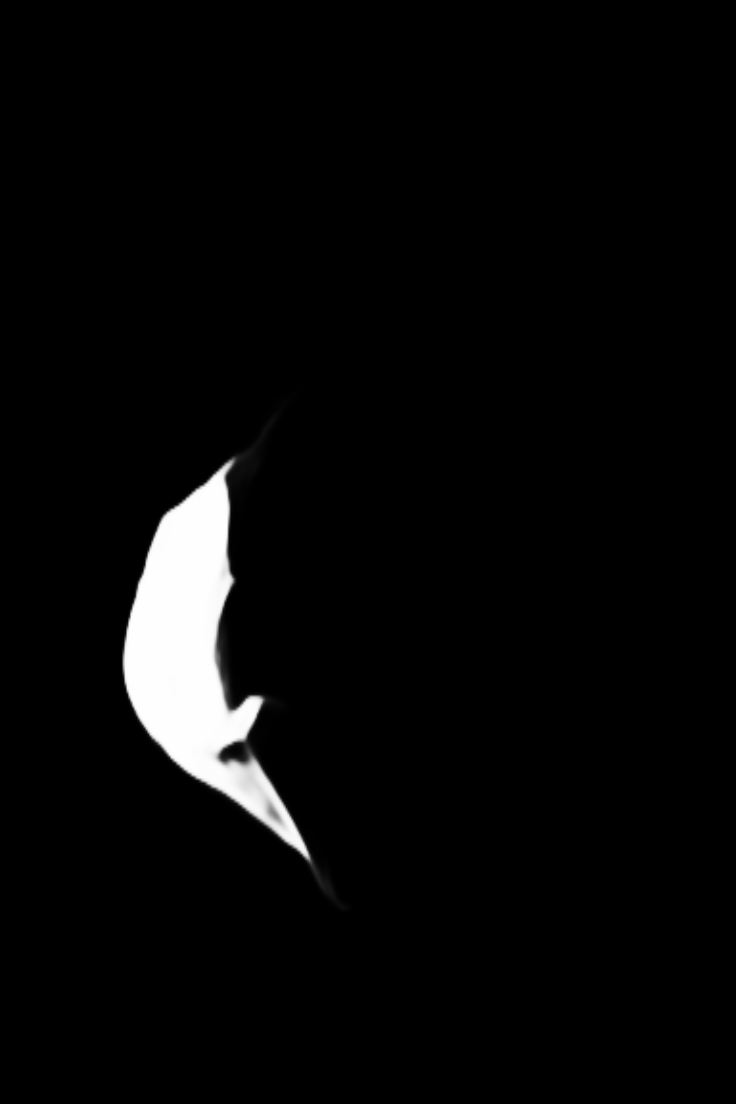}}&
    {\includegraphics[width=0.093\linewidth,height=0.08\linewidth]{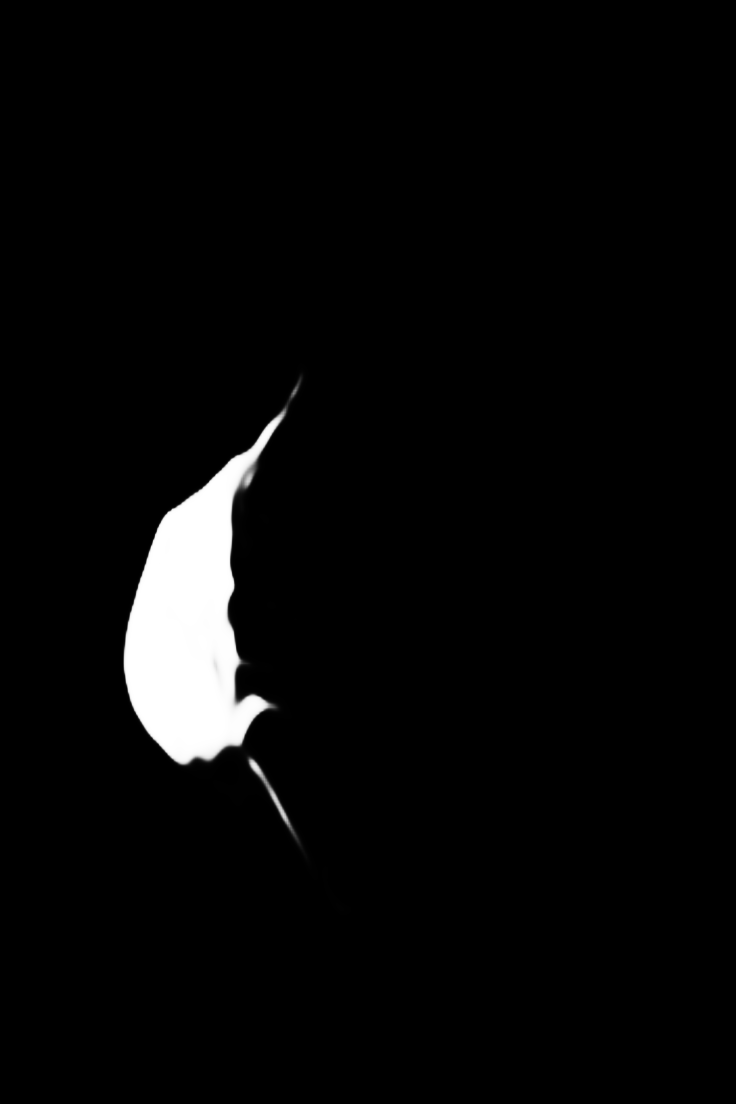}}&
    {\includegraphics[width=0.093\linewidth,height=0.08\linewidth]{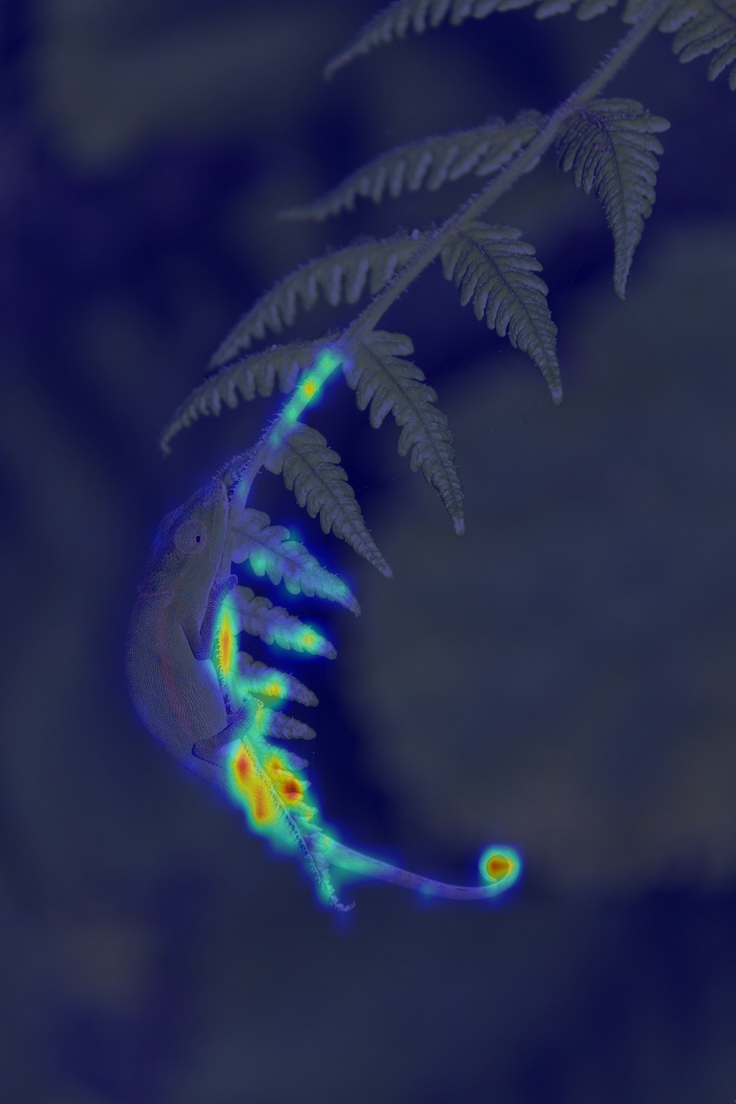}}&
    {\includegraphics[width=0.093\linewidth,height=0.08\linewidth]{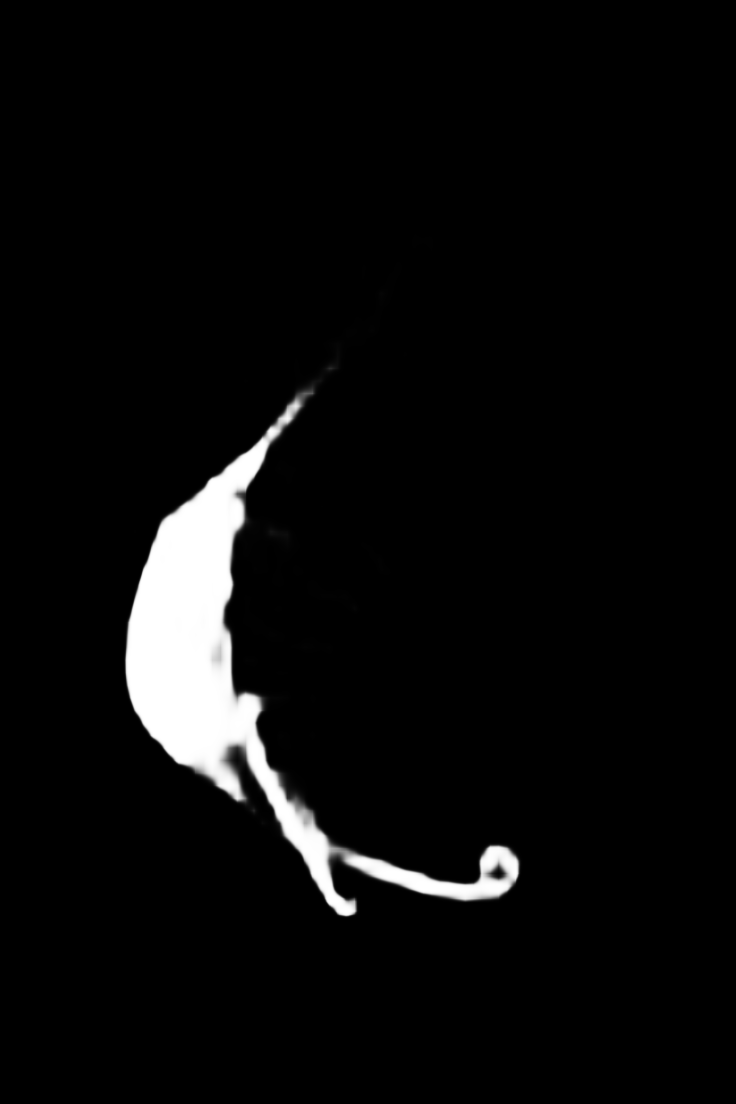}}\\
        
    \footnotesize{Image} &
     \footnotesize{GT} &\footnotesize{SINet~\cite{fan2020camouflaged}} &\footnotesize{PFNet~\cite{mei2021Ming}}&\footnotesize{C2FNet \cite{sun2021c2fnet}} &\footnotesize{ERRNet \cite{ji2022fast}} &\footnotesize{ZoomNet\cite{ZoomNet_CVPR2022}}&\footnotesize{LSR+~\cite{lv2023towards}}&\footnotesize{Uncertainty} &\footnotesize{Ours}\\
   \end{tabular}
   \end{center}
    \caption{
    % Predictions of competing camouflaged object detection models and ours.
    Visual comparison of our camouflage predictions with the state-of-the-art models.
    } 
    \label{fig:cod_samples_comparison}
\end{figure*}

\subsection{Performance Comparison}
% \Jing{I will come back to this part!}
% \AX{I'm here}
\noindent\textbf{Quantitative Analysis:}
We compare the performance of our SOD branch with SOTA SOD models as shown in Table~\ref{tab:benchmark_sod_model_comparison}. One observation from Table~\ref{tab:benchmark_sod_model_comparison} is that the structure-preserving strategy is widely used in the state-of-the-art saliency detection models, \eg SCRN~\cite{scrn_sal}, F$^3$Net~\cite{wei2020f3net}, ITSD~\cite{zhou2020interactive}, and it can indeed improve model performance.
% Table~\ref{tab:benchmark_sod_model_comparison} shows that we achieve 5/6 best performance, except on SOC testing dataset~\cite{fan2018SOC}. The main reason is that there exists texture images in SOC, which may be treated as camouflaged object, thus influence our performance. We will investigate this issue further.
Our method shows significant improvement in performance on four evaluation metrics compared to other SOD methods, except for the SOD dataset~\cite{sod_dataset}. Due to the small size of the SOD dataset~\cite{sod_dataset}(300 images), we believe that fluctuations in predictions are reasonable.
We also compare the performance of our COD branch with SOTA COD models in Table~\ref{tab:benchmark_cod_model_comparison}. Except for COD10k\cite{fan2020camouflaged}, where our method is slightly inferior to ZoomNet \cite{ZoomNet_CVPR2022}, our method shows significant superiority over all other COD methods on all datasets. The reason for this may be that ZoomNet \cite{ZoomNet_CVPR2022} was tested at resolution $384 \times 384$, while our method was tested at resolution $352 \times 352$, and resolution can affect the performance of COD. The consistent best performance of our camouflage model further illustrates the effectiveness of the joint learning framework. 
% As there exists limited deep camouflaged object detection models [][][],
% % (SINet~\cite{fan2020camouflaged} in particular), 
% due to similar setting of saliency detection, we re-train existing saliency detection models with the camouflaged object detection training dataset~\cite{fan2020camouflaged,le2019anabranch}, and test on the existing camouflaged object detection testing set. The performance of these models is shown in Table~\ref{tab:benchmark_cod_model_comparison} and Fig.~\ref{fig:cod_samples_comparison}. 

\noindent\textbf{Qualitative Analysis:}
Further, we show predictions of ours and SOTA models of SOD method in Fig.~\ref{fig:sod_samples_comparison}, and COD method in Fig.~\ref{fig:cod_samples_comparison}, where the \enquote{Uncertainty} is obtained based on the prediction from the discriminator.
Fig.~\ref{fig:sod_samples_comparison} shows that we produce both accurate prediction and reasonable uncertainty estimation, where the brighter areas of the uncertainty map indicate the less confident regions. It can be observed that our approach can better distinguish the boundaries between salient objects and the background. Fig.~\ref{fig:cod_samples_comparison} illustrates that our proposed joint learning approach and random-sampling based foreground cropping can better localize camouflaged targets.
% , while also having more accurate boundaries compared to other methods.
Further, the produced uncertainty map clearly represents model awareness of the prediction,
% towards the current prediction, 
leading to interpretable prediction for the downstream tasks.

\noindent\textbf{Run-time Analysis:}
For COD task, the inference time of our model is 53.9 ms per image. And for SOD task, the inference time of our model is 40.4 ms per image on an NVIDIA GeForce RTX 3090 GPU, which is comparable to the state-of-the-art model in terms of speed.
% \Jing{1 or 2 sentences sota run-time comparison}
% COD 349s 6473 img; SOD 679s 16784 image
% \Jing{more sod models}

% \noindent\textbf{Running time comparison:}
% \Jing{show the inference time with fps}

% \Jing{running time/model size comparison?}

% including PiCANet, CPD, SCRN, CSNet, PoolNet, F3Net, ITSD, BASNet, NLDF, EGNet and AFNet.

% \subsection{COD performance comparison}
% \Jing{more cod models}

% \Jing{running time/model size comparison?}

% Similar to the visual comparison of SOD models, we also show the predictions of competing COD models and ours in Fig. \ref{fig:cod_samples_comparison}, where the \enquote{Uncertainty} is the uncertainty map of our prediction.

% \noindent\textbf{Running time comparison:}
% \Jing{similarly, show the inference time with fps}

\subsection{Ablation Study}
We extensively analyze the proposed joint learning framework to explain the effectiveness of our strategies, and show the performance of our SOD and COD models in Table~\ref{tab:ablation_sod_model} and Table~\ref{tab:ablation_cod_model} respectively. Note that, unless otherwise stated, we do not perform multi-scale training for the related models.
% % , and on the camouflaged object detection dataset in 
% Table~\ref{tab:ablation_cod_model} for SOD and COD respectively.
% and
% % present extra experiments to fully explore our model, and 
% show model performance
% % of ablation study
% % on the salient object detection dataset 
% in Table~\ref{tab:ablation_sod_model} and
% % , and on the camouflaged object detection dataset in 
% Table~\ref{tab:ablation_cod_model} for SOD and COD respectively.
% Unless other clarified, the following experiments are based on the augmented SOD dataset.

\noindent\textbf{Train each individual task:
% without data interaction, connection modeling and uncertainty estimation:
}
% The basic baseline of our framework is to directly train each task separately. 
We use the same \enquote{Feature encoder}, \enquote{Prediction decoder} in Fig.~\ref{fig:network_overview} to train the SOD model with original DUTS dataset and the COD model trained without random-sampling based foreground cropping following the same training related setting as in the \enquote{Training details} section, and show their performance as \enquote{SSOD} and \enquote{SCOD}, respectively. And we used the augmented DUTS dataset and foreground cropping COD training dataset to train the SOD model and the COD model separately, the results are shown as \enquote{ASOD} and \enquote{ACOD}.
The comparable performance of \enquote{SSOD} and \enquote{SCOD} with their corresponding SOTA models proves the effectiveness of our prediction decoder.
% Further, the two data augmentation based models show slightly performance improvement compared with training directly with the raw dataset, especially for the COD task, where foreground cropping is applied. We generated the augmented SOD dataset via data interaction (see Sec.~\ref{subsub:data_interaction}), where the basic assumption is that easy samples for COD can serve as bard samples for SOD (see Fig.~\ref{fig:relationship_sod_cod}). Experimental results indicated limited performance improvement, which can be caused by our effective prediction decoder. We will further investigate data distribution issue with more effective data augmentation techniques in the future.
Further, the two data augmentation based models show clear performance improvement compared with training directly with the raw dataset, especially for the COD task, where foreground cropping is applied. We generated the augmented SOD dataset via data interaction (see Sec.~\ref{subsub:data_interaction} and Fig.~\ref{fig:relationship_sod_cod}).
 % , where the basic assumption is that easy samples for COD can serve as bard samples for SOD (see Fig.~\ref{fig:relationship_sod_cod}). 
 Experimental results show a reasonable performance improvement,
 % in performance, 
 indicating that our proposed data augmentation techniques are
 % is more 
 effective in enriching the diversity of the training data.
 % the data distribution.

% By comparing \enquote{SSOD} and \enquote{ASOD}, and comparing \enquote{SCOD} and \enquote{ACOD}, we have demonstrated the effectiveness of the data augmentation we proposed.

\begin{table*}[t!]
  \centering
  \scriptsize
  \renewcommand{\arraystretch}{1.3}
  \renewcommand{\tabcolsep}{1.2mm}
  \caption{Ablation studies on the salient object detection datasets.
  }
  \begin{tabular}{l|cccc|cccc|cccc|cccc|cccc|cccc}
  % \hline
\toprule
  &\multicolumn{4}{c|}{DUTS~\cite{wang2017learning}}&\multicolumn{4}{c|}{ECSSD~\cite{yan2013hierarchical}}&\multicolumn{4}{c|}{DUT~\cite{Manifold-Ranking:CVPR-2013}}&\multicolumn{4}{c|}{HKU-IS~\cite{li2015visual}}&\multicolumn{4}{c|}{PASCAL-S~\cite{pascal_s_dataset}}&\multicolumn{4}{c}{SOD~\cite{sod_dataset}} \\
    Method & $S_{\alpha}\uparrow$&$F_{\beta}\uparrow$&$E_{\xi}\uparrow$&$\mathcal{M}\downarrow$& $S_{\alpha}\uparrow$&$F_{\beta}\uparrow$&$E_{\xi}\uparrow$&$\mathcal{M}\downarrow$& $S_{\alpha}\uparrow$&$F_{\beta}\uparrow$&$E_{\xi}\uparrow$&$\mathcal{M}\downarrow$& $S_{\alpha}\uparrow$&$F_{\beta}\uparrow$&$E_{\xi}\uparrow$&$\mathcal{M}\downarrow$& $S_{\alpha}\uparrow$&$F_{\beta}\uparrow$&$E_{\xi}\uparrow$&$\mathcal{M}\downarrow$& $S_{\alpha}\uparrow$&$F_{\beta}\uparrow$&$E_{\xi}\uparrow$&$\mathcal{M}\downarrow$ \\ \hline
%   Ours & .899 & .866 & .937 & .032 & .933 & .935 & .960 & .030 & .850 & .782 & .884 & .051 & .931 & .924 & .867 & .026 & . & . & . & . & . & . & . & .  \\ \hline 
    Ours & .900 & .875 & .937 & .030 & .929 & .935 & .955 & .029 & .841 & .777 & .876 & .050 & .921 & .920 & .958 & .026 & .866 & .867 & .910 & .058 & .835 & .839 & .871 & .072 \\
    % \textbf{SOTA}& \textbf{.900} & \textbf{.876} & \textbf{.937} & \textbf{.032}  & \textbf{.933} & \textbf{.935} & \textbf{.960} & \textbf{.030} & \textbf{.853} & \textbf{.796} & \textbf{.888} & \textbf{.050} & \textbf{.931} & \textbf{.924} & \textbf{.967} & \textbf{.026} & \textbf{.869} & \textbf{.856} & \textbf{.902} & \textbf{.062}  & \textbf{.836} & \textbf{.836} & \textbf{.867} & \textbf{.067}  \\ 
    \hline 

    SSOD   & .887 & .850 & .924 & .035 & .924 & .924 & .952 & .032 & .837 & .767 & .871 & .053 & .917 & .910 & .954 & .028 & .861 & .854 & .904 & .062 & .848 & .844 & .885 & .065\\
    % SSOD   &.890 & .854 & .928 & .035 & .926 & .925 & .953 & .031 & .838 & .766 & .869 & .055 & .917 & .910 & .954 & .028 & .862 & .855 & .906 & .061 & .841 & .837 & .871 & .069\\
    ASOD &.891 & .860 & .928 & .033 & .927 & .928 & .953 & .031 & .842 & .777 & .875 & .050 & .917 & .914 & .955 & .027 & .863 & .859 & .906 & .061 & .836 & .835 & .874 & .070  \\  
    % ASOD &.890 & .853 & .926 & .035 & .928 & .928 & .955 & .030 & .835 & .764 & .866 & .055 & .919 & .912 & .956 & .027 & .864 & .856 & .906 & .061 & .839 & .836 & .874 & .069  \\  
    % ASOD &.891 & .856 & .929 & .034 & .926 & .925 & .953 & .032 & .838 & .769 & .872 & .052 & .917 & .911 & .955 & .027 & .860 & .850 & .902 & .061 & .845 & .842 & .875 & .066 \\
    JSOD1 & .892 & .857 & .928 & .034 & .929 & .930 & .955 & .030 & .836 & .765 & .869 & .054 & .919 & .913 & .956 & .027 & .862 & .856 & .906 & .062 & .839 & .840 & .875 & .071  \\  
    JSOD2 &.895 & .864 & .932 & .032 & .931 & .933 & .957 & .029 & .839 & .771 & .873 & .052 & .921 & .916 & .959 & .026 & .862 & .857 & .906 & .062 & .842 & .844 & .877 & .068\\
    JSOD3 & .893 & .861 & .931 & .033 & .928 & .931 & .956 & .030 & .835 & .766 & .869 & .053 & .919 & .914 & .956 & .027 & .863 & .859 & .908 & .060 & .844 & .848 & .883 & .065 \\
    
    % AdSOD &.891 & .859 & .929 & .034 & .929 & .929 & .956 & .030 & .839 & .769 & .870 & .053 & .918 & .912 & .955 & .027 & .863 & .854 & .904 & .061 & .844 & .841 & .876 & .067 \\ % 
    % AdSOD &.891 & .856 & .928 & .034 & .926 & .925 & .952 & .031 & .839 & .769 & .872 & .052 & .919 & .912 & .956 & .027 & .862 & .853 & .906 & .061 & .845 & .844 & .880 & .066 \\ % 
    % AdSOD &.889 & .854 & .927 & .034 & .926 & .925 & .953 & .031 & .839 & .770 & .873 & .051 & .918 & .912 & .956 & .027 & .862 & .856 & .905 & .061 & .837 & .834 & .869 & .068 \\ % 
   % CSOD & . & . & . & . & . & . & . & . & . & . & . & . & . & . & . & . & . & . & . & . & . & . & . & .   \\ %CSOD\_1+conf\_w=0.007
   % \hline 
\bottomrule
  \end{tabular}
  \label{tab:ablation_sod_model}
%   \vspace{-5mm}
\end{table*}

\begin{table*}[t!]
  \centering
  \scriptsize
  \renewcommand{\arraystretch}{1.3}
  \renewcommand{\tabcolsep}{2.8mm}
  \caption{Ablation studies on the camouflaged object detection datasets.}
  \begin{tabular}{l|cccc|cccc|cccc|cccc}
  % \hline
  \toprule
  &\multicolumn{4}{c|}{CAMO}&\multicolumn{4}{c|}{CHAMELEON}&\multicolumn{4}{c|}{COD10K}&\multicolumn{4}{c}{NC4K} \\
    Method & $S_{\alpha}\uparrow$&$F_{\beta}\uparrow$&$E_{\xi}\uparrow$&$\mathcal{M}\downarrow$& $S_{\alpha}\uparrow$&$F_{\beta}\uparrow$&$E_{\xi}\uparrow$&$\mathcal{M}\downarrow$ &  $S_{\alpha}\uparrow$ & $F_{\beta}\uparrow$ & $E_{\xi}\uparrow$ & $\mathcal{M}\downarrow$&  $S_{\alpha}\uparrow$ & $F_{\beta}\uparrow$ & $E_{\xi}\uparrow$ & $\mathcal{M}\downarrow$  \\
  \hline
%   Ours  & .803 & .759 & .853 & .076 & .894 & .848 & .943 & .030 & .817 & .726 & .892 & .035 & .& .& .& .   \\ 
   Ours  & .803 & .768 & .858 & .071 & .892 & .848 & .948 & .025 & .817 & .733 & .895 & .033 & .856 & .824 & .913 & .040  \\ 
   % ZoomNet \cite{ZoomNet_CVPR2022} & .789 & .741 & .829 & .076 & .865 & .823 & .939 & .031 & .821 & .741 & .866 & .032 & .839 & .796 & .867 & .046 \\
   \hline

    % SCOD & .793 & .762 & .842 & .076 & .896 & .859 & .949 & .029 & .805 & .717 & .883 & .036 & .841 & .808 & .898 & .047 \\ 
    SCOD & .784 & .754 & .843 & .077 & .895 & .859 & .946 & .030 & .797 & .708 & .880 & .037 & .833 & .803 & .897 & .049 \\ 
    ACOD & .805 & .771 & .863 & .071 & .886 & .840 & .943 & .029 & .810 & .711 & .887 & .036 & .851 & .809 & .906 & .043 \\ 
 
    JCOD1 & .816 & .781 & .863 & .068 & .893 & .845 & .946 & .027 & .815 & .722 & .889 & .035 & .856 & .817 & .906 & .042    \\   
    JCOD2 &  .805 & .772 & .860 & .074 & .886 & .841 & .938 & .029 & .812 & .719 & .891 & .035 & .853 & .814 & .907 & .042   \\ 
%   JCOD3 & .793 & .747 & .850 & .078  & .887 & .840 & .943 & .029 & .807 & .717 & .885 & .037 & .& .& .& . \\
  JCOD3  &  .808 & .785 & .868 & .068 & .892 & .853 & .951 & .027 & .810 & .723 & .890 & .034 & .850 & .815 & .906 & .041 \\
%   JCOD3 & .784 & .752 & .840 & .076  & .891 & .849 & .947 & .028 & .799 & .716 & .884 & .036 & .836 & .808 & .898 & .047 \\  
   % AdCOD  & .803 & .773 & .864 & .071 & .887 & .839 & .946 & .028 & .807 & .712 & .891 & .035 & .850 & .812 & .909 & .042 \\ 
%   AdCOD & .783 & .753 & .844 & .078  & .895 & .853 & .953 & .027 & .796 & .789 & .884 & .037 & .834 & .805 & .898 & .048  \\ 
   % CCOD  &  &  &  &  &  &  &  &  &  &  &  &  &  &  &  &   \\      
%   CCOD & .780 & .754 & .844 & .079 & .888 & .850 & .953 & .027 & .798 & .718 & .884 & .036 & .832 & .803 & .895 & .048  \\
%   JCOD4 &  &  &  &   & . & . & . & . & . & . & . & .  \\
   % \hline 
   \bottomrule

  \end{tabular}
  \label{tab:ablation_cod_model}
  \vspace{-2mm}
\end{table*}

\noindent\textbf{Joint training of SOD and COD:}
% As the basic joint learning baseline, we can 
% We train the \enquote{Feature encoder} and \enquote{Prediction decoder} within a joint learning pipeline to achieve simultaneous salient object detection and camouflaged object detection. The performance is shown as \enquote{JSOD1} and \enquote{JCOD1}, respectively. The slightly improved performance compared with the uni-task setting indicates effectiveness of the joint learning framework, especially for COD.
We train the \enquote{Feature encoder} and \enquote{Prediction decoder} within a joint learning pipeline to achieve simultaneous SOD and COD. The performance is reported as \enquote{JSOD1} and \enquote{JCOD1}, respectively. For the COD task, there was a slight improvement in performance compared to the uni-task setting, indicating that under the joint learning framework, SOD can provide effective prediction optimization for COD. For SOD task, there was a slight decrease in performance, which we believe is due to the lack of consideration of the \enquote{contradicting} attribute between the two tasks. The subsequent experiments in the paper fully demonstrate this point.

% for the saliency detection task and camouflaged object detection task respectively. 
% Compared with \enquote{JSOD1}  and \enquote{ASOD}, it is found that simply conducting joint learning of two tasks does not offer much assistance to SOD. However, the improved performance of \enquote{JCOD1}  and \enquote{ACOD} demonstrated that the network architecture with a shared decoder can help COD branch learn fine-grained structure prediction.

\noindent\textbf{Joint training of SOD and COD with contrastive learning:}
% \noindent\textbf{Joint training of SOD and COD with the similarity measure module}\\
We add the task connection constraint to the joint learning framework, \ie~the contrastive module in particular, and show performance as \enquote{JSOD2} and \enquote{JCOD2} respectively. As discussed in Sec.~\ref{sub:contradicting_modeling}, our contrastive module is designed to enhance the context information, and the final results show performance improvement for SOD. However, we observe deteriorated performance for COD when the contrastive module is applied. We have analyzed the predictions and find that the context enhancement strategy via contrastive learning can be a double-edged sword, which is effective for SOD but leads to performance deterioration for COD. Different from the conventional way of constructing positive/negative pairs based on augmentation or category information, SOD and COD are both class-agnostic tasks, and our positive/negative pairs are designed based on contrast information. Experimental results explain its effectiveness for high-contrast based foreground detection, \ie~salient object detection, while minimal context difference between foreground and background of COD poses new challenges for applying contrastive learning effectively to achieve distinguishable foreground/background feature representation.
% In general, we can observe improved performance, especially for the COD10K dataset~\cite{fan2020camouflaged}, which verifies effectiveness of our similarity measure module.

\noindent\textbf{Joint adversarial training of SOD and COD:}
Based on the joint learning framework (\enquote{JSOD1} and \enquote{JCOD1}), we further introduce the adversarial learning pipeline, and show performance as \enquote{JSOD3} and \enquote{JCOD3}. We observe relatively comparable performance of \enquote{JSOD3} (\enquote{JCOD3}) to \enquote{JSOD1} (\enquote{JCOD1}), explaining that the adversarial training pipeline will not sacrifice model deterministic performance. Note that with adversarial training, our model can output prediction uncertainty with single forward, serving as an auxiliary output to explain confidence of model output (see \enquote{Uncertainty} in Fig.~\ref{fig:sod_samples_comparison} and Fig.~\ref{fig:cod_samples_comparison}).

\noindent\textbf{The proposed joint framework:} We report our final model performance with both the contrastive module and the adversarial learning solution
% \AX{with multi-scale} 
as \enquote{Ours}. As a dual-task learning framework, \enquote{Ours} shows improved performance compared with models with each individual strategy, \ie~contrastive learning and adversarial training. As discussed in Sec.~\ref{sec:introduction}, the former is introduced to model the task-wise correlation, and the latter is presented to model the inherent uncertainty within the two tasks. Although these two strategies show limitations for some specific datasets, we argue that as a class-agnostic task, both our contrast based positive/negative pair construction for contrastive learning and residual learning based discriminator learning within the adversarial training pipeline are effective in general, and more investigation will be conducted to further explore their contributions for the joint learning of the contradictory tasks.

\subsection{Framework Analysis}
As discussed in Sec.~\ref{sec:task_relationship}, SOD and COD are correlated from both task's point of view and the data's perspective. In this Section, we further analyze their relationships and the inherent uncertainty modeling techniques for SOD and COD.

% \Jing{I'm here!}

\subsubsection{Data interaction analysis}
SOD and COD are both context based tasks
% often related to the surrounding environment 
(see Fig.~\ref{fig:relationship_sod_cod}), and can be transformed into each other, where the former represents the attribute of object(s) with high-contrast and the latter is related to concealment.
% Therefore, it is inadequate to simply classify SOD and COD as binary tasks with only dataset differences. Salience and camouflage inherently exhibit opposition and correlation, and certain objects possess both salient and camouflage attributes, as illustrated in Fig. \ref{fig:selected_easy_samples_from_cod}.
Considering the opposite object attribute of saliency and camouflage, we introduce a simple data selection strategy as data augmentation for saliency detection. 
% Additionally, to account for cases where certain samples may have a small MAE metric due to the inherently  small size of the object, we proposed $wMAE$, defined in Eq.~\eqref{wmae}. 
Based on the nature of the two task, we explicitly connected the SOD and COD datasets. Experimental results show that incorporating an additional $3.8\%$ of data, specifically 403 out of 10,553 images, led to performance improvement for SOD, comparing \enquote{ASOD} and \enquote{SSOD} in Tabel~\ref{tab:ablation_sod_model}.

\subsubsection{Task interaction analysis}
% --Contradicting modeling related experiments
In our preliminary version~\cite{aixuan_cod_sod21},
% previous paper, 
we used the entire PASCAL VOC 2007 as a bridge dataset to model the contradictory properties of SOD and COD via similarity modeling. Here, we apply contrative learning based on contrast information instead, which is proven effective for SOD, comparing \enquote{JSOD2} and \enquote{JSOD1} in Tabel~\ref{tab:ablation_sod_model}. As contrastive learning is sensitive to the positive/negative pools, and PASCAL VOC 2007 dataset contains samples that pose challenges for either SOD or COD to decide the foreground, we thus selected a portion of the images from the bridge dataset as the updated PASCAL dataset. Specifically, we tested the PASCAL VOC 2007 dataset using the trained SOD and COD models to obtain the weighted MAE of the SOD and COD prediction maps. Then, we selected 200 images from the PASCAL VOC 2007 dataset with the smallest weighted MAE as the new bridge dataset for training the contradicting modeling module. 
The contradicting module is trained every 5 steps of the other modules to avoid involving feature conflicting for COD. Although our contrastive learning solution is proven effective for SOD, the final performance still shows deteriorated performance of COD, comparing \enquote{JCOD2} and \enquote{JCOD1} in Tabel~\ref{tab:ablation_cod_model}. The main reason is that the contrastive learning module tries to push the feature spaces of foreground and background to be close as Eq.~\eqref{latent_loss}, while the main task of COD is to distinguish the foreground from the background. The contradicting objectives pose challenges for the COD task to converge.
% \YC{Please ref to either Table or figure for these conclusions. How the above conclusion is obtained.}
% contradicting modeling and data interaction conflicts. 
% When training the contrastive module, it is essential to note that we use the prediction in Eq.~\eqref{fore_back_fea} during training the contradicting modeling module, but we only use the prediction for dividing foreground and background features, and this is the contrastive module update do not affect the Prediction Decoder.
% salient foreground stands out from salient background, while the camouflage foreground is hidden in the camouflage background, 
 % Moreover, we select positive and negative correlation pairs based on the similarity of salient and camouflage foreground and background pairs for the same image.

\subsubsection{Discriminator analysis}
Considering that
% both 
the uncertainty regions of both tasks
% and the hard-to-predict components 
are associated with the image, we concatenate the prediction/ground truth with the image, and feed it to the discriminator. We define the portions of a network's incorrect predictions as areas that are difficult to learn following
% , i.e. the residual between the  ground truth and the prediction $|pred - gt|$, which is similar to 
\cite{franchi2022latent}. 
% For ground truth without noise as input, the unpredictable region is $0$. But the subjective nature of SOD and labeling uncertainty of COD, there is noise in the ground truth image (Section\ref{subsub:Inherent_Uncertainty}),  so when the noisy ground truth  is used as input, the corresponding discriminator output should not be $0$, but should correspond to the noisy region. 
% However, since noise cannot be directly labeled, and considering the consistency of the distribution of the complete dataset, we believe that in the framework of adversarial learning and consistency regularization, the discriminator network can fit the uncertain areas of the labels during the training phase, as shown in Fig.
% However, inherent uncertainty cannot be directly labeled, but annotations in the same dataset often follow the same distribution, with correctly annotated regions dominating the distribution. Therefore, we used adversarial learning to help model the uncertainty. 
In the early stages of training, the network fits the correctly annotated regions, and in later training, the predicted maps gradually approach the ground truth maps with the uncertainty/noise annotations \cite{huang2019o2u,han2018co,ma2018dimensionality}.
When introducing image information, the areas that are difficult to predict or annotated incorrectly (inherent uncertainty) can be gradually discovered under the guidance of RGB image.

\subsection{Hyper-parameters analysis}
In our joint learning framework, several hyper-parameters affect our final performance, including the maximum iterations, the base learning rates, weights for the contrastive learning loss function and the adversarial loss function.
We found that although the training dataset size of SOD is three times of the COD dataset,
% is roughly 3:1, 
the COD images are more complex than the SOD images. Therefore, we kept the same numbers of iterations for SOD and COD tasks.
Due to the overlapping regions of saliency and camouflage, for the contrastive learning module, we trained it every 5 steps to avoid involving too much conflicting to COD.
% too much impact on the main network when establishing the conflicting properties, and 
With the same goal, we set the weight of the contrastive loss to 0.1.
For the \enquote{Confidence estimation} module, we observed that
% the network predictions gradually approach the ground truth in the later stages of training, and 
excessively large adversarial training loss may lead to over-fitting on noise.
% the network fitting to noise.
Our main goal of using the adversarial learning is to provide reasonable uncertainty estimation. In this case, we define the ground truth output of the discriminator as the residual between the main network prediction and the corresponding ground truth, and set the weight of Eq.~\eqref{adv_pred_sod} and Eq.~\eqref{adv_pred_cod} as 1.0, to achieve trade-off between model performance and effective uncertainty estimation.

\section{Conclusion}
% In this paper, we have 
In this paper, we proposed the first joint salient object detection and camouflaged object detection framework to
% within a unified uncertainty-aware framework, which 
explore the contradicting nature of these two tasks. Firstly, we conducted an in-depth analysis on the intrinsic relationship of the two tasks.
% between saliency and camouflage.
% , which are inherently contradictory attributes. 
Based on it, we designed a contrastive module to model the task-wise correlation, and
% natural oppositional properties of foreground and background features for saliency and camouflaged. 
% Secondly, we have observed overlapping regions of the same image for
% % here is an overlap between 
% saliency and camouflaged detection. Thus, 
a data interaction strategy to achieve
% was introduced to SOD,
% data augmentation, 
% achieving
% to help SOD achieve 
context-aware data augmentation for SOD. Secondly, considering that camouflage is a local attribute, we proposed random sampling-based foreground-cropping as the COD data augmentation technique.
% , which significantly improves the performance of COD. 
Finally,
% we used an 
uncertainty-aware learning is explored to produce uncertainty estimation with single forward. 
Experimental results across different datasets prove the effectiveness of our proposed joint learning framework. We observed that although contrast-based task-wise contrastive learning is proven effective for SOD, it damages the performance of COD due to the contradicting attribute of these two tasks. More investigation will be conducted to further explore informative feature representation learning via contrastive learning for class-agnostic tasks.
% module to help model inherent uncertainty and network prediction difficulties with adversarial learning.
% In the future, we plan to extend our work to generic object detection.

% % use section* for acknowledgment
% \ifCLASSOPTIONcompsoc
   % The Computer Society usually uses the plural form
%   \section*{Acknowledgments}
%   This research was supported in part by National Natural Science Foundation of China (62271410).
% \else
   % regular IEEE prefers the singular form
%   \section*{Acknowledgment}
%  \fi

\bibliographystyle{ieeetr}
\bibliography{SOD_COD_Reference}

%  \clearpage

%  You can push biographies down or up by placing
% a \vfill before or after them. The appropriate
% use of \vfill depends on what kind of text is
% on the last page and whether or not the columns
% are being equalized. 

\vfill
\end{document}